\documentclass[11pt,a4paper]{article}
\usepackage[hyperref]{emnlp-ijcnlp-2019}
\usepackage{times}
\usepackage{graphicx}
\usepackage{xspace}
\usepackage{latexsym}
\usepackage{amsmath}
\usepackage{amssymb}
\usepackage{xcolor}
\usepackage{caption}
\usepackage{subcaption}
\usepackage{booktabs}
\usepackage{bm}
\usepackage{enumitem}
\usepackage{diagbox}

\PassOptionsToPackage{hyphens}{url}
\hypersetup{breaklinks=true}
\aclfinalcopy %

\newcommand{\hypothesis}[1]{{\bf #1}\xspace}
\newcommand{\para}[1]{\noindent {\bf #1}\xspace}
\newcommand{\empara}[1]{\noindent {\em #1}\xspace}
\newcommand{\figref}[1]{Figure \ref{#1}}

\newcommand{\secref}[1]{\S\ref{#1}}
\newcommand{\tableref}[1]{Table \ref{#1}}

\newcommand{\importance}{\mathcal{I}}
\newcommand{\real}{\mathbb{R}}

\title{Many Faces of Feature Importance: Comparing Built-in and Post-hoc Feature Importance in Text Classification}

\author{Vivian Lai \and Zheng Cai \and Chenhao Tan\\
  Department of Computer Science \\
  University of Colorado Boulder \\
  Boulder, CO \\
  {\tt vivian.lai, jon.z.cai, chenhao.tan@colorado.edu} \\
}

\date{}

\begin{document}
\maketitle

\begin{abstract}
Feature importance is commonly used to explain machine predictions.
While feature importance can be derived from a machine learning model with a variety of methods,
the consistency of feature importance via different methods remains understudied.
In this work,
we 
systematically compare feature importance from built-in mechanisms in a model such as attention values and post-hoc methods that approximate model behavior such as LIME.
Using text classification as a testbed, we find that 1)
no matter which method we use, important features from traditional models such as SVM and XGBoost are more similar with each other, than with deep learning models;
2) post-hoc methods tend to generate more similar important features for two models than built-in methods.
We further demonstrate how such similarity varies across instances.
Notably, important features do {\em not} 
always resemble each other better when two models agree on the predicted label than when they disagree.
\end{abstract}

\section{Introduction}
\label{sec:intro}

As machine learning models are 
adopted in societally important tasks such as recidivism prediction and loan 
approval, explaining machine predictions has become increasingly important \citep{doshi2017towards,lipton2016mythos}.
Explanations 
can potentially
improve the trustworthiness of algorithmic decisions for decision makers,  facilitate model developers in debugging, and even allow regulators to identify biased algorithms.

A popular approach to explaining machine predictions is to identify important features for a particular prediction \citep{luong-etal-2015-effective,ribeiro2016should,lundberg2017unified}.
Typically, these explanations assign a value to each feature (usually a word in NLP), and thus enable visualizations such as highlighting top $k$ features.

In general, there are two classes of methods:
1) built-in feature importance that is embedded in the machine learning model such as coefficients in linear models and attention values in attention mechanisms;
2) post-hoc feature importance through credit assignment based on the model such as LIME.
It is well recognized that robust evaluation of 
feature importance is challenging \citep[][inter alia]{jain2019,nguyen2018comparing}, which is further complicated by different use cases of explanations (e.g., for decision makers vs. for developers). 
Throughout this work, we refer to machine learning models that learn from data as {\em models} and methods to obtain local explanations (i.e., feature importance in this work) for a prediction by a model as {\em methods}.

\begin{table*}
\small
\centering
\begin{tabular}{l|p{3cm}|p{2.8cm}|p{3cm}|p{3cm}}
\toprule
\multicolumn{5}{p{15cm}}{%
One of favorite places to eat on the King W side, simple and relatively quick.
I typically always get the chicken burrito and the small is enough for me for dinner.
Ingredients are always fresh and watch out for the hot sauce cause it's skull scratching hot.
Seating is limited so be prepared to take your burrito outside or you can even eat at Metro Hall Park.
} \\ 
\midrule
\diagbox[innerwidth=1.7cm]{methods}{models} & SVM ($\ell_2$) & XGBoost & LSTM with attention & BERT \\
\hline
built-in & sauce, seating, park, prepared, even, always, can, fresh, quick, favorite & is, can, quick, fresh, at, to, always, even, favorite, and & me, be, relatively, enough, always, fresh, ingredients, prepared, quick, favorite & ., ingredients, relatively, quick, places, enough, dinner, typically, me, i \\
\hline
LIME & the, dinner, be, quick, and, even, you, always, fresh, favorite & you, to, fresh, quick, at, can, even, always, and, favorite & dinner, ingredients, typically, fresh, places, cause, quick, and, favorite, always & one, watch, to, enough, limited, cause, and, fresh, hot, favorite \\
\bottomrule
\end{tabular}
\vspace{-0.1in}
\caption{10 most important features (separated by comma) identified by different methods for different models for the given review. In the interest of space, we only show built-in and LIME here.
}
\label{tb:example}
\vspace{-0.1in}
\end{table*}

While prior research tends to focus on the internals of models in designing and evaluating methods of explanations, e.g., how well explanations reflect the original model \citep{ribeiro2016should}, we 
view feature importance itself as a subject of study,
and aim to provide a systematic characterization of important features obtained via different methods for different models.
This view is particularly important when explanations are used to support decision making because they are the only exposure to the model for decision makers.
It would be desirable that explanations are consistent across different instances.
In comparison, debugging represents a distinct use case where developers often know the mechanism of the model beyond explanations.
Our view also connects to studying explanation as a {\em product} 
in cognitive studies of explanations \citep{lombrozo2012explanation}, and is complementary to the model-centric perspective.

Given a wide variety of models and methods to generate feature importance, there are basic open questions such as how similar 
important features are between models and 
methods,
how important features distribute across instances,
and what linguistic properties important features tend to have.
We use text classification as a testbed to answer these questions.
We consider built-in importance from both traditional models such as linear SVM and neural models with attention mechanisms, 
as well as post-hoc importance based on LIME and SHAP.
\tableref{tb:example} shows important features for a Yelp review in sentiment classification.
Although most approaches consider ``fresh'' and ``favorite'' important, there exists significant variation.

We 
use three text classification tasks to characterize the overall similarity between important features.
Our analysis reveals the following insights:

\begin{itemize}[itemsep=0pt,leftmargin=*,topsep=2pt]
    \item (Comparison between approaches) Deep learning models generate more different important features from traditional models such as SVM and XGBoost.
    Post-hoc methods tend to reduce the dissimilarity between models by making important features more similar than the built-in method. 
    Finally, {different approaches do not generate more similar important features even if we focus on the most important features (e.g., top one feature)}.
    \item (Heterogeneity between instances) {Similarity between important features is not always greater when two models agree on the predicted label}, and longer instances are less likely to share important features.
    \item  (Distributional properties) {Deep models generate more diverse important features with higher entropy, which indicates lower consistency across instances}.
    Post-hoc methods bring the POS distribution closer to background distributions.
\end{itemize}

In summary, our work systematically compares important features 
from different methods for different models, and sheds light on how different models/methods induce important features.
Our work 
takes the first step to understand important features as a product and helps inform the adoption of feature importance for different purposes.
Our code is available at \url{https://github.com/BoulderDS/feature-importance}.

\section{Related Work}
\label{sec:related}

To provide further background for our work, we summarize current popular approaches to generating and evaluating explanations of machine predictions, with an emphasis on feature importance.

\para{Approaches to generating explanations.}
A battery of approaches have been recently proposed to explain machine predictions (see 
\citet{guidotti2019survey}
for an overview), including example-based approaches that identifies ``informative'' examples in the training data \citep[e.g.,][]{kim2016examples} and rule-based approaches that reduce complex models to simple rules \citep[e.g.,][]{malioutov2017learning}.
Our work focuses on characterizing properties of feature-based approaches.
Feature-based approaches tend to identify important features in an instance and enable visualizations with important features highlighted.
We discuss several directly related post-hoc methods here and introduce the built-in methods in \secref{sec:approach}.
A popular approach, LIME, fits a sparse linear model to approximate model predictions locally \citep{ribeiro2016should};
\citet{lundberg2017unified} present a unified framework based on Shapley values, which can be computed with different approximation methods for different models.
Gradients are popular for identifying important features in deep learning models since these models are usually differentiable \citep{shrikumar2017learning}, for instance, \citet{li2015visualizing} uses gradient-based saliency to compare LSTMs with simple recurrent networks.

\para{Definition and evaluation of explanations.}
Despite a myriad of studies on 
approaches to explaining machine predictions,
 explanation is a rather overloaded term and evaluating explanations is challenging.
\citet{doshi2017towards} lays out three levels of evaluations: functionally-grounded evaluations based on proxy automatic tasks, human-grounded evaluations with laypersons on proxy tasks, and application-grounded based on expert performance in the end task.
In text classification, \citet{nguyen2018comparing} shows that automatic evaluation based on word deletion moderately correlate with human-grounded evaluations that ask crowdworkers to infer machine predictions based on explanations.
However, explanations that help humans infer machine predictions may not actually help humans make better decisions/predictions.
In fact, recent studies find that feature-based explanations alone have limited improvement on human performance in detecting deceptive reviews and media biases \citep{lai+tan:19,horne2019rating}.

In another recent debate,
\citet{jain2019} 
examine attention as an explanation mechanism based on how well attention values correlate with gradient-based feature importance and whether they exclusively lead to the predicted label, and conclude that attention is not explanation. 
Similarly, \citet{serrano:2019} show that attention is not a fail-safe indicator for explaining machine predictions based on intermediate representation erasure.
However, \citet{pinter:2019} argue that attention can be explanation depending on the definition of explanations (e.g., plausibility and faithfulness).

In comparison, we treat 
feature importance itself as a subject of study and compare different approaches to obtaining feature importance from a model.
Instead of providing a normative judgment with respect to what makes good explanations, our goal is to allow decision makers or model developers to make informed decisions based on 
properties of important features 
using different models and methods.

\section{Approach}
\label{sec:approach}

In this section, we first formalize the problem of obtaining feature importance and then introduce the models and methods that we consider in this work. Our main contribution 
is to compare important features identified for a particular instance through different methods for different models.

\para{Feature importance.}
For any instance $t$ and a machine learning model $m: t \rightarrow y \in \{0, 1\}$, we use method $h$ to obtain feature importance on an interpretable representation of $t$, $\importance_{t}^{m,h} \in \real^d$, where $d$ is the dimension of the interpretable representation.
In the context of text classification, we use unigrams as the interpretable representation.
Note that the machine learning model does not necessarily use the interpretable representation.
Next, we introduce the models and methods in this work.

\para{Models ($m$).}
We include both recent deep learning models for NLP and popular machine learning models that are not based on neural networks.
In addition, we make sure that the chosen models have some built-in mechanism for inducing feature importance and describe the built-in feature importance as we introduce the model.\footnote{For instance, we do not consider LSTM as a model here due to the lack of commonly-accepted built-in mechanisms.
}

\begin{itemize}[itemsep=0pt,topsep=0pt,leftmargin=*]
    \item Linear SVM with $\ell_2$ (or $\ell_1$) regularization. Linear SVM has shown strong performance in text categorization \citep{joachims1998text}. 
    The absolute value of coefficients in these models is typically considered a measure of feature importance \citep[e.g.,][]{ott2011finding}.
    We also consider $\ell_1$ regularization because $\ell_1$ regularization is often used to induce sparsity in the model.
    \item Gradient boosting tree (XGBoost).
    XGBoost represents an ensembled tree algorithm that shows strong performance in competitions \citep{chen2016xgboost}. 
    We use the default option in XGBoost to measure feature importance with the average training loss gained when using a feature for splitting.
    \item LSTM with attention (often shortened as LSTM in this work). Attention is a commonly used technique in deep learning models for NLP \citep{bahdanau2014neural}.
    The intuition is to assign a weight to each token before aggregating into the final prediction (or decoding in machine translation).
    We use the dot product formulation in \citet{luong-etal-2015-effective}.
    The weight on each token has been commonly used to visualize the importance of each token.
    To compare with the previous bag-of-words models, we use the average weight of each type (unique token) in this work to measure feature importance.
    \item BERT. BERT represents an example architecture based on Transformers, which could show different behavior from LSTM-style recurrent networks \citep{devlin2018bert,vaswani2017attention,huggingface}.
    It also achieves state-of-the-art performance in many NLP tasks.
    Similar to LSTM with attention, we use the average attention values of 12 heads used by the first token at the final layer (the representation passed to fully connected layers) to measure feature importance for BERT.\footnote{We also tried to use the max of 12 heads and previous layers, and the average of the final layer is more similar to SVM ($\ell_2$) than the average of first layer. Results are in the supplementary material. 
    \citet{vig:2019} show that attention in BERT tends to be on first words, neighboring words, and even separators.
    The complex choices for 
    BERT further motivate our work to view feature importance as a subject of study.
    } 
    Since BERT uses a subword tokenizer, for each word, 
    we aggregate the attention on related subparts.
    BERT also requires special processing due to the length constraint; please refer to the supplementary material for details.
    As a result, we focus on presenting LSTM with attention in the main paper for ease of understanding.
\end{itemize}

\para{Methods ($h$).} For each model, in addition to the built-in feature importance that we described above, we consider the following two popular methods for extracting 
post-hoc feature importance (see the supplementary material for details of using the post-hoc methods).

\begin{itemize}[itemsep=0pt,topsep=0pt,leftmargin=*]
    \item LIME \citep{ribeiro2016should}. LIME generates post-hoc explanations by fitting a local sparse linear model to approximate model predictions.
    As a result, the explanations are sparse.
    \item SHAP \citep{lundberg2017unified}. SHAP unifies several interpretations of feature importance through Shapley values.
    The main intuition is to account the importance of a feature by examining the change in prediction outcomes for all the combinations of other features.
    \citet{lundberg2017unified} propose various approaches to approximate the computation for different classes of models (including gradient-based methods for deep models).
\end{itemize}

Note that feature importances obtained via all approaches are all local, because the top features are conditioned on an instance (i.e., words present in an instance) even for the built-in method for SVM and XGBoost.

\para{Comparing feature importance.} 
Given $\importance_{t}^{m,h}$ and $\importance_{t}^{m',h'}$, we use Jaccard similarity based on the top $k$ features with the greatest absolute feature importance, 
$\frac{|\operatorname{TopK}(\importance_{t}^{m,h}) \cap \operatorname{TopK}(\importance_{t}^{m',h'})|}{|\operatorname{TopK}(\importance_{t}^{m,h}) \cup \operatorname{TopK}(\importance_{t}^{m',h'})|}$,
as our main similarity metric for two reasons.
First, the most typical way to use feature importance for interpretation purposes is to show the most important features \citep{lai+tan:19,ribeiro2016should,horne2019rating}.
Second, some models and methods inherently generate sparse feature importance, so most feature importance values are 0.

It is useful to discuss the implication of similarity before we proceed.
On the one hand, it is possible that different models/methods identify the same set of important features (high similarity) and the performance difference in prediction is due to how different models weigh these important features.
If this were true, the choice of model/method would have mattered little for visualizing important features.
On the other hand, a low similarity poses challenges for choosing which model/method to use for displaying important features.
In that case, this work aims to develop an understanding of how the 
similarity varies depending on models and methods, instances, and features.
We leave it to future work for examining the impact on human interaction with feature importance.
Low similarity may enable model developers to understand the differences between models, but may lead to challenges for decision makers to get a consistent picture of what the model relies on.

\section{Experimental Setup and Hypotheses}
\label{sec:experiment}

Our goal is to 
characterize the similarities and differences between feature importances obtained with different methods and different models.
In this section, we first present our experimental setup and then formulate our hypotheses.

\para{Experimental setup.}
We consider the following three text classification tasks in this work.
We choose to focus on classification because classification is the most common scenario used for examining feature importance and the associated human interpretation \citep[e.g.,][]{jain2019}.

\begin{itemize}[itemsep=-5pt,topsep=0pt,leftmargin=*]
    \item Yelp \citep{yelp2019}. We set up a binary classification task to predict whether a review is positive (rating $\geq 4$) or negative (rating $\leq 2$).
    As the original dataset is huge, we subsample 12,000 reviews for this work.
    \item SST \citep{socher2013recursive}. It is a sentence-level sentiment classification task and represents a common benchmark. 
    We only consider the binary setup here.
    \item Deception detection \citep{ott2013negative,ott2011finding}. This dataset was created by extracting genuine reviews from TripAdvisor and collecting deceptive reviews using Turkers. It is relatively small with 1,200 reviews and represents a distinct task from sentiment classification.
\end{itemize}

For all the tasks, we use 20\% of the dataset as the test set.
For SVM and XGBoost, we use cross validation on the other 80\% to tune hyperparameters.
For LSTM with attention and BERT, we use 10\% of the dataset as a validation set, and choose the best hyperparameters based on the validation performance.
We use spaCy to tokenize and obtain part-of-speech tags for all the datasets \citep{honnibal2017spacy}.
\tableref{tb:accuracy} shows the accuracy on the test set and our results are comparable to prior work.
Not surprisingly, BERT achieves the best performance in all three tasks.
For important features, we use $k\leq 10$ for Yelp and deception detection, and $k\leq 5$ for SST as it is a sentence-level task.
See supplementary materials for details of preprocessing, learning, and dataset statistics.

\begin{table}[t]
\small
\centering
\begin{tabular}{lrrr}
\toprule
Model &  Yelp & SST & Deception \\
\midrule
SVM ($\ell_2$) & 92.3 & 80.8 & 86.3 \\
SVM ($\ell_1$) & 91.5 & 79.2 & 84.4 \\
XGBoost & 88.8 & 75.9 & 83.4 \\
LSTM w/ attention & 93.9 & 82.6 & 88.4 \\
BERT & 95.5 & 92.2 & 90.9 \\
\bottomrule
\end{tabular}
\caption{Accuracy on the test set.}
\label{tb:accuracy}
\end{table}

\para{Hypotheses.}
We aim to examine the following three research questions in this work:
1) How similar are important features between models and methods?
2) What factors relate to the heterogeneity across instances?
3) What words tend to be chosen as important features?

\empara{Overall similarity.}
Here we focus on discussing comparative hypotheses, but we would like to note that it is important to understand to what extent important features are similar across models (i.e., the value of similarity score).
First, as deep learning models and XGBoost are nonlinear, we hypothesize that built-in feature importance is more similar between SVM ($\ell_1$) and SVM ($\ell_2$) than other model pairs (\hypothesis{H1a}).
Second, LIME generates more similar important features to SHAP than to built-in feature importance because both LIME and SHAP make additive assumptions, while built-in feature importance is based on drastically different models (\hypothesis{H1b}).
It also follows that post-hoc explanations of different models show higher similarity than built-in explanations across models.
Third, the similarity with small $k$ is higher (\hypothesis{H1c}) because hopefully, all models and methods agree what the most important features are.

\empara{Heterogeneity between instances.}
Given a pair of (model, method) combinations, our second question is concerned with 
how
instance-level properties 
affect the similarity in important features between different combinations.
We hypothesize that 1) when two models agree on the predicted label, the similarity between important features is greater (\hypothesis{H2a});
2) longer instances are less likely to share similar important features (\hypothesis{H2b}).
3) instances with higher type-token ratio,\footnote{Type-token ratio is defined as the number of unique tokens divided by the number of tokens.} which might be more complex, are less likely to share similar important features (\hypothesis{H2c}).

\empara{Distribution of important features.}
Finally, we examine what words tend to be chosen as important features.
This question certainly depends on the nature of the task, but
we would like to understand how consistent different models and methods are.
We hypothesize that
1) deep learning models generate more diverse important features (\hypothesis{H3a});
2) adjectives are more important in sentiment classification,
while pronouns are more important in deception detection as shown in prior work (\hypothesis{H3b}).

\section{Similarity between Instance-level Feature Importance}
\label{sec:similarity}

\begin{figure}[t]
\centering
\includegraphics[width=0.3\textwidth]{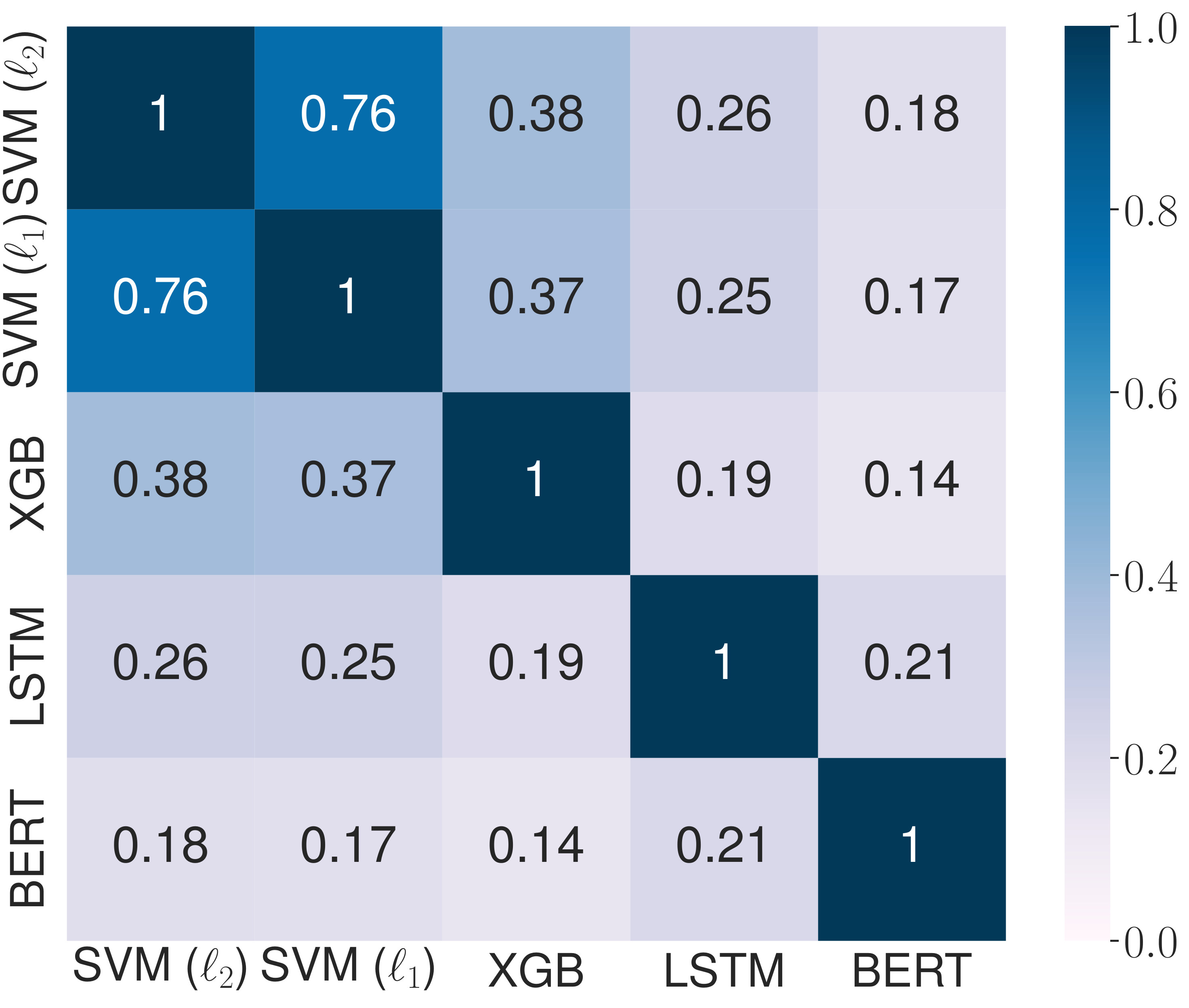}
\caption{Jaccard similarity between the top 10 features of different models based on built-in feature importance on Yelp. The similarity is the greatest between SVM ($\ell_2$) and SVM ($\ell_1$), while LSTM with attention and BERT pay attention to quite different features from other models.
}
\label{fig:yelp_heatmap}
\end{figure}

\begin{figure*}[t]
\centering
Similarity comparison between models using the built-in method \\
\begin{subfigure}[t]{0.43\textwidth}
  \centering
  \includegraphics[width=\textwidth]{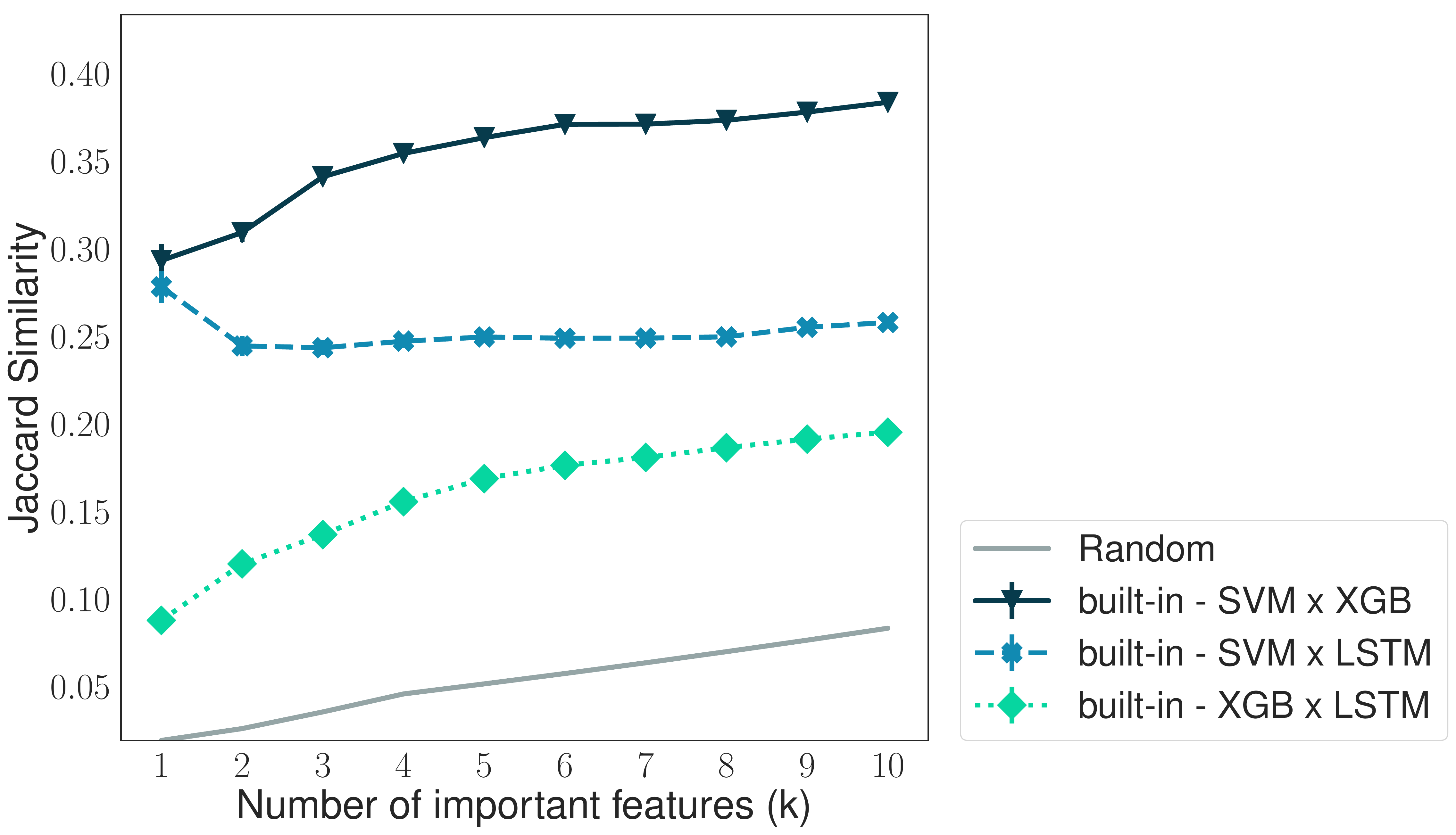}
  \caption{Yelp}
  \label{fig:yelp_methods}
\end{subfigure}
\hfill
\begin{subfigure}[t]{0.27\textwidth}
  \centering
  \includegraphics[trim=0 0 6.5in 0,clip,width=\textwidth]{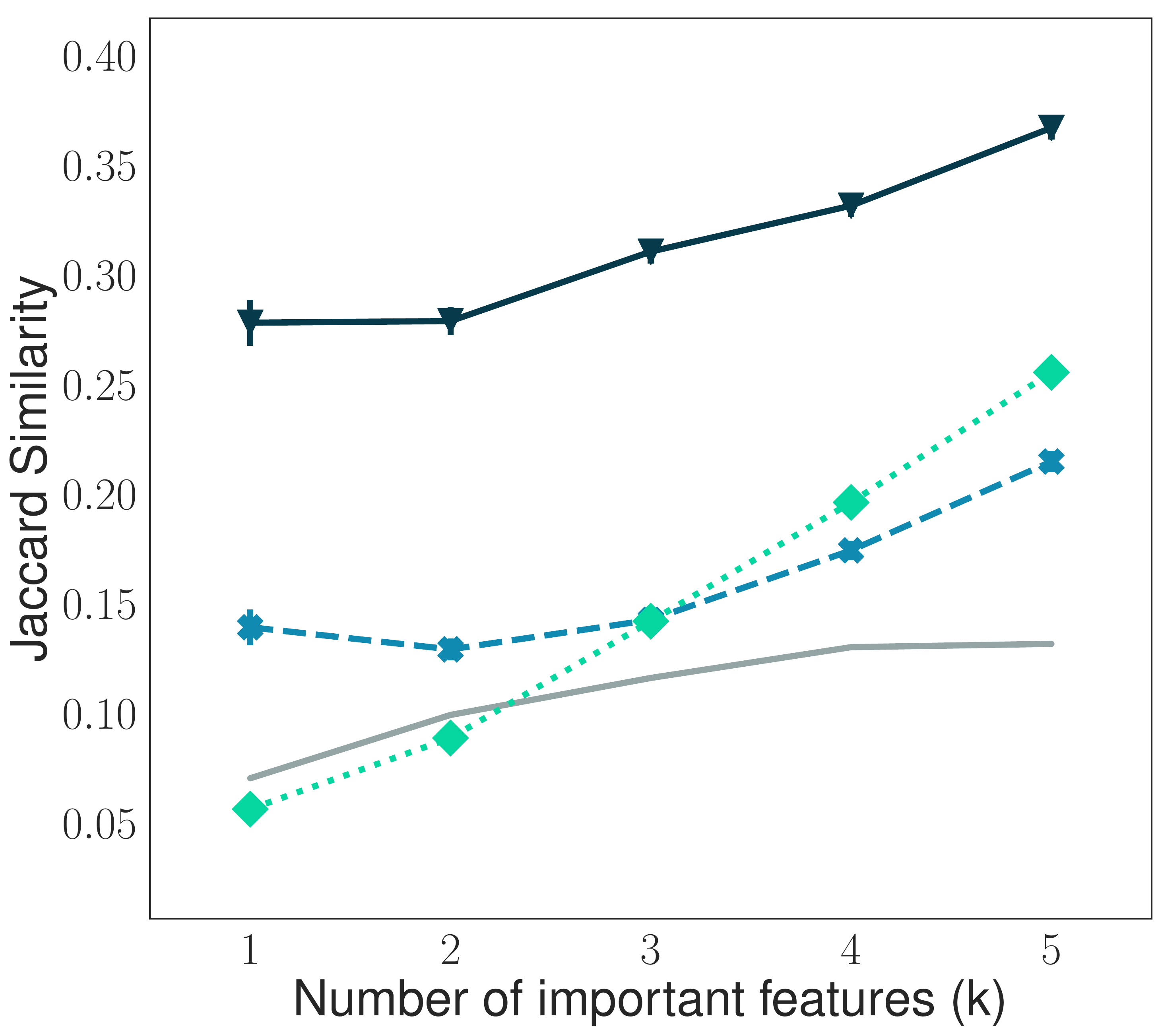}
  \caption{SST}
  \label{fig:sst_methods}
\end{subfigure}
\hfill
\begin{subfigure}[t]{0.265\textwidth}
  \centering
  \includegraphics[trim=0 0 6.5in 0,clip,width=\textwidth]{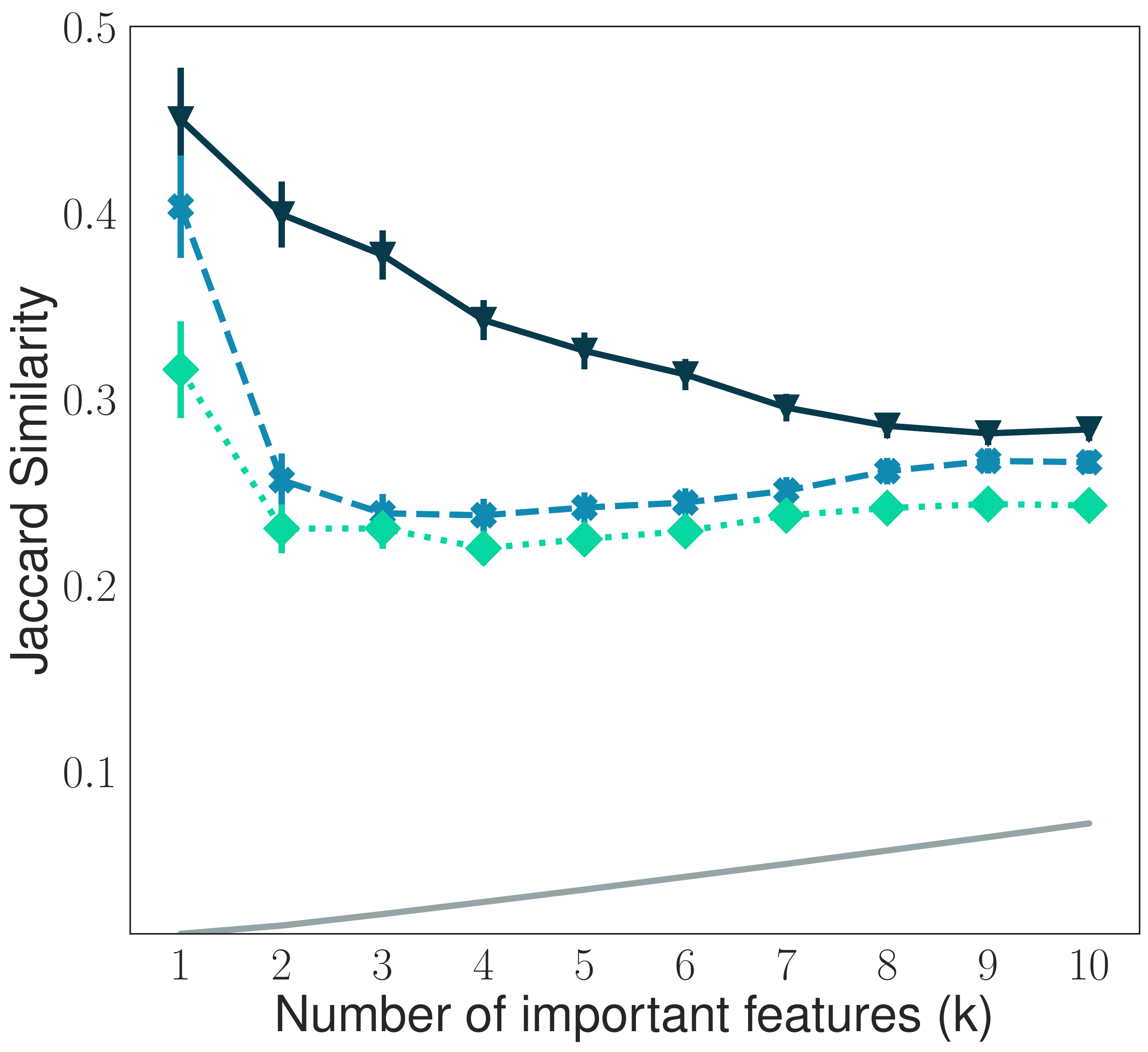}
  \caption{Deception}
  \label{fig:deception_methods}
\end{subfigure}\\\bigskip
Comparison between the built-in method and post-hoc methods \\
\begin{subfigure}[t]{0.44\textwidth}
  \centering
  \includegraphics[width=\textwidth]{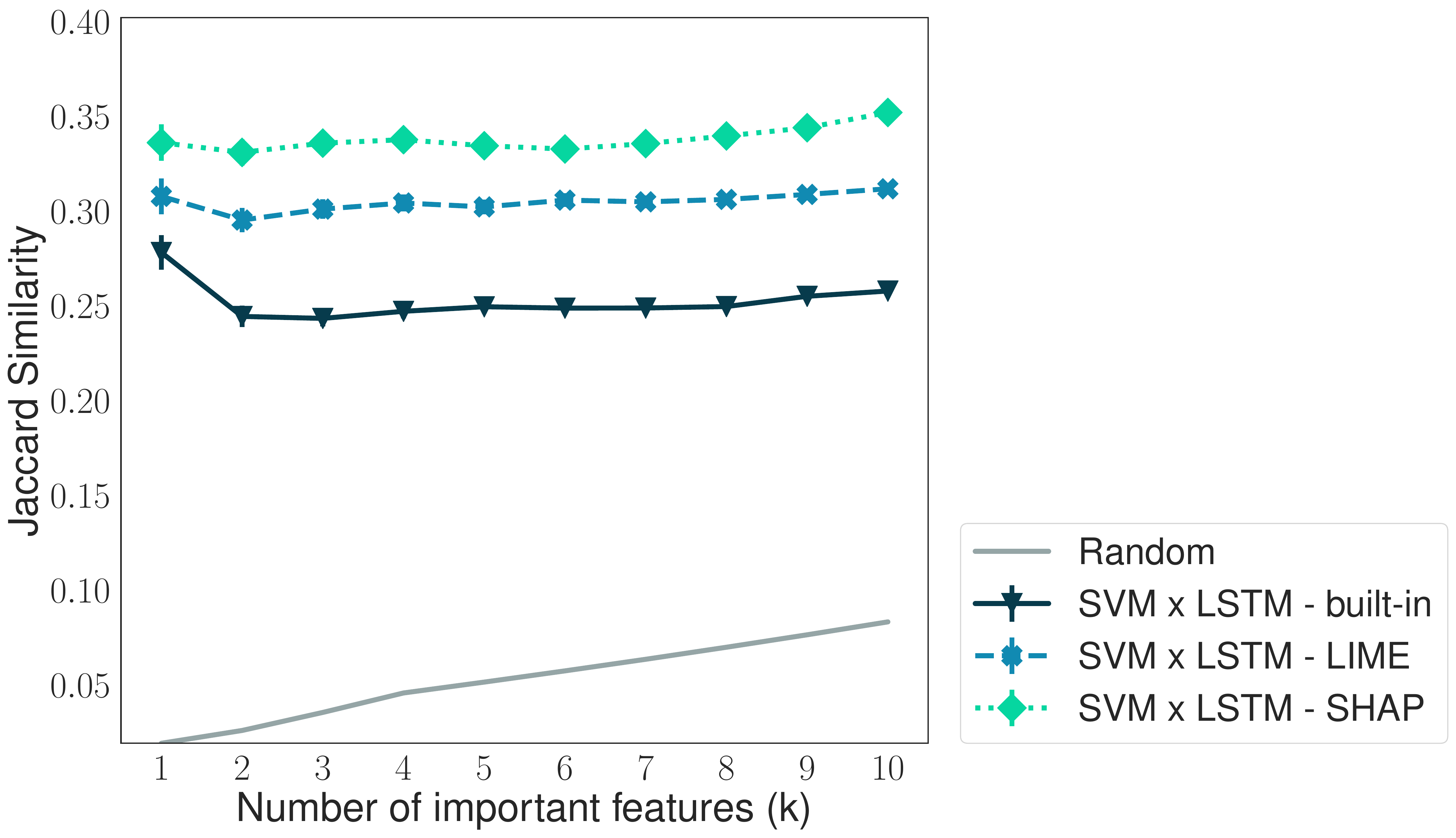}
  \caption{Yelp}
  \label{fig:yelp_methods}
\end{subfigure}
\hfill
\begin{subfigure}[t]{0.27\textwidth}
  \centering
  \includegraphics[trim=0 0 6.5in 0,clip,width=\textwidth]{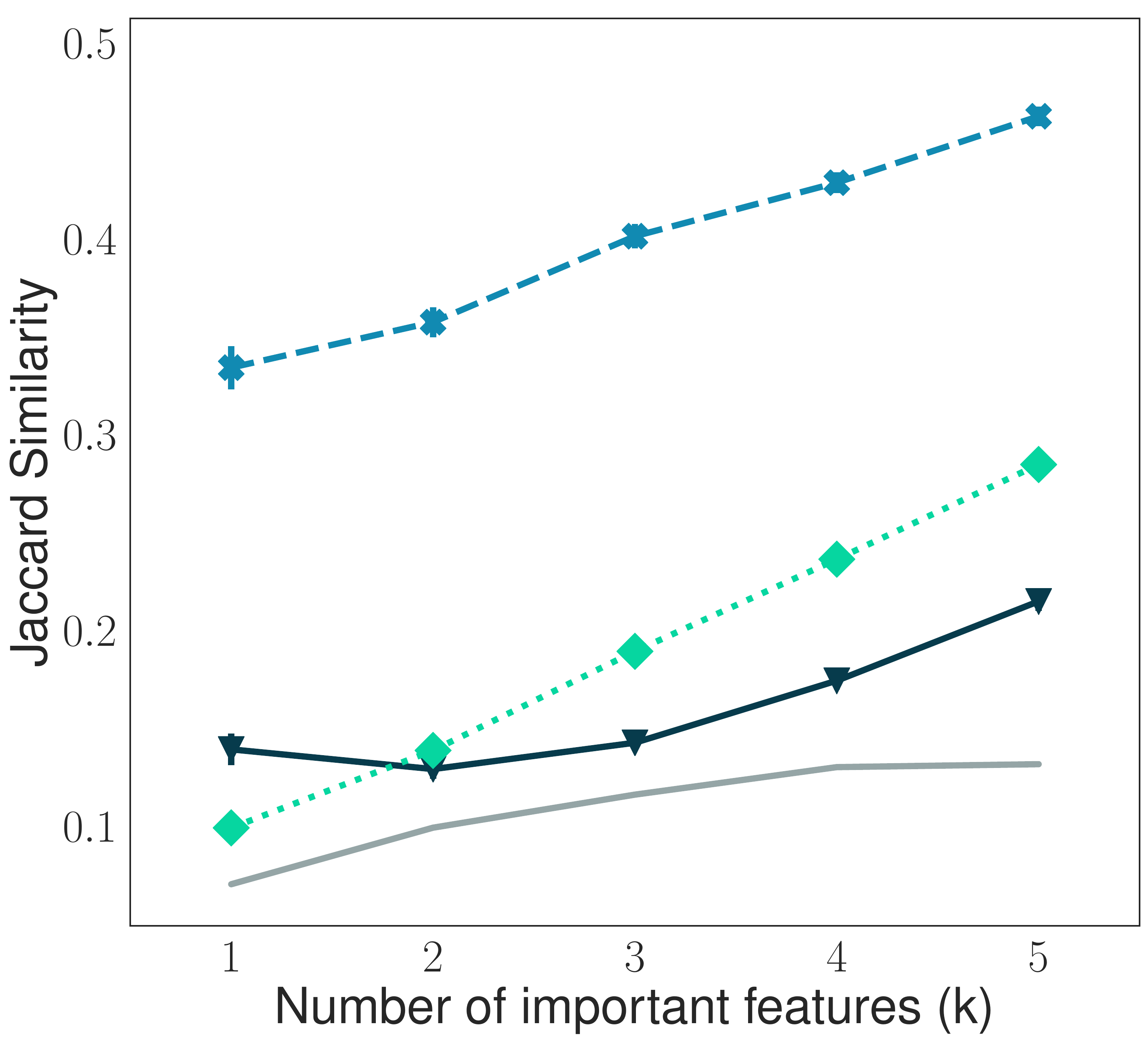}
  \caption{SST}
  \label{fig:sst_methods}
\end{subfigure}
\hfill
\begin{subfigure}[t]{0.27\textwidth}
  \centering
  \includegraphics[trim=0 0 6.5in 0,clip,width=\textwidth]{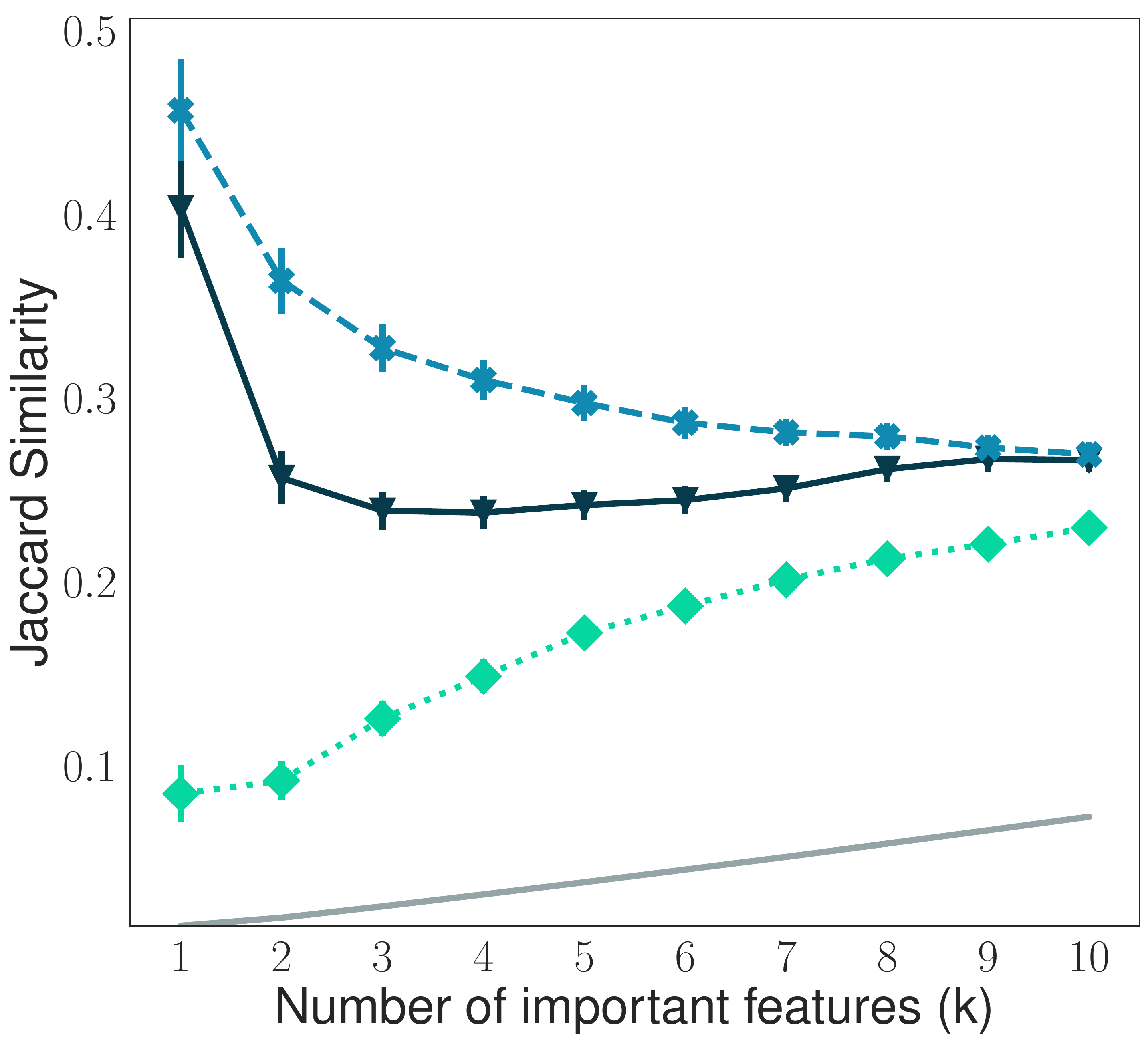}
  \caption{Deception}
  \label{fig:deception_methods}
\end{subfigure}
\caption{Similarity comparison between models with the same method.
$x$-axis represents the number of important features that we consider, while $y$-axis shows the Jaccard similarity.
{\em Error bars represent standard error throughout the paper.}
The top row compares three pairs of models using the built-in method, while the second row compares three methods on SVM and LSTM with attention (LSTM in figure legends always refers to LSTM with attention in this work).
The random line is derived using the average similarity between two random samples of $k$ features from 100 draws.
}
\label{fig:similarity}
\end{figure*}

We start by examining the overall similarity between different models using different methods.
In a nutshell, we compute the average Jaccard similarity of top $k$ features for each pair of $(m, h)$ and $(m', h')$.
To facilitate effective comparisons, we first fix the method and compare the similarity of different models, and then fix the model and compare the similarity of different methods.
\figref{fig:yelp_heatmap} shows the similarity between different models using the built-in feature importance for the top 10 features in Yelp ($k=10$).
Consistent with {\bf H1a}, SVM ($\ell_2$) and SVM ($\ell_1$) are very similar to each other, and LSTM with attention and BERT clearly lead to quite different top 10 features from the other models.
As the number of important features ($k$) can be useful for evaluating the overall trend, we thus focus on line plots as in \figref{fig:similarity} in the rest of the paper.
This heatmap visualization represents a snapshot for 
 $k=10$ using the built-in method.
Also, 
we only include SVM ($\ell_2$) in the main paper for ease of visualization and sometimes refer to it in the rest of the paper as SVM.

\begin{figure*}[t!]
\centering
\begin{subfigure}[t]{0.43\textwidth}
  \centering
  \includegraphics[width=\textwidth]{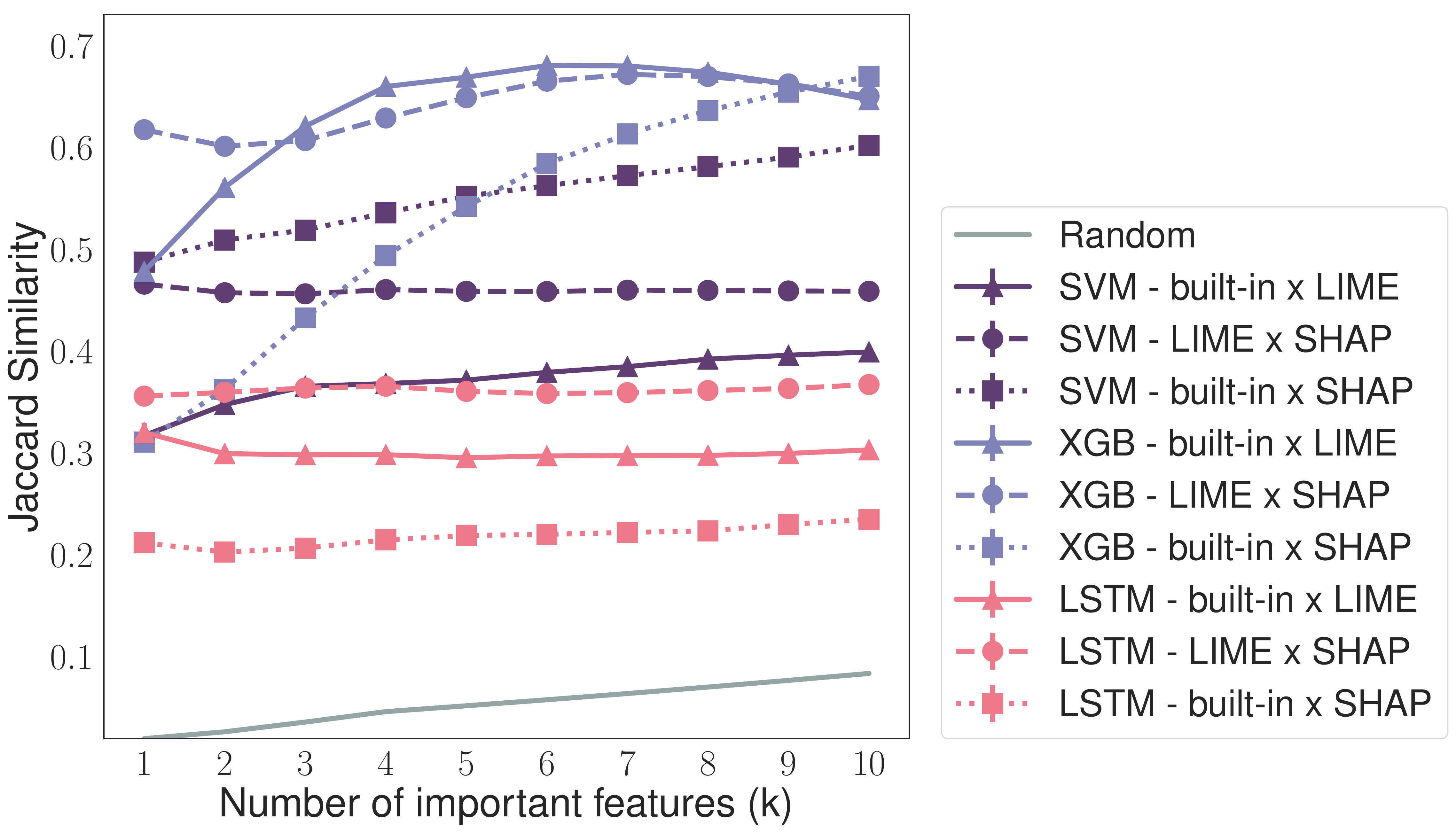}
  \caption{Yelp}
  \label{fig:yelp_models}
\end{subfigure}
\hfill
\begin{subfigure}[t]{0.27\textwidth}
  \centering
  \includegraphics[trim=0 0 6.5in 0,clip,width=\textwidth]{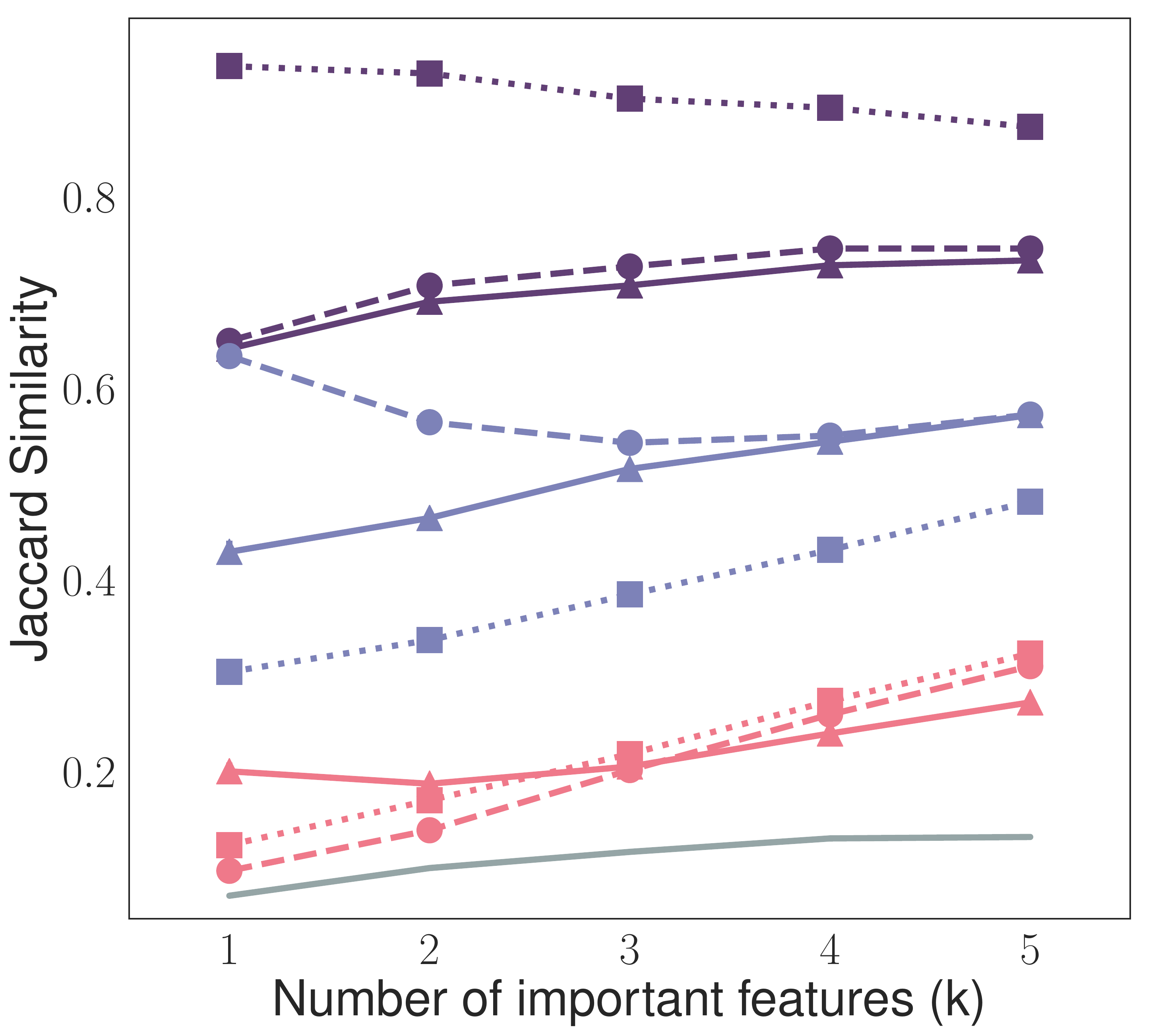}
  \caption{SST}
  \label{fig:sst_models}
\end{subfigure}
\hfill
\begin{subfigure}[t]{0.27\textwidth}
  \centering
  \includegraphics[trim=0 0 6.5in 0,clip,width=\textwidth]{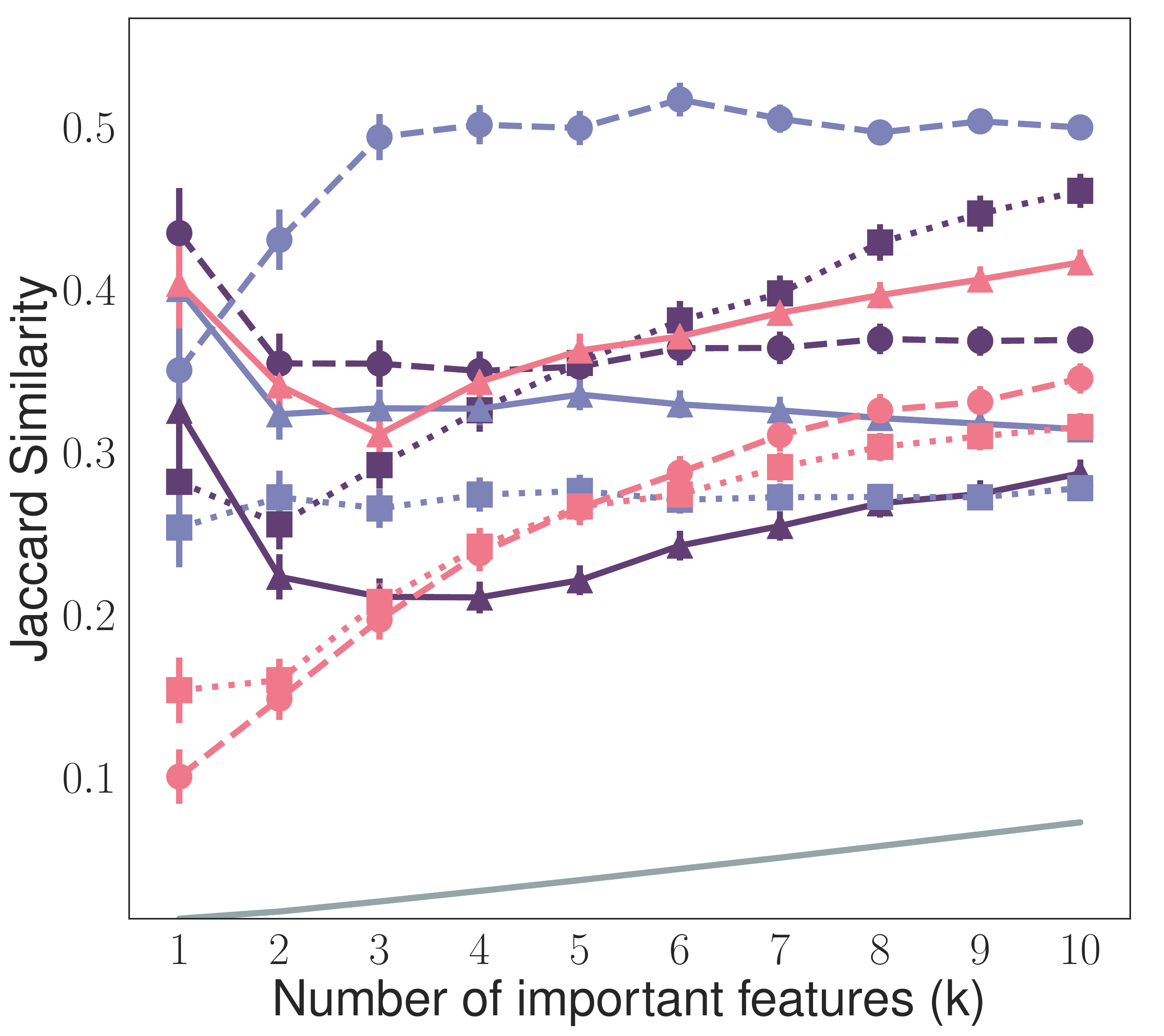}
  \caption{Deception}
  \label{fig:deception_models}
\end{subfigure}
\caption{Similarity comparison between methods using the same model.
The similarity between different methods based on LSTM with attention is generally lower than other methods.
Similar results hold for BERT (see the supplementary material).
}
\label{fig:similarity_models}
\end{figure*}

\begin{figure*}[t]
\centering
\begin{subfigure}[t]{0.43\textwidth}
  \centering
  \includegraphics[width=\textwidth]{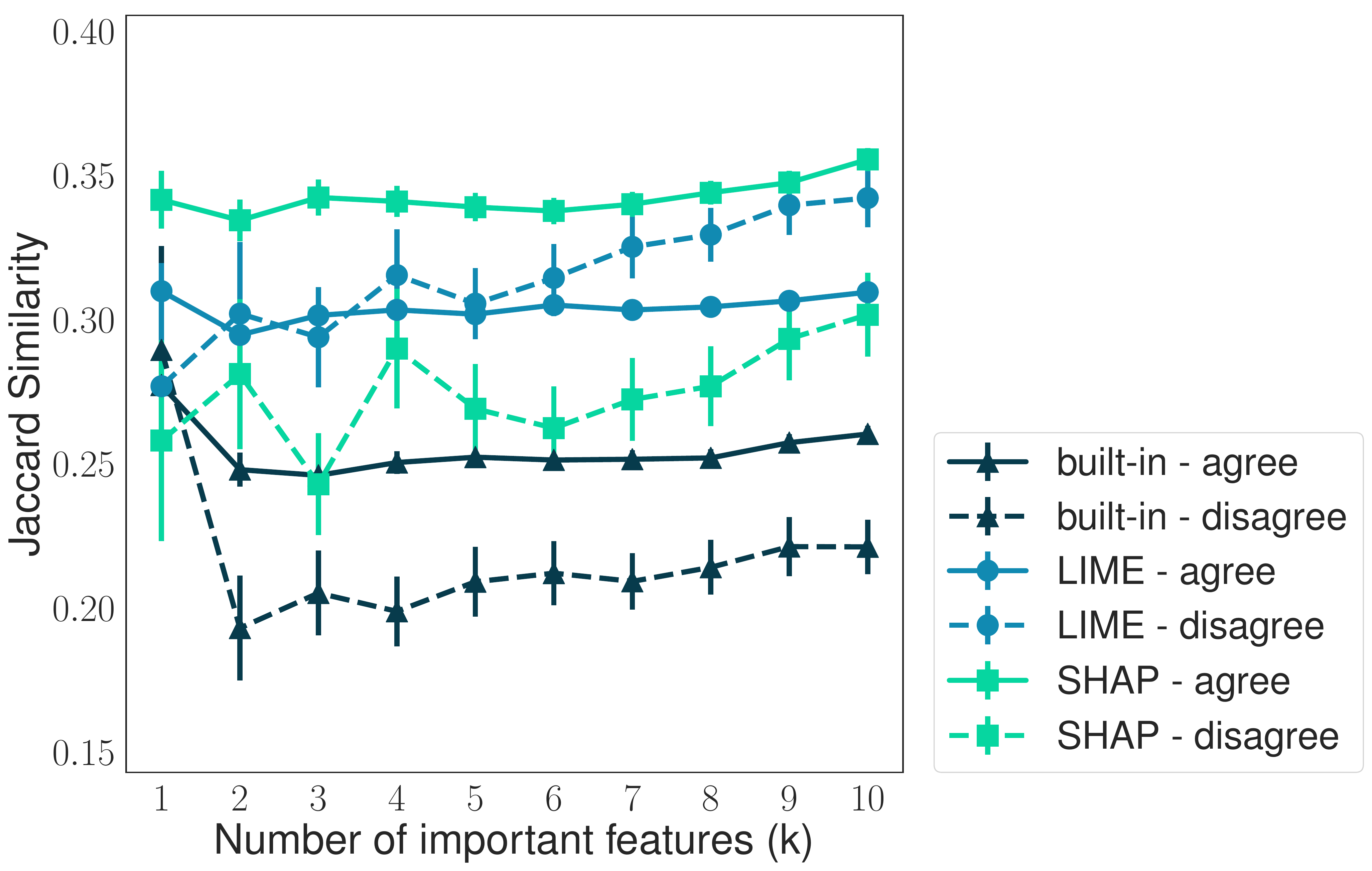}
  \caption{Yelp}
  \label{fig:yelp_svm_lstm_att_pred}
\end{subfigure}
\hfill
\begin{subfigure}[t]{0.27\textwidth}
  \centering
  \includegraphics[trim=0 0 5.5in 0,clip,width=\textwidth]{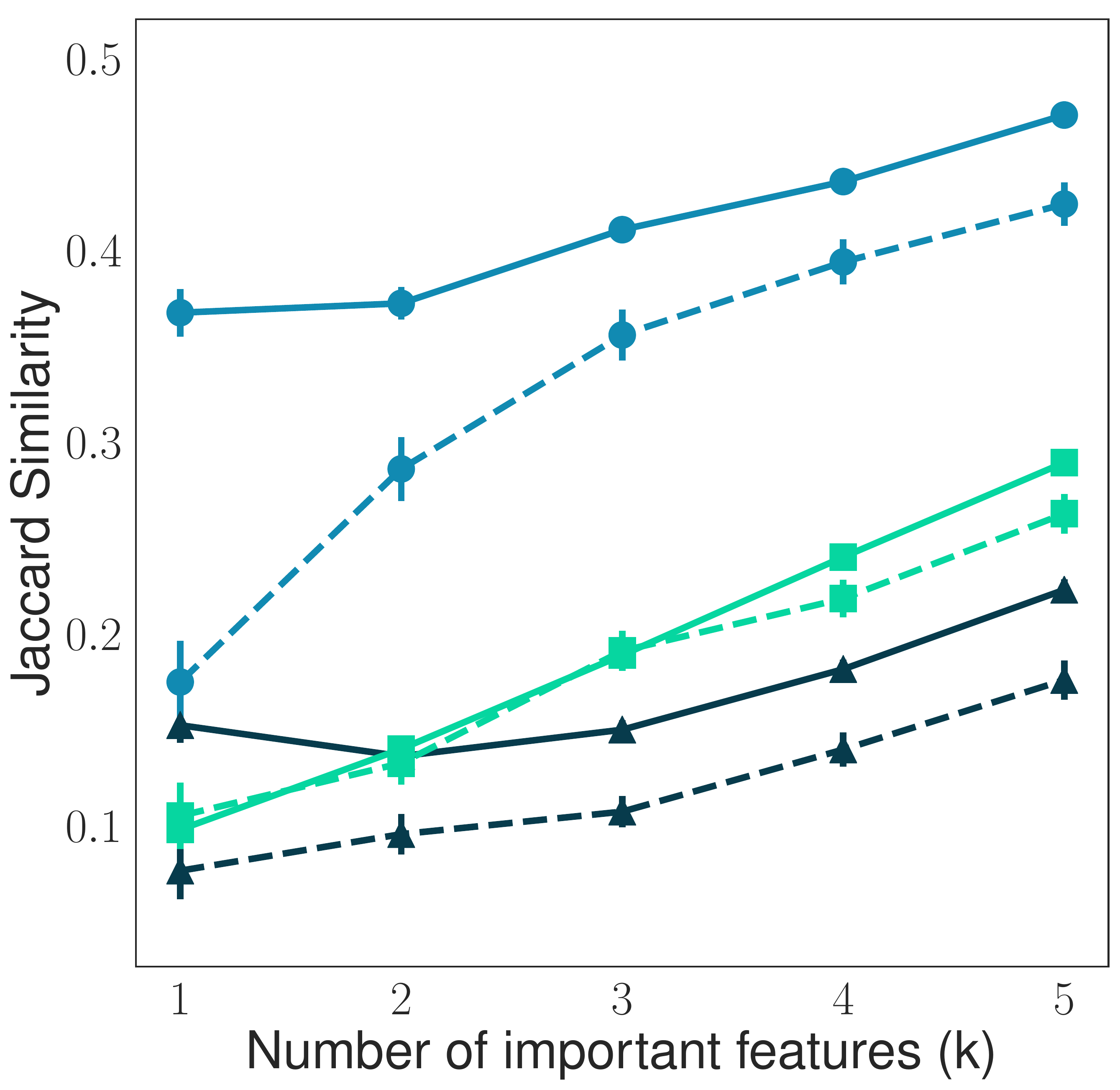}
  \caption{SST}
  \label{fig:sst_svm_lstm_att_pred}
\end{subfigure}
\hfill
\begin{subfigure}[t]{0.27\textwidth}
  \centering
  \includegraphics[trim=0 0 5.5in 0,clip,width=\textwidth]{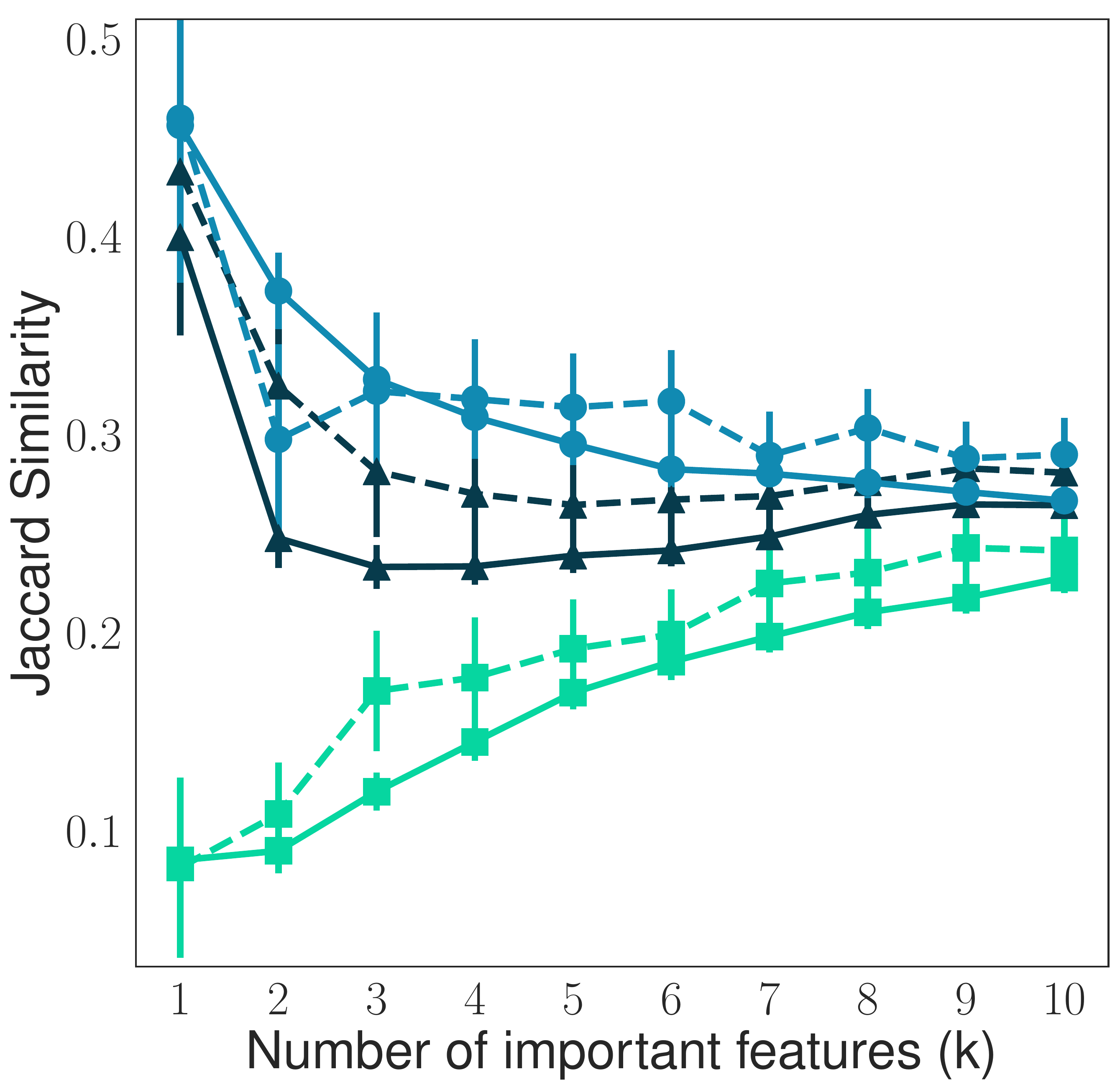}
  \caption{Deception}
  \label{fig:deception_svm_lstm_att_pred}
\end{subfigure}
\caption{
Similarity between SVM ($\ell_2$) and LSTM with attention with different methods grouped by whether these two models agree on the predicted label.
The similarity 
is not always greater when they agree on the predicted labels than when they disagree.
}
\label{fig:prediction}
\end{figure*}

\begin{figure*}[t!]
\centering
\begin{subfigure}[t]{0.425\textwidth}
  \centering
  \includegraphics[width=\textwidth]{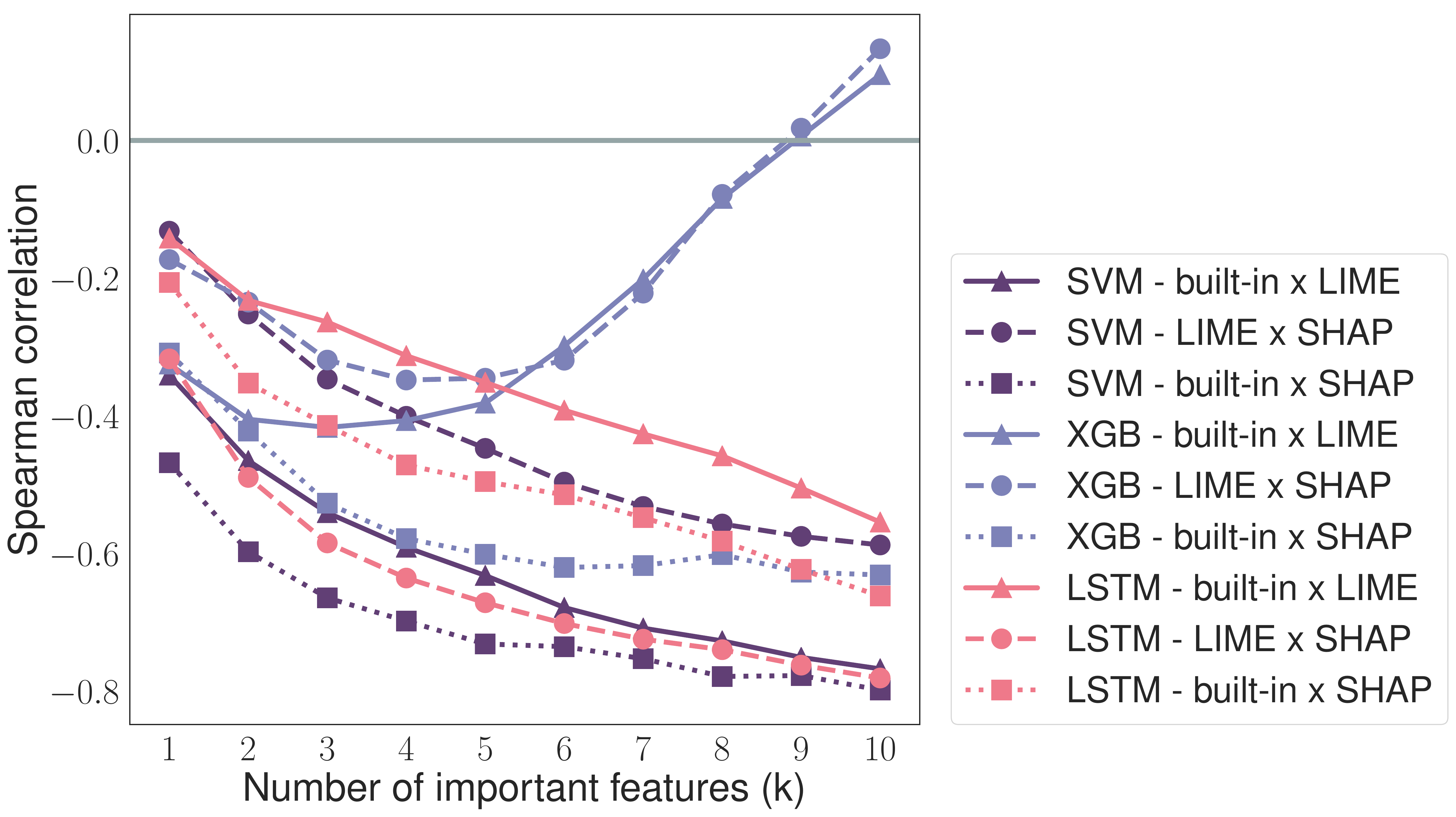}
  \caption{Yelp}
  \label{fig:yelp_models_length}
\end{subfigure}
\hfill
\begin{subfigure}[t]{0.28\textwidth}
  \centering
  \includegraphics[trim=0 0 6.5in 0,clip,width=\textwidth]{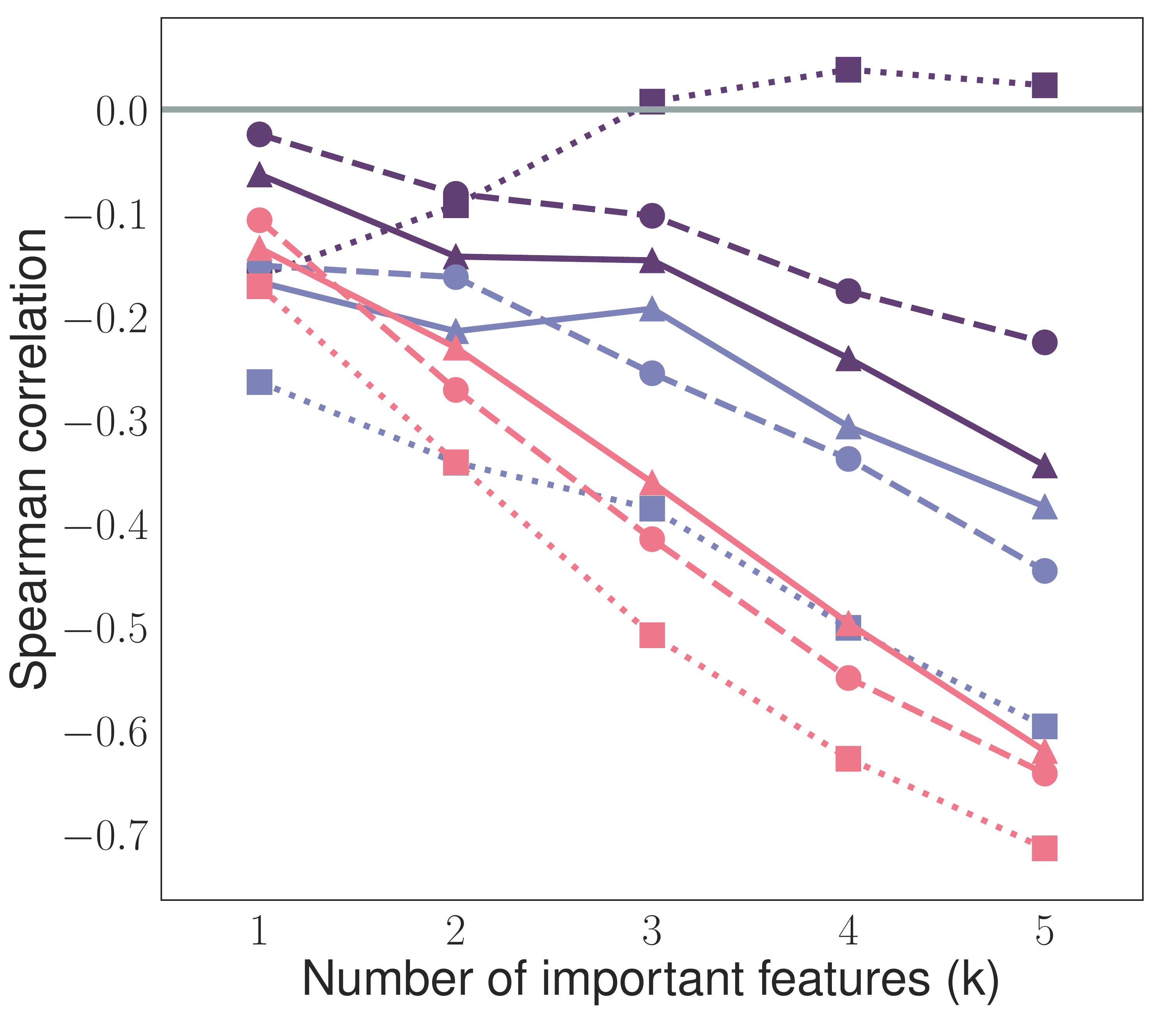}
  \caption{SST}
  \label{fig:sst_models_length}
\end{subfigure}
\hfill
\begin{subfigure}[t]{0.28\textwidth}
  \centering
  \includegraphics[trim=0 0 6.5in 0,clip,width=\textwidth]{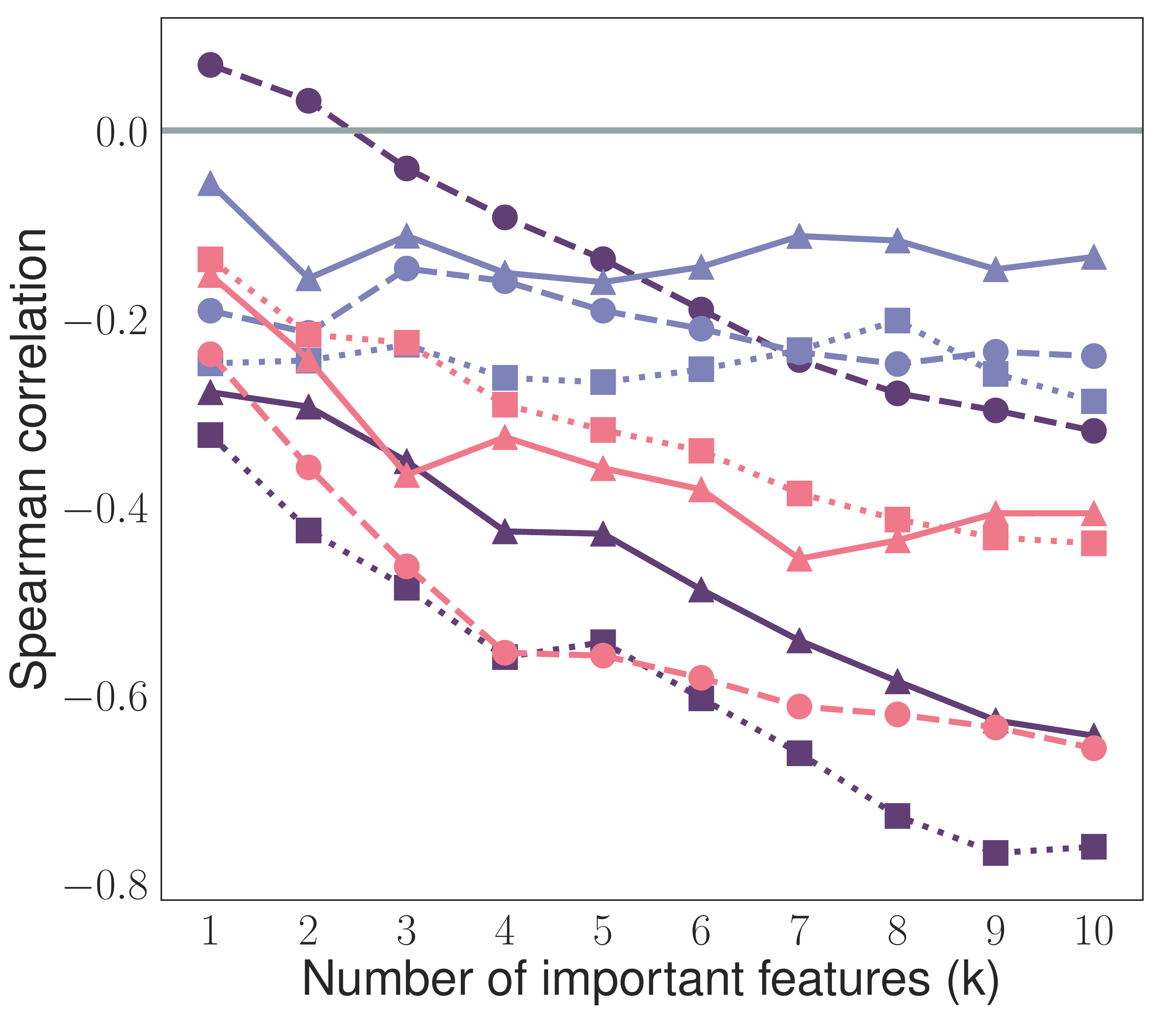}
  \caption{Deception}
  \label{fig:deception_models_length}
\end{subfigure}
\caption{In most cases, the similarity between feature importance is negatively correlated with length. 
Here we only show the comparison between different methods based on the same model.
Similar results hold for comparison between different models using the same method.
For ease of comparison, the gray line 
marks the value $0$. Generally as $k$ grows, relationship becomes even more negatively correlated. 
}
\label{fig:length}
\end{figure*}

\para{No matter which method we use,
important features from SVM and XGBoost are more similar with each other, than with 
deep learning models (\figref{fig:similarity}).}
First, we compare the similarity of feature importance between different models using the same method.
Using the built-in method (first row in \figref{fig:similarity}), the solid line (SVM x XGBoost) is always above the other lines, usually by a significant margin, suggesting that deep learning models such as LSTM with attention are less similar to traditional models.
In fact, the similarity between XGBoost and LSTM with attention is lower than random samples for $k=1,2$ in SST.
Similar results also hold for BERT (see supplementary materials).
Another interesting observation is that post-hoc methods
tend to generate greater similarity than built-in methods, especially for LIME 
(the dashed line (LIME) is always above the solid line (built-in) in the second row of \figref{fig:similarity}).
This is likely because LIME only depends on the model behavior (i.e., what the model predicts) and does not account for how the model works.

\begin{figure*}[t]
\centering
\begin{subfigure}[t]{0.4\textwidth}
  \centering
  \includegraphics[width=\textwidth]{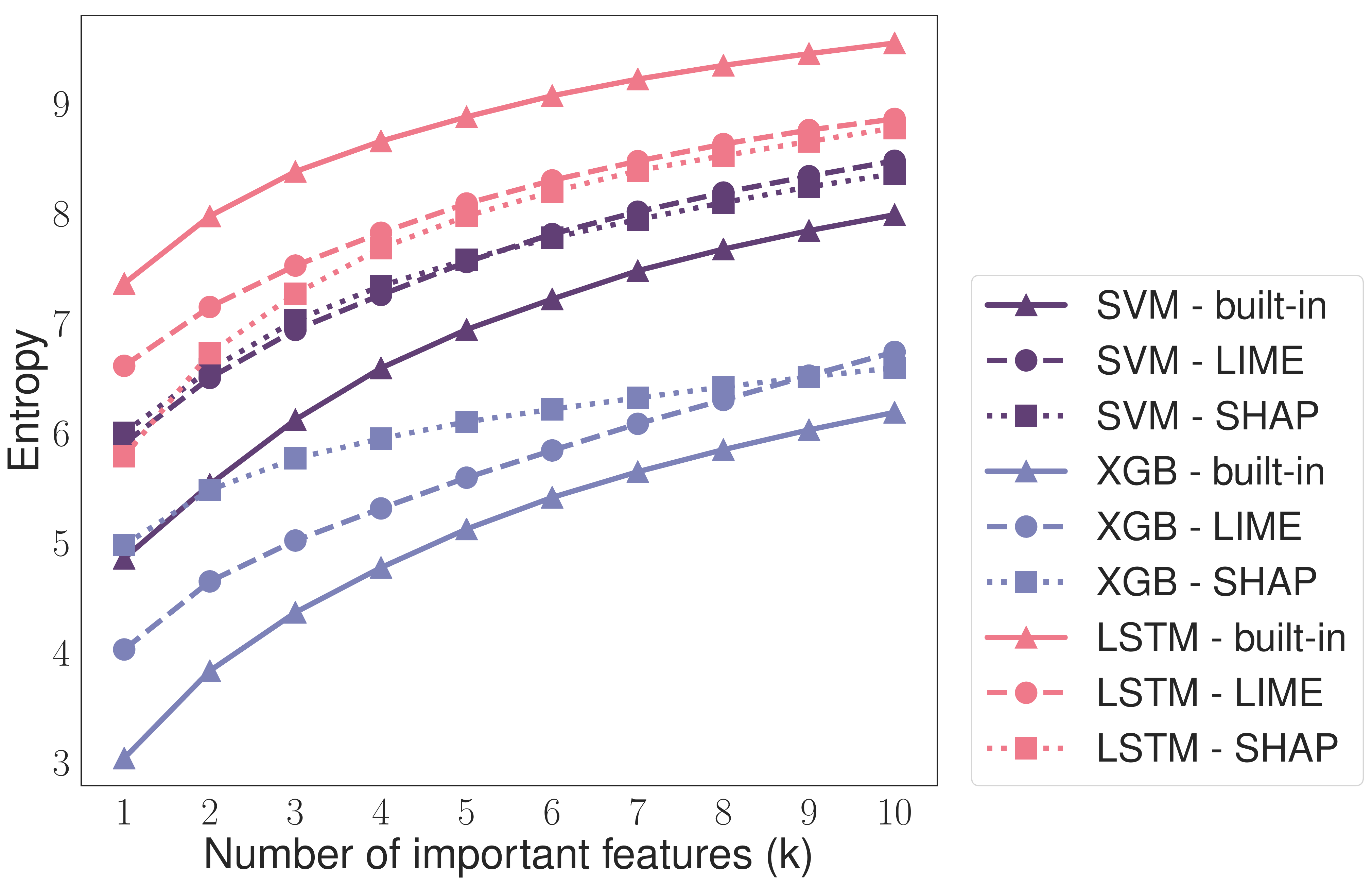}
  \caption{Yelp}
  \label{fig:yelp_models_entropy}
\end{subfigure}
\hfill
\begin{subfigure}[t]{0.275\textwidth}
  \centering
  \includegraphics[trim=0 0 5in 0,clip,width=\textwidth]{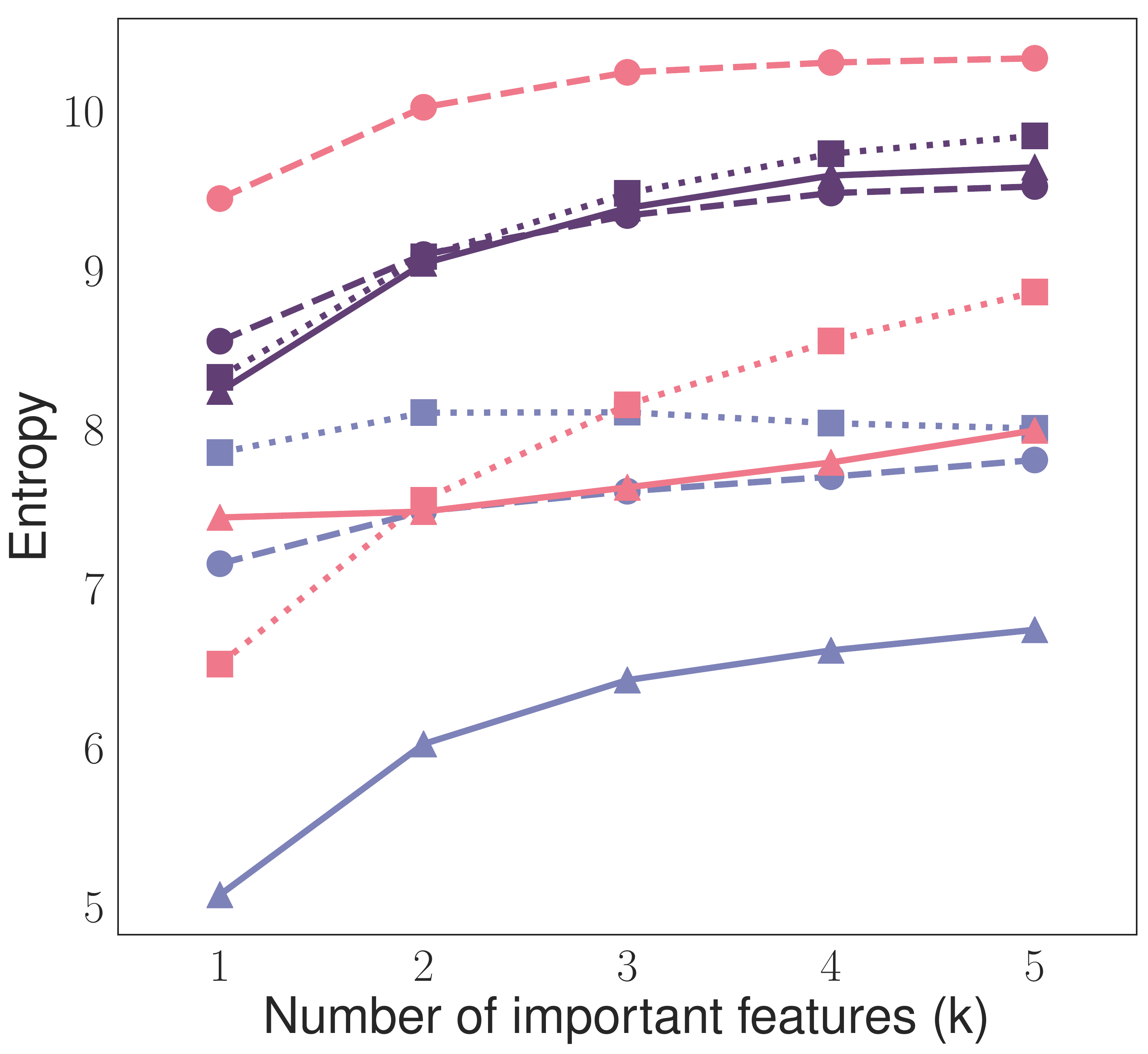}
  \caption{SST}
  \label{fig:sst_models_entropy}
\end{subfigure}
\hfill
\begin{subfigure}[t]{0.27\textwidth}
  \centering
  \includegraphics[trim=0 0 5in 0,clip,width=\textwidth]{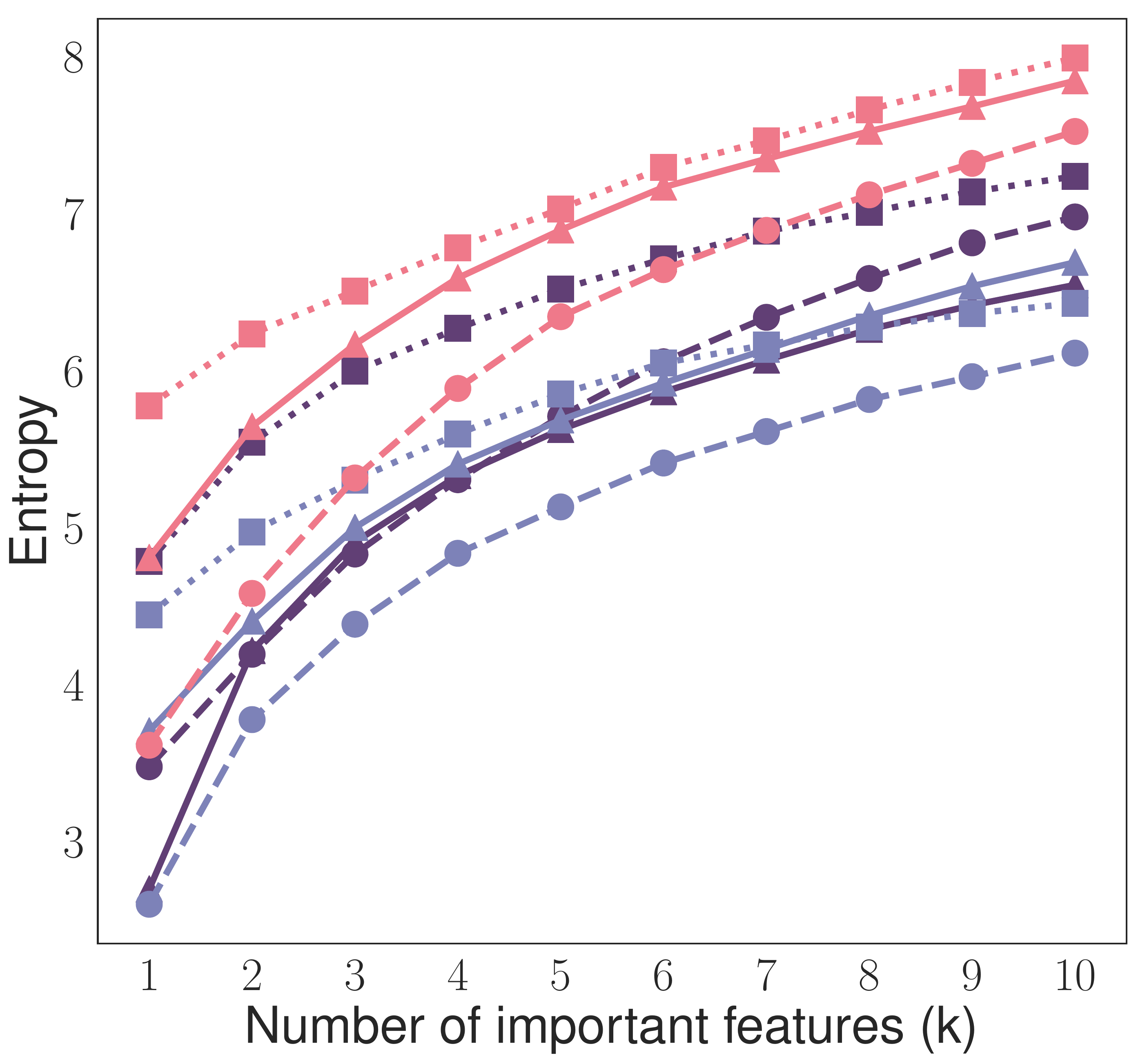}
  \caption{Deception}
  \label{fig:deception_models_entropy}
\end{subfigure}
\caption{The entropy of important features. LSTM with attention generates more diverse important features than SVM and XGBoost.
}
\label{fig:entropy}
\end{figure*}

\begin{figure*}[t]
\centering
\begin{subfigure}[t]{0.38\textwidth}
  \centering
  \includegraphics[width=\textwidth]{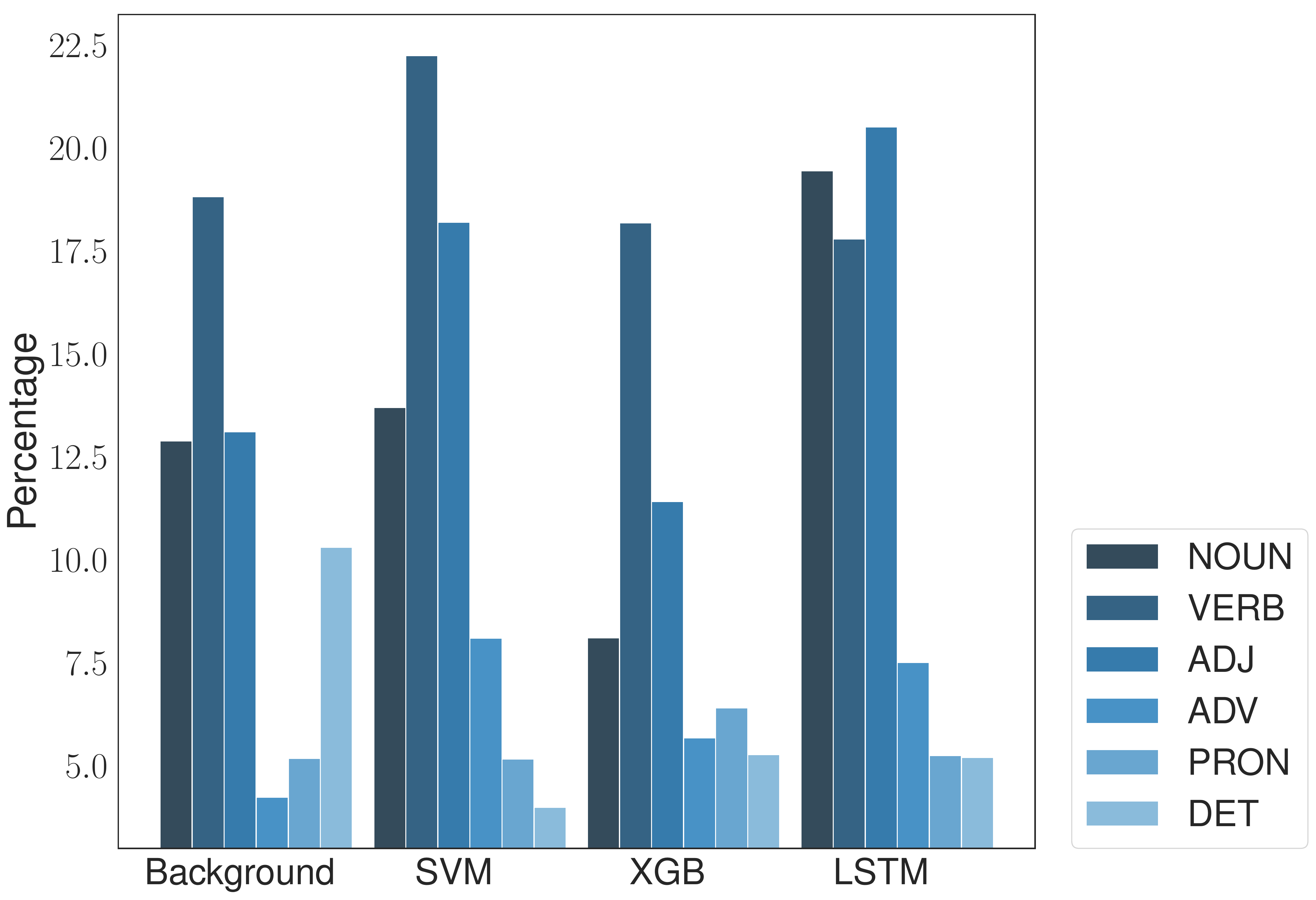}
  \caption{Yelp}
  \label{fig:yelp_pos}
\end{subfigure}
\hfill
\begin{subfigure}[t]{0.3\textwidth}
  \centering
  \includegraphics[trim=0 0 3.5in 0,clip,width=\textwidth]{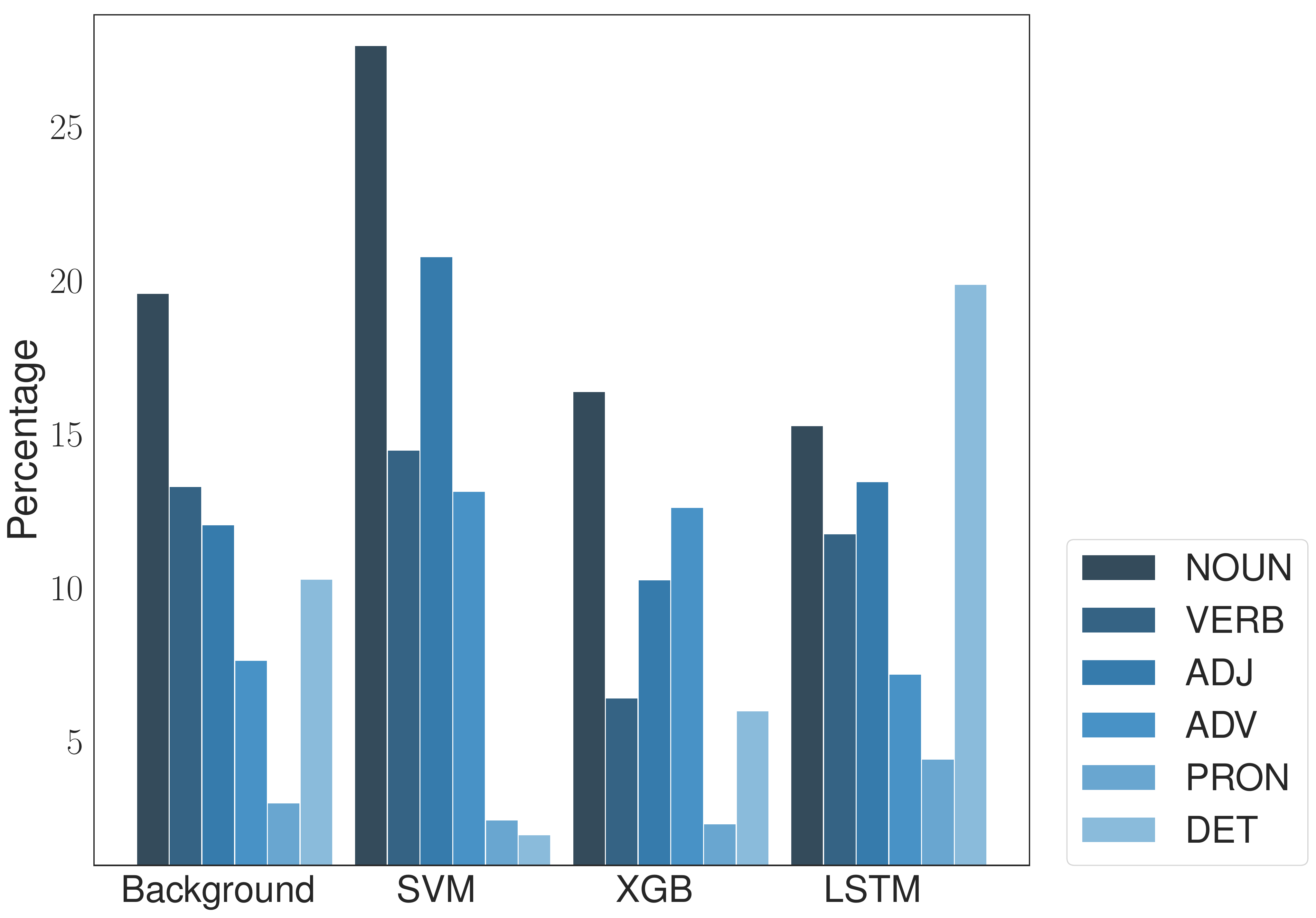}
  \caption{SST}
  \label{fig:sst_pos}
\end{subfigure}
\hfill
\begin{subfigure}[t]{0.3\textwidth}
  \centering
  \includegraphics[trim=0 0 3.5in 0,clip,width=\textwidth]{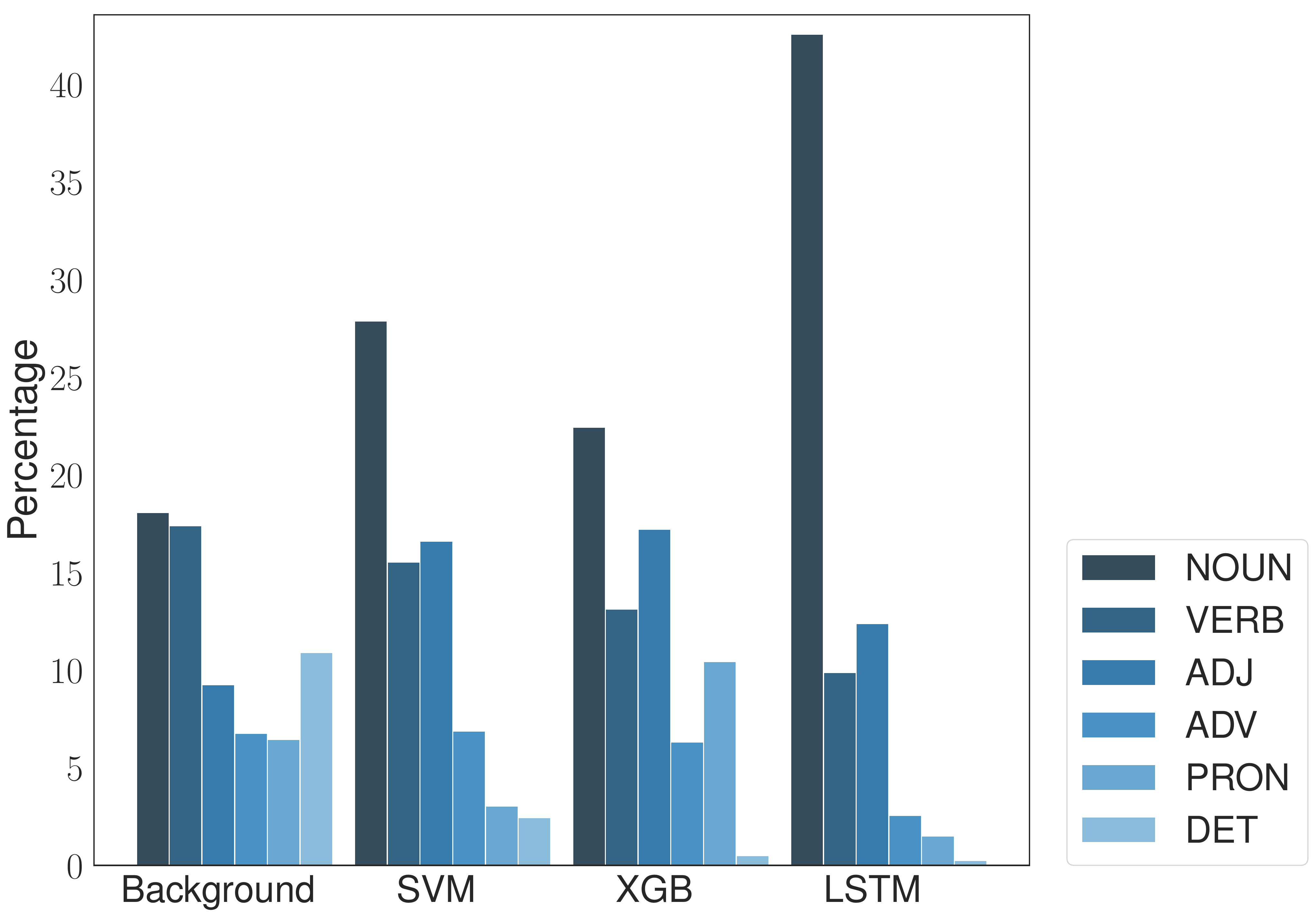}
  \caption{Deception}
  \label{fig:deception_pos}
\end{subfigure}
\caption{Part-of-speech tag distribution with the built-in method. ``Background'' shows the distribution of all words in the test set.
LSTM with attention puts a strong emphasis on nouns in deception detection, but is not necessarily more different from the background than other models.
}
\label{fig:pos}
\end{figure*}

\begin{figure*}[t]
\centering
\begin{subfigure}[t]{0.44\textwidth}
  \centering
  \includegraphics[width=\textwidth]{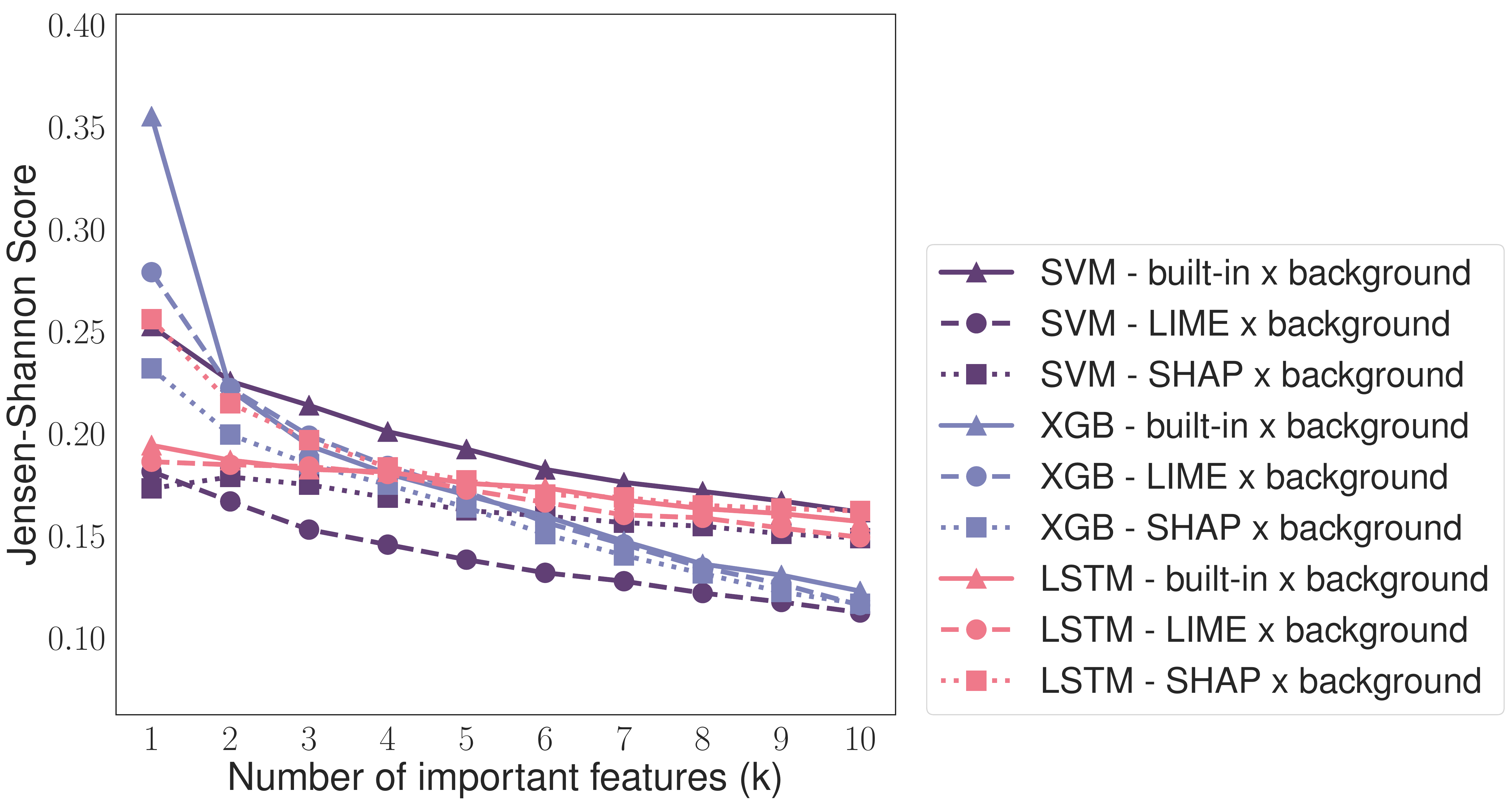}
  \caption{Yelp}
  \label{fig:yelp_models_jensenshannon_pos}
\end{subfigure}
\hfill
\begin{subfigure}[t]{0.275\textwidth}
  \centering
  \includegraphics[trim=0 0 7.7in 0,clip,width=\textwidth]{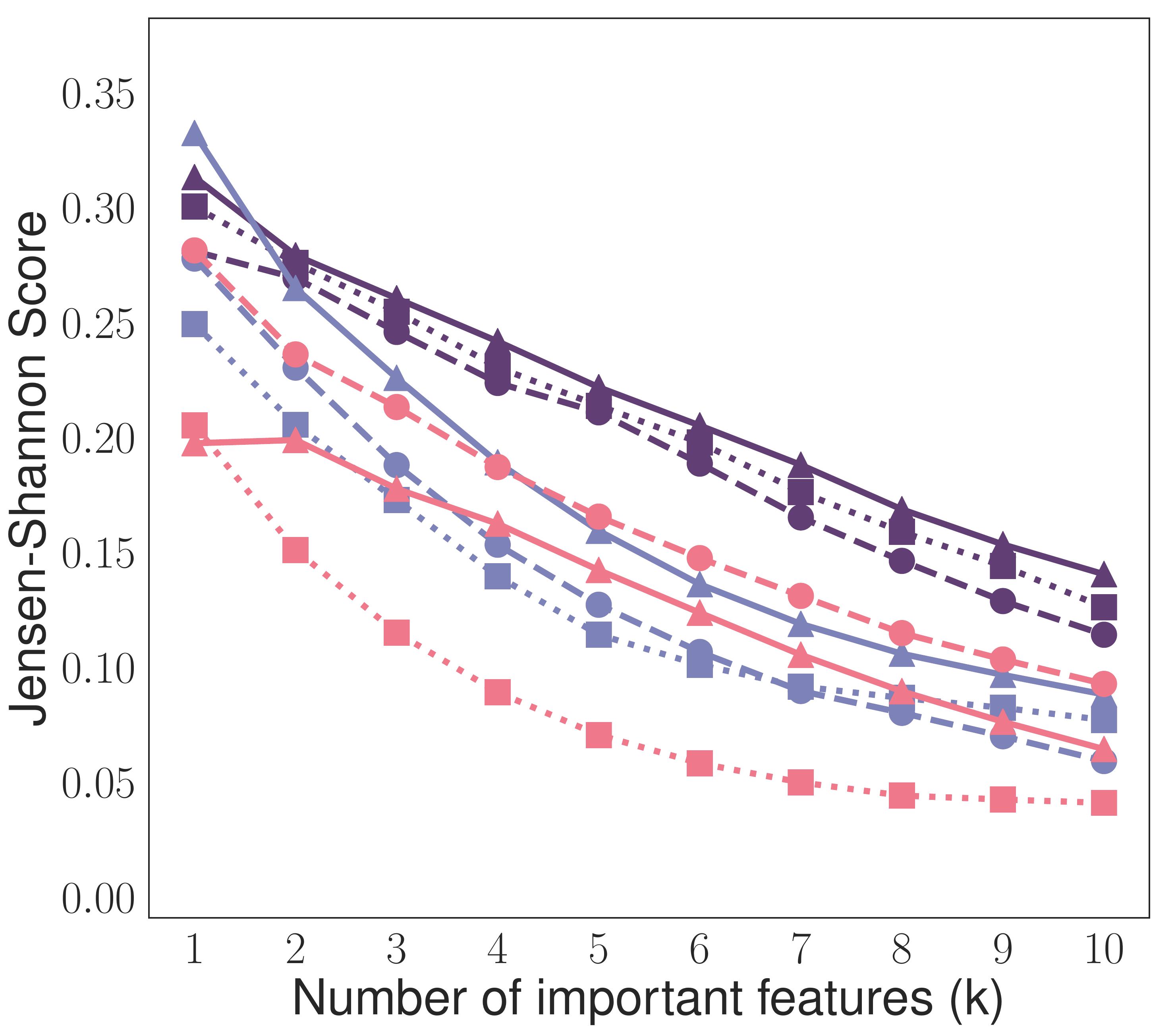}
  \caption{SST}
  \label{fig:sst_models_jensenshannon_pos}
\end{subfigure}
\hfill
\begin{subfigure}[t]{0.27\textwidth}
  \centering
  \includegraphics[trim=0 0 7.7in 0,clip,width=\textwidth]{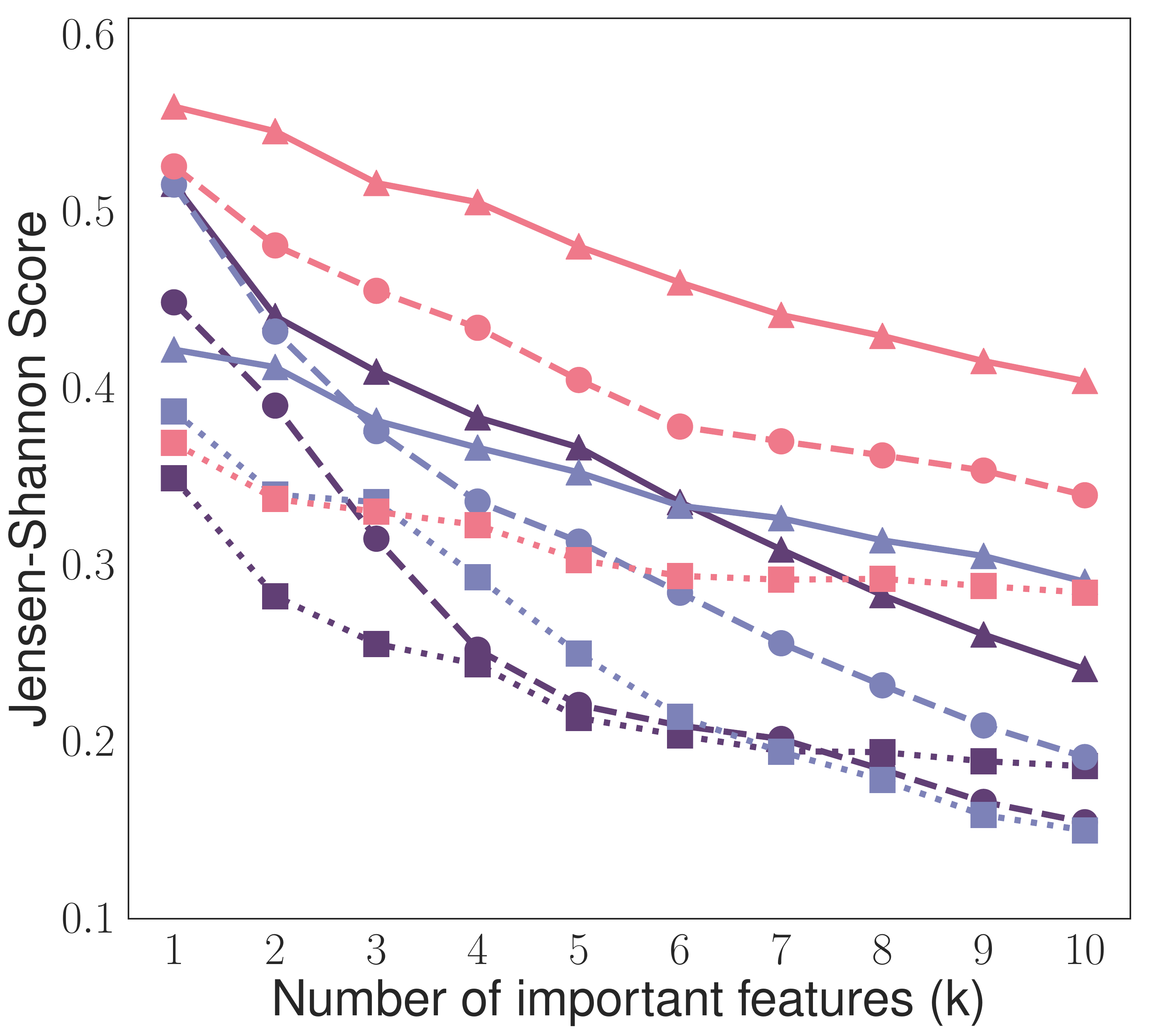}
  \caption{Deception}
  \label{fig:deception_models_jensenshannon_pos}
\end{subfigure}
\caption{Distance of the part-of-speech tag distributions between important features and all words (background).
Distance is generally smaller with post-hoc methods for all models, 
although some exceptions exist for LSTM with attention and BERT.
}
\label{fig:jensenshannon_pos_main}
\end{figure*}

\para{The similarity between important features from different methods tends to be lower for LSTM with attention (\figref{fig:similarity_models}).}
Second, we compare the similarity of feature importance derived from the same model with different methods.
For deep learning models such as LSTM with attention, the similarity between feature importance generated by different methods is the lowest, especially comparing LIME with SHAP.
Notably, the results are much more cluttered in deception detection.
Contrary to {\bf H1b}, we do not observe that LIME is more similar to SHAP than built-in.
The order seems to depend on both the task and the model: even within SST, the similarity between built-in and LIME can rank as third, second, or first. %
In other words, 
post-hoc methods generate more similar important features when we compare different models, but 
that is not the case when we fix the model.
It is reassuring that 
that similarity between any pairs is above random, with a sizable margin in most cases (BERT on SHAP is an exception; see supplementary materials).

\para{Relation with $k$.}
As the relative order between different approaches can change with $k$, we have so far only focused on relatively consistent patterns over $k$ and classification tasks.
Contrary to {\bf H1c}, the similarity between most approaches 
is not drastically greater for small $k$,
which suggests that different approaches may not even agree on the most important features.
In fact, there is no consistent trend as $k$ grows:
similarity mostly {\em increases} in SST (while our hypothesis is that it decreases), increases or stays level in Yelp, and shows varying trends in deception detection.

\section{Heterogeneity between Instances}
\label{sec:instance}

Given the overall low similarity between different methods/models,
we next investigate how the similarity may vary across instances.

\para{The similarity between models is not always greater when two models agree on the predicted label (\figref{fig:prediction}).}
One hypothesis for the overall low similarity between models is that 
different models tend to give different predictions therefore they choose different features to support their decisions.
However, we find that the similarity between models is not particularly high when they agree on the predicted label, and are sometimes even lower than when they disagree.
This is true for LIME in Yelp and for all methods in deception detection.
In SST, the similarity when the models agree on the predicted label is generally greater than when they disagree.
We 
show the comparison between SVM ($\ell_2$) and LSTM here, and
similar results hold for other combinations (see supplementary materials).
This observation suggests that feature importance may not connect with the %
predicted labels: different models 
agree for different reasons and also 
disagree for different reasons.

\para{The similarity between models and methods is generally negatively correlated with length but positively correlated with type-token ratio (\figref{fig:length}).}
Our results support {\bf H2b}:
Spearman correlation between length and similarity is mostly below $0$, which indicates that the longer an instance is, the less similar the important features are.
The negative correlation becomes stronger as $k$ grows, indicating that length has a stronger effect on similarity when we consider more top features.
However, this is not true in the case of LIME and SHAP 
where the correlation 
between length and similarity are occasionally above 0 and sometimes even the declining relationship with $k$ does not hold.
Our result on type-token ratio is opposite to {\bf H2c}: the greater the type-token ratio, the higher the similarity (see supplementary materials).
We believe that the reason is that type-token ratio is strongly negatively correlated with length (the Spearman correlation for Yelp, SST and deception dataset is -0.92, -0.59 and -0.84 respectively).
In other words, type-to-token ratio becomes 
redundant to length and fails to capture text complexity beyond length.

\section{Distribution of Important Features}
\label{sec:linguistic}

Finally, we examine the distribution of important features obtained from 
different approaches.
These results may partly explain our 
previously observed low similarity in feature importance.

\para{Important features show higher entropy using LSTM with attention and lower entropy with XGBoost (\figref{fig:entropy}).}
As expected from {\bf H3a}, 
LSTM with attention (the pink lines)
 are usually at the top (similar results for BERT in the supplementary material).
Such a high entropy can contribute to the 
low similarity between LSTM with attention and other models.
However, as the order in similarity between SVM and XGBoost is less stable, entropy cannot be the sole cause.

\para{Distribution of POS tags (\figref{fig:pos} and \figref{fig:jensenshannon_pos_main}).}
We further examine the linguistic properties of important words.
Consistent with {\bf H3b}, adjectives are more important in sentiment classification than in deception detection.
On the contrary to our hypothesis, we found that pronouns do not always play an important role in deception detection.
Notably, LSTM with attention puts a strong emphasis on nouns in deception detection.
In all cases, determiners are under-represented among important words.
With respect to the distance of part-of-speech tag distributions between important features and all words (background), 
post-hoc methods tend to bring important words closer to the background words, which echoes the previous observation that post-hoc methods tend to increase the similarity between important words (\figref{fig:jensenshannon_pos_main}).

\section{Concluding Discussion}
\label{sec:conclusion}

In this work, we provide the first systematic characterization of feature importance obtained from different approaches.
Our results show that different approaches can sometimes lead to very different important features, but there exist some consistent patterns between models and methods.
For instance, deep learning models tend to generate diverse important features that are different from traditional models;
post-hoc methods lead to more similar important features than built-in methods.

As important features are increasingly adopted for varying use cases (e.g., decision making vs. model debugging), we hope to encourage more work in understanding the space of important features, and how they should be used for different purposes.
While we focus on consistent patterns across classification tasks, it is certainly interesting to investigate how properties related to tasks and data affect the findings.
Another promising direction is to understand whether more concentrated important features (lower entropy) lead to better human performance in supporting decision making.

\section*{Acknowledgments}
We thank Jim Martin, anonymous reviewers, and members of the NLP+CSS research group at CU Boulder for their insightful comments and discussions.
This work was supported in part by NSF grant IIS-1849931.

\bibliographystyle{acl_natbib}
\bibliography{emnlp2018}

\appendix

\section{Preprocessing and Computational Details}

\para{Preprocessing.}
We used spaCy for tokenization and part-of-speech tagging. 
All the words are lowercased.
\tableref{tb:stats} shows basic data statistics.

\begin{table}[h]
\centering
\begin{tabular}{lr}
\toprule
dataset & average tokens \\
\midrule
Yelp & 134.6 \\ 
SST & 20.0 \\
Deception & 163.7 \\
\bottomrule
\end{tabular}
\caption{Dataset statistics.}
\label{tb:stats}
\end{table}

\para{Hyperparameter tuning.}
Hyperparameters for both SVM and XGBoost are tuned using cross validation.
The only hyperparameter tuned for SVM includes C.
We try a range of Cs from log space -5 to 5.
The finalized value of C ranges between 1 and 5.
Hyperparameters tuned for XGBoost include learning rate, max depth of tree, gamma, number of estimators and colsample by tree.
We lay out the range of values tried in the process of hyperparameter tuning, learning rate: 0.1 to 0.0001, max depth of tree: 3 to 7, gamma: 1 to 10, number of estimators: 1000 to 10000 and colsample by tree: 0.1 to 1.0.
Hyperparameters for LSTM with attention are tuned using the validation dataset which comprises 10\% of the entire dataset.
They include embedding dimension, hidden dimension, learning rate, number of epochs and the type of optimizer. 
The range of values tried in the process of hyperparameter tuning, hidden dimension: 256 and 512, learning rate: 0.01 to 0.0001, number of epochs: 3 to 20 and type of optimizer: SGD and adam.

\para{BERT fine-tuning.}
We fine-tuned BERT from a pre-trained BERT model provided by its original release and pytorch implementation \cite{huggingface}.
We use the same architecture of 12 layers Transformer with 12 attention heads. The hidden dimension of each layer is 768. The vocabulary size is 30522. The initial learning rate we use is $5*e^{-5}$, and we add an extra $\ell_2$ regularization on the parameters that are not bias terms or normalization layer with a coefficient of $0.01$. We do early stopping according to the validation set within the first 20 epochs with batch size no larger than 4. The attention weights we consider are the self-attention weights of the first token of each text instance, namely the attention weights from ``[CLS]'', since according to BERT's design, the first token will generate the sentence representation fed into the classification layer. For the three target tasks, we choose different maximum lengths according to their natural length. For the deception detection task, the maximum sequence length is 300 tokens. For the SST binary classification task, we choose the default 128 tokens as the maximum length and for the yelp review classification task we use 512 tokens. 
\para{BERT alignment.} Given that BERT tokenizes a text instance 
with its own tokenizer, we map the important features from BERT tokens to %
tokenize results from spaCy we used for other models. To be specific, we generate token start-end information as a tuple and call it token spans. We show an example for text instance ``It's a good day.'':

\noindent\textbf{tokenization 1}: [It's], [a], [good], [day], [.]\\
\textbf{token spans 1}: (0,3),(4,4),(5,8),(9,11),(12,12)\\
\textbf{tokenization 2}: [It], ['s], [a], [go], [od], [day], [.]\\
\textbf{token spans 2}: (0,1), (2,3), (4,4), (5,6), (7,8), (9,11), (12,12)

With the span information, we can identify how a token in the first tokenization relates to tokens in the second tokenization and then aggregate all the attention values to the sub-parts.
Formally,

$W^{(1)}_{(i,j)} = \sum_{(s, t) \text{ s.t. } t \geq i, s \leq j} \operatorname{min}(1, \frac{t-i+1}{t-s+1},\\ \frac{j-s+1}{t-s+1}) W^{(2)}_{(k,p)}.$

\noindent 
In other words, for partial span overlapping, we allocate the weight according to the span overlapping ratio. For example: if $\text{span}_i^{(1)} = (2,5) $ and $\text{span}_{k-1}^{(2)} = (2,3), \text{span}_{k}^{(2)} = (4,6)$, then $W^{(1)}_{(2,5)} = W^{(2)}_{(2,3)} + \frac{2}{3}W^{(2)}_{(4,6)} $. Here $W^{(2)}$ represents the importance weight according to the second tokenization, $W^{(1)}_{(i,j)}$ represents the aligned feature importance for the token that has span $(i,j)$ in the first tokenization.
By definition, $\sum_{(i, j)} W^{(1)}_{(i, j)} = \sum_{(i, j)} W^{(2)}_{(i, j)} = 1$ for attention values.

\para{LIME.}
We use the LimeTextExplainer and write a wrapper function that returns actual probabilities of the respective model.
Since the LinearSVM generates only binary predictions, we return 0.999 and 0.001 instead.
We use 1,000 samples for fitting the local classifier.

\para{SHAP.} We use a LinearExplainer for linear SVM, a TreeExplainer for XGBoost, and adapt the gradient-based DeepExplainer for our neural models.
The main adaptation required for the neural method is to view the embedding lookup layer as a matrix multiplication layer so that the entire network is differentiable on the input token ids.

\section{Additional Figures}

\begin{figure}
\centering
\includegraphics[width=0.5\textwidth]{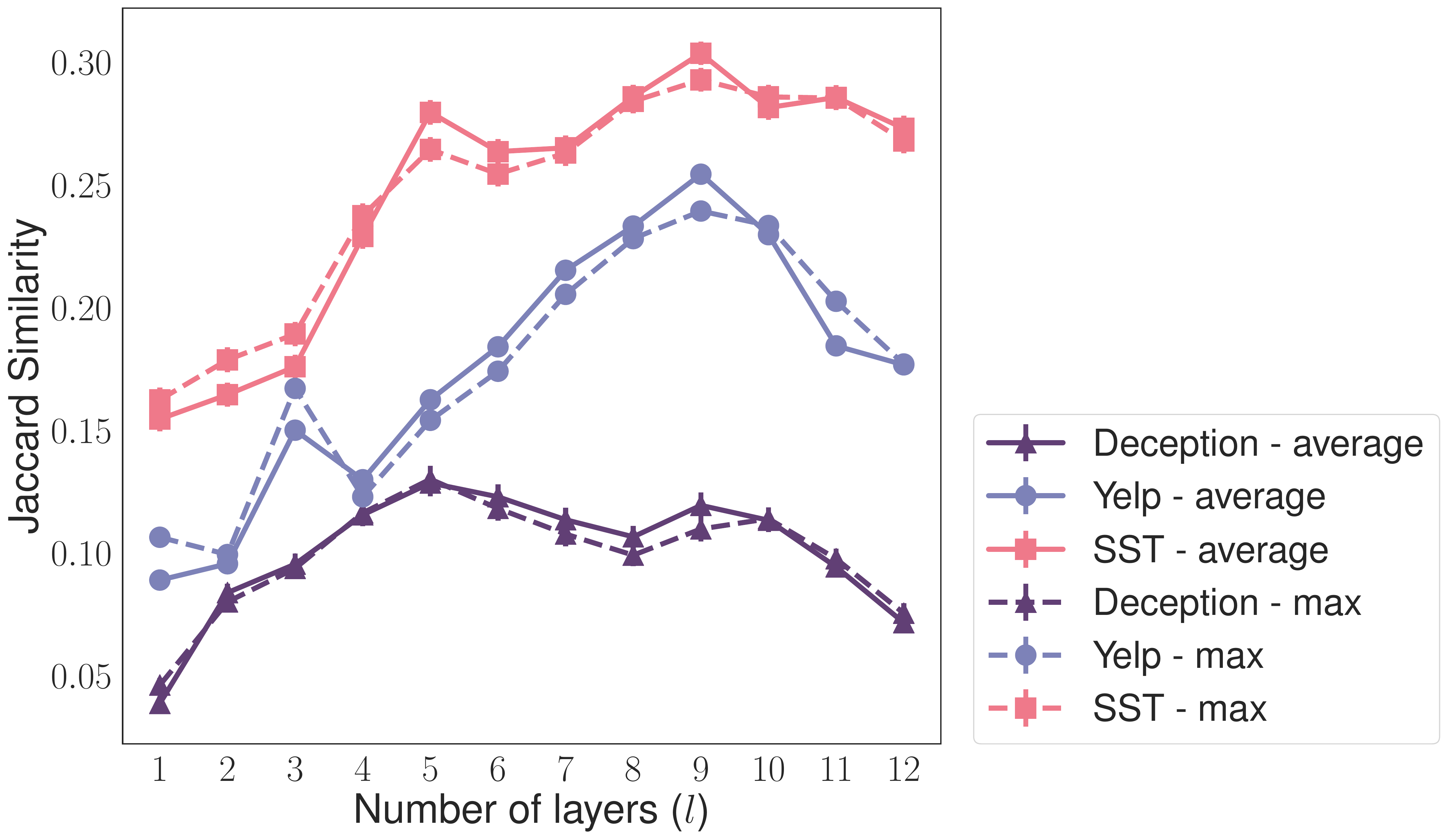}
\caption{Similarity comparison between BERT layers using average or maximum attention heads score ($k=10$).
In general, similarity becomes greater as $l$ increases, but the last layer is not necessarily the greatest.
Similarity is slightly higher when average attention heads score is computed.
}
\label{fig:layer}
\end{figure}

\para{Similarity between BERT layers and SVM ($\ell_2$).}
Important features using the final layer are more similar to that from SVM ($\ell_2$) than using the first layer.
See \figref{fig:layer}.

\begin{figure*}[t!]
\centering
Similarity comparison between models using the built-in method \\
\begin{subfigure}[t]{0.42\textwidth}
  \centering
  \includegraphics[width=\textwidth]{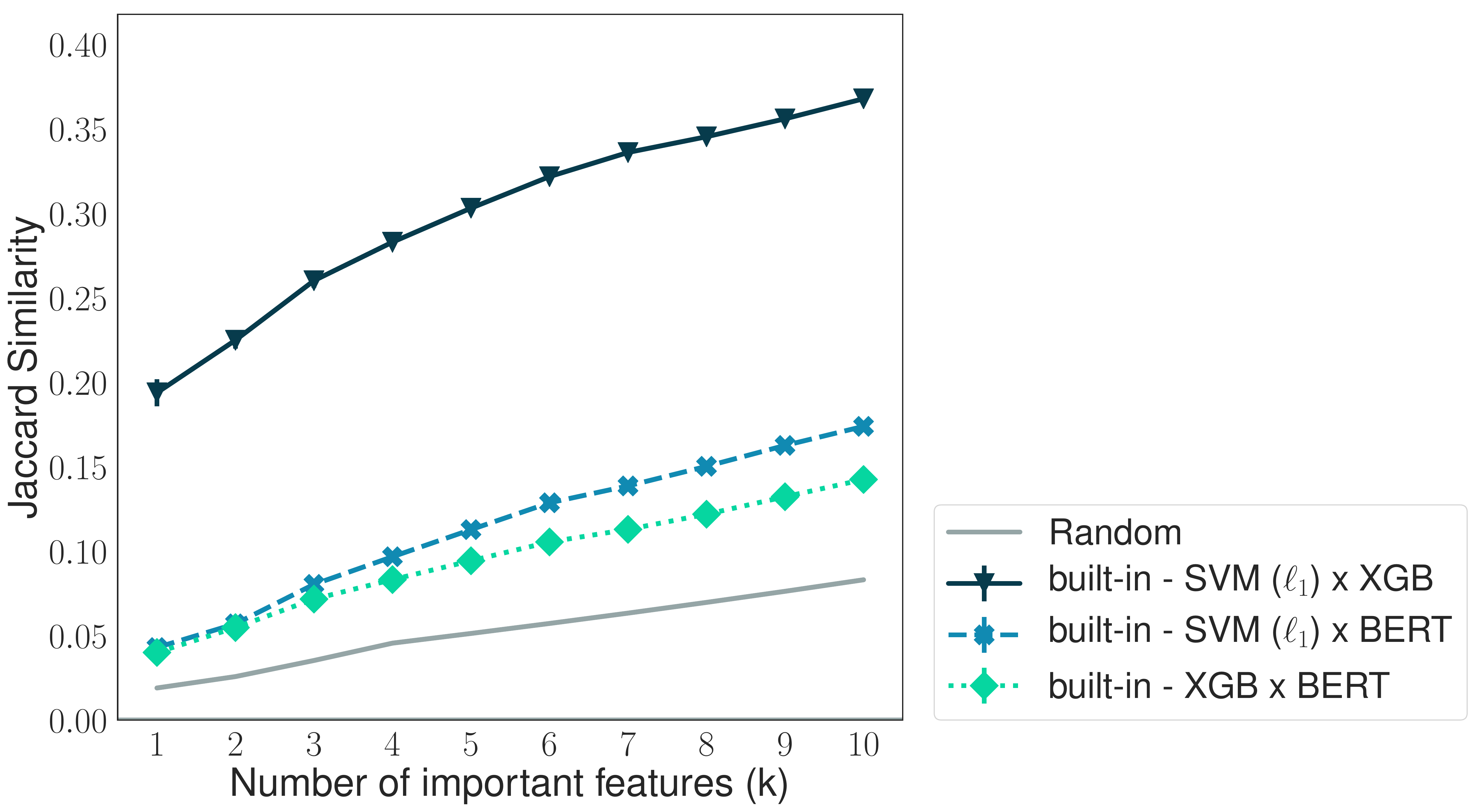}
  \caption{Yelp}
  \label{fig:yelp_methods}
\end{subfigure}
\hfill
\begin{subfigure}[t]{0.27\textwidth}
  \centering
  \includegraphics[trim=0 0 7.1in 0,clip,width=\textwidth]{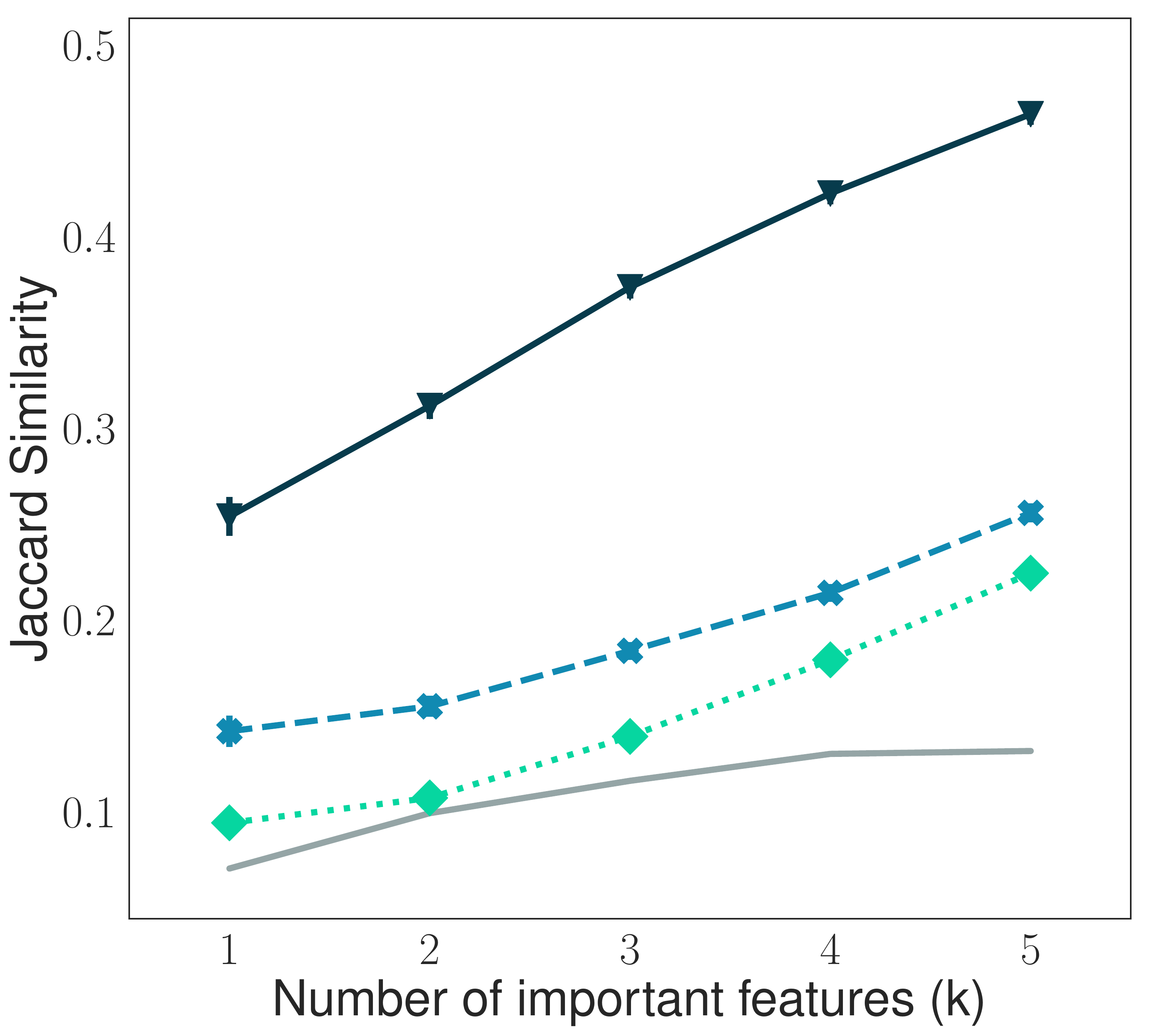}
  \caption{SST}
  \label{fig:sst_methods}
\end{subfigure}
\hfill
\begin{subfigure}[t]{0.27\textwidth}
  \centering
  \includegraphics[trim=0 0 7.1in 0,clip,width=\textwidth]{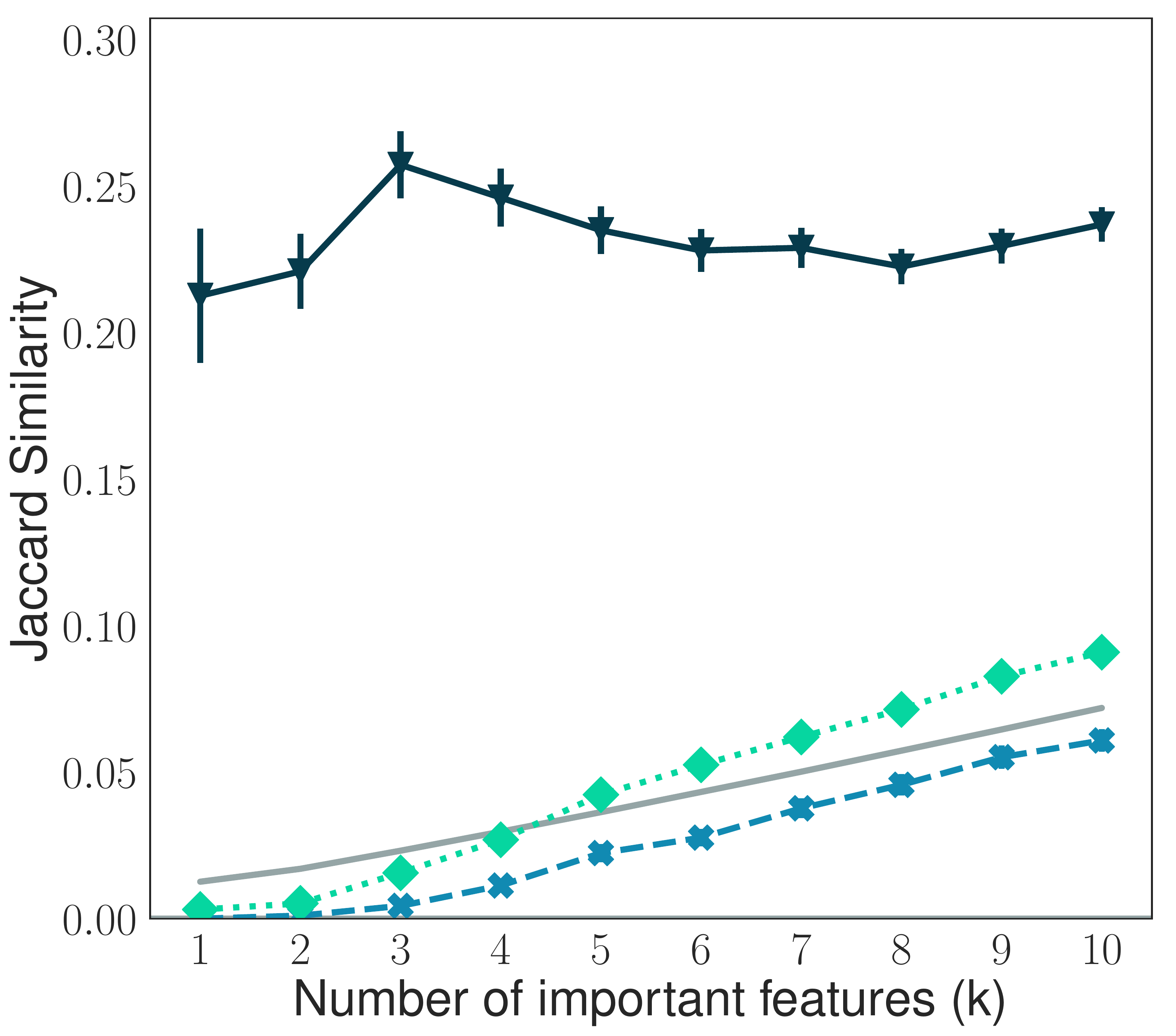}
  \caption{Deception}
  \label{fig:deception_methods}
\end{subfigure}\\\bigskip
Comparison between the built-in method and post-hoc methods \\
\begin{subfigure}[t]{0.42\textwidth}
  \centering
  \includegraphics[width=\textwidth]{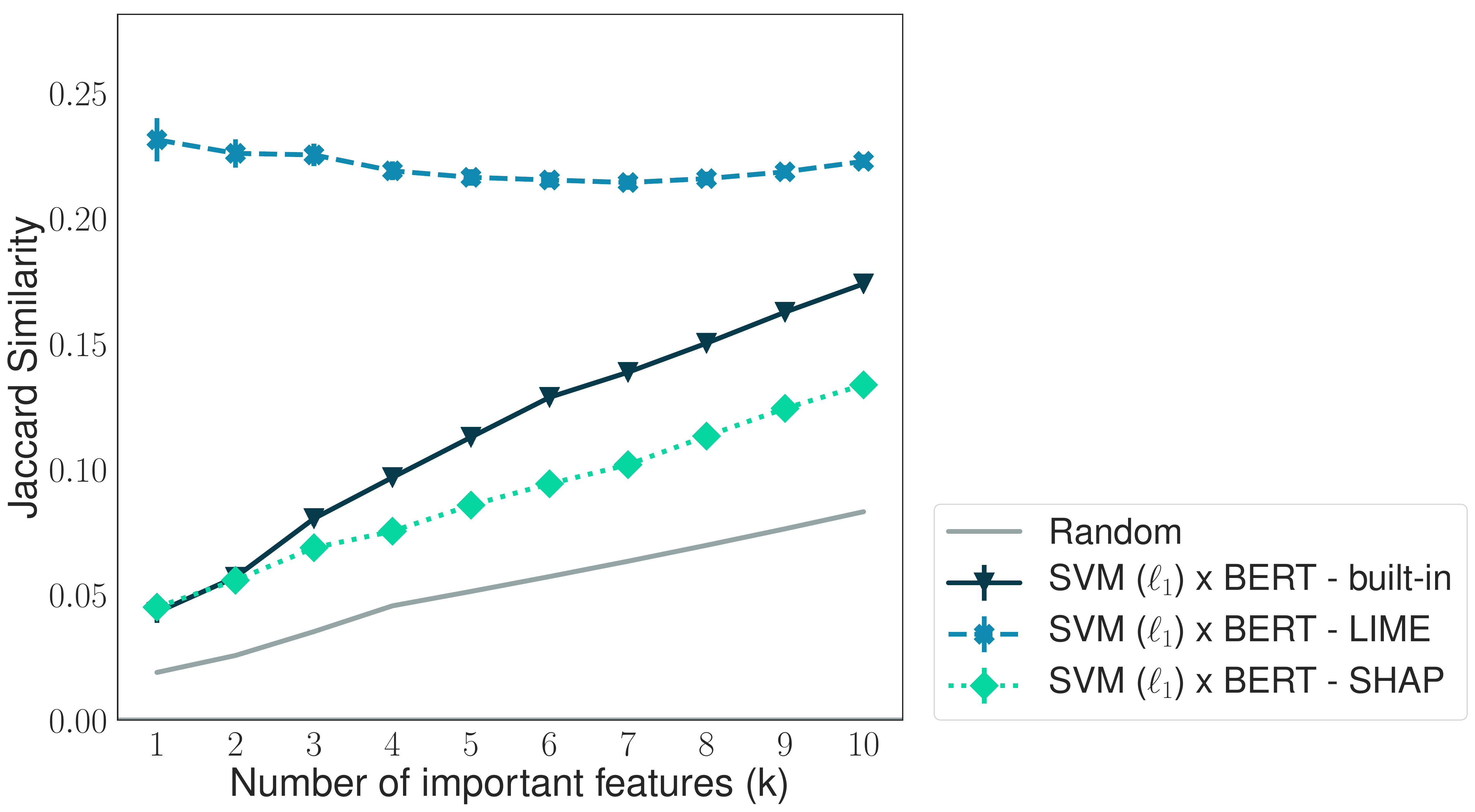}
  \caption{Yelp}
  \label{fig:yelp_methods}
\end{subfigure}
\hfill
\begin{subfigure}[t]{0.27\textwidth}
  \centering
  \includegraphics[trim=0 0 7.1in 0,clip,width=\textwidth]{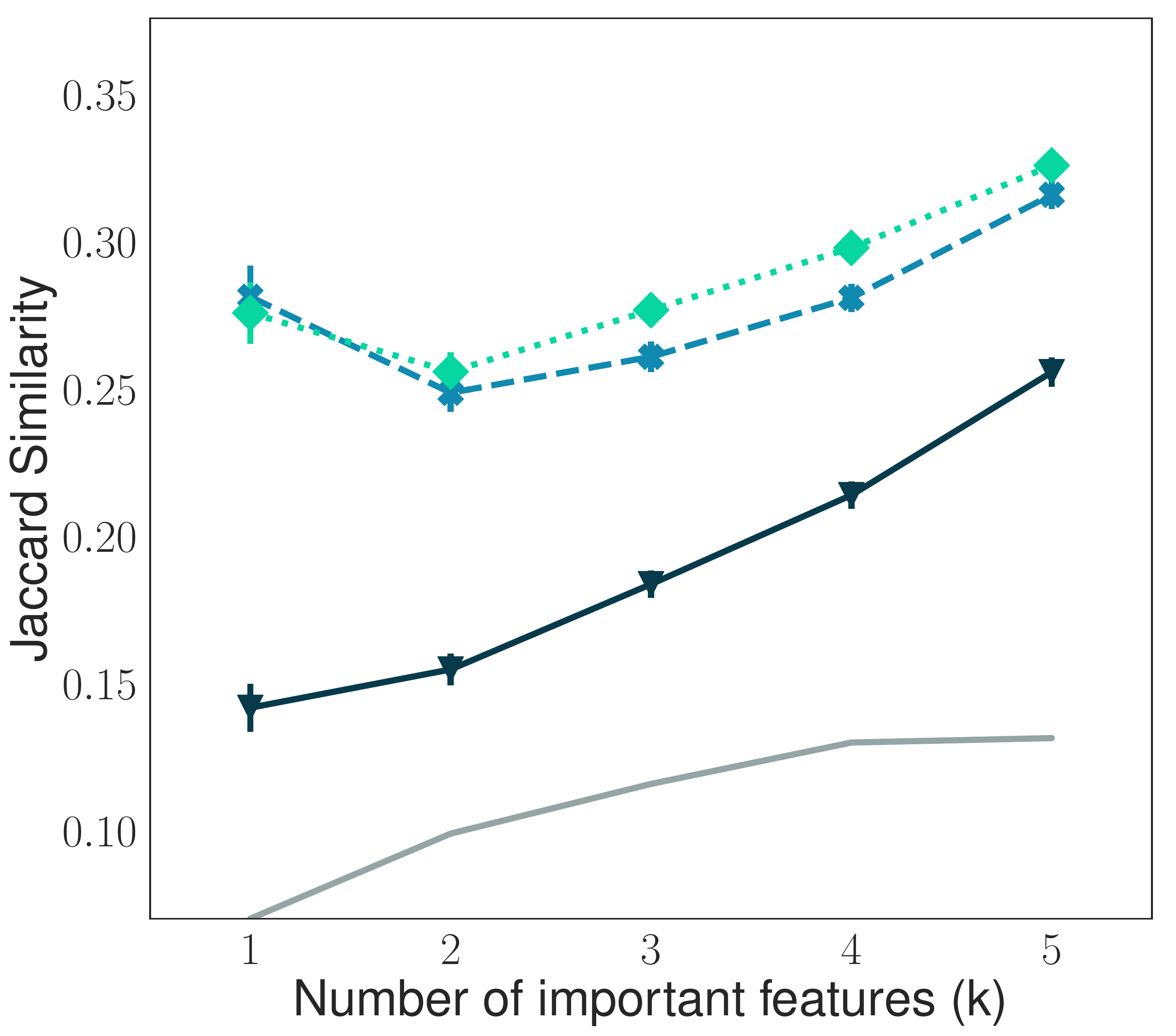}
  \caption{SST}
  \label{fig:sst_methods}
\end{subfigure}
\hfill
\begin{subfigure}[t]{0.27\textwidth}
  \centering
  \includegraphics[trim=0 0 7.1in 0,clip,width=\textwidth]{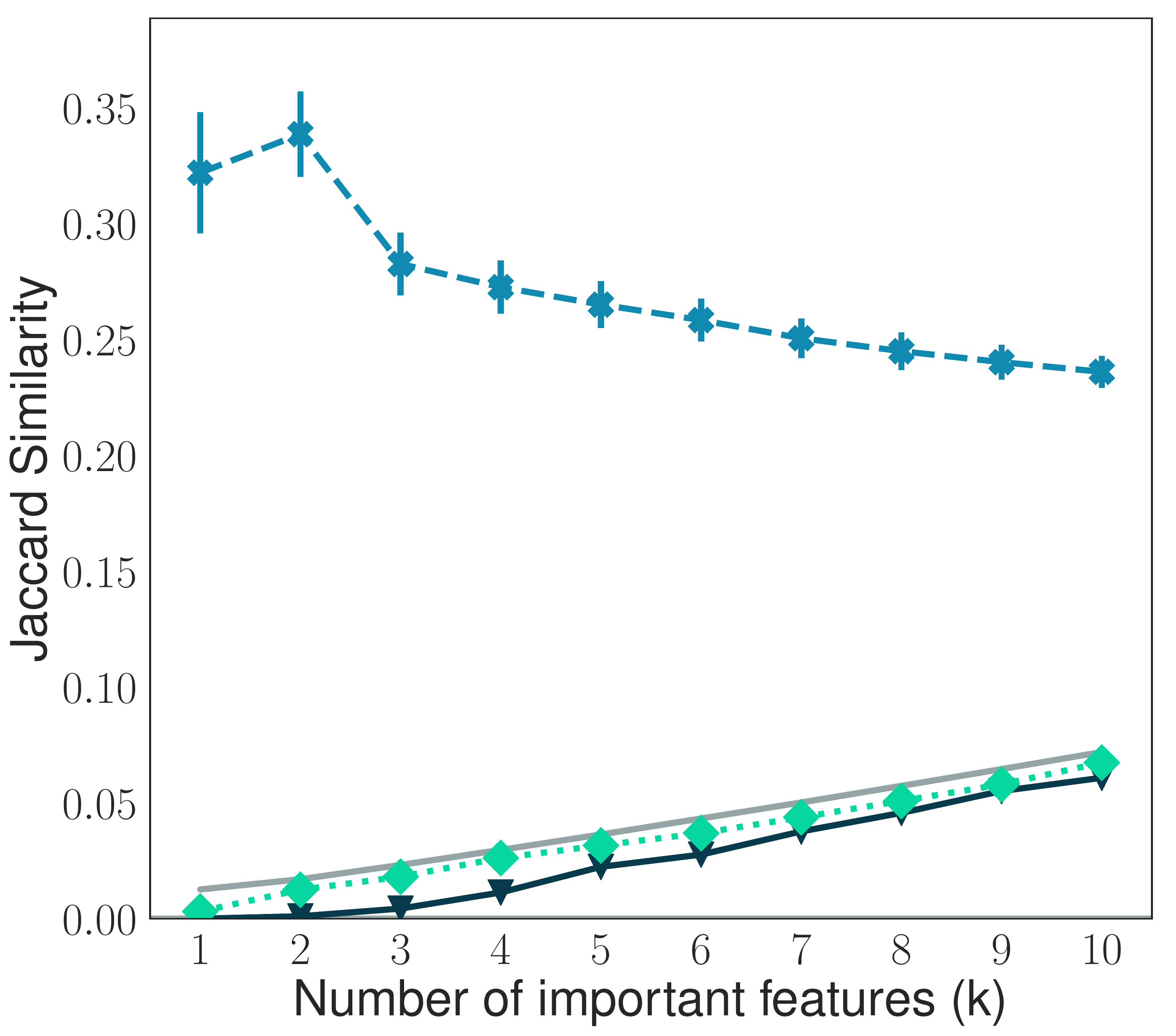}
  \caption{Deception}
  \label{fig:deception_methods}
\end{subfigure}
\caption{Similarity comparison between models with the same method.
$x$-axis represents the number of important features that we consider, while $y$-axis shows the Jaccard similarity.
{\em Error bars represent standard error throughout the paper.}
The top row compares three pairs of models using the built-in method, while the second row compares three methods on SVM ($\ell_1$) and BERT.
The random line is derived using the average similarity between two random samples of $k$ features from 100 draws.
}
\label{fig:similarity_supp}
\end{figure*}

\para{Built-in similarity is much lower with deep learning models, and post-hoc methods ``smooth'' the distance.} 
Similar results are observed in SVM ($\ell_1$) and BERT.
See \figref{fig:similarity_supp}.

\begin{figure*}[t!]
\centering
\begin{subfigure}[t]{0.42\textwidth}
  \centering
  \includegraphics[width=\textwidth]{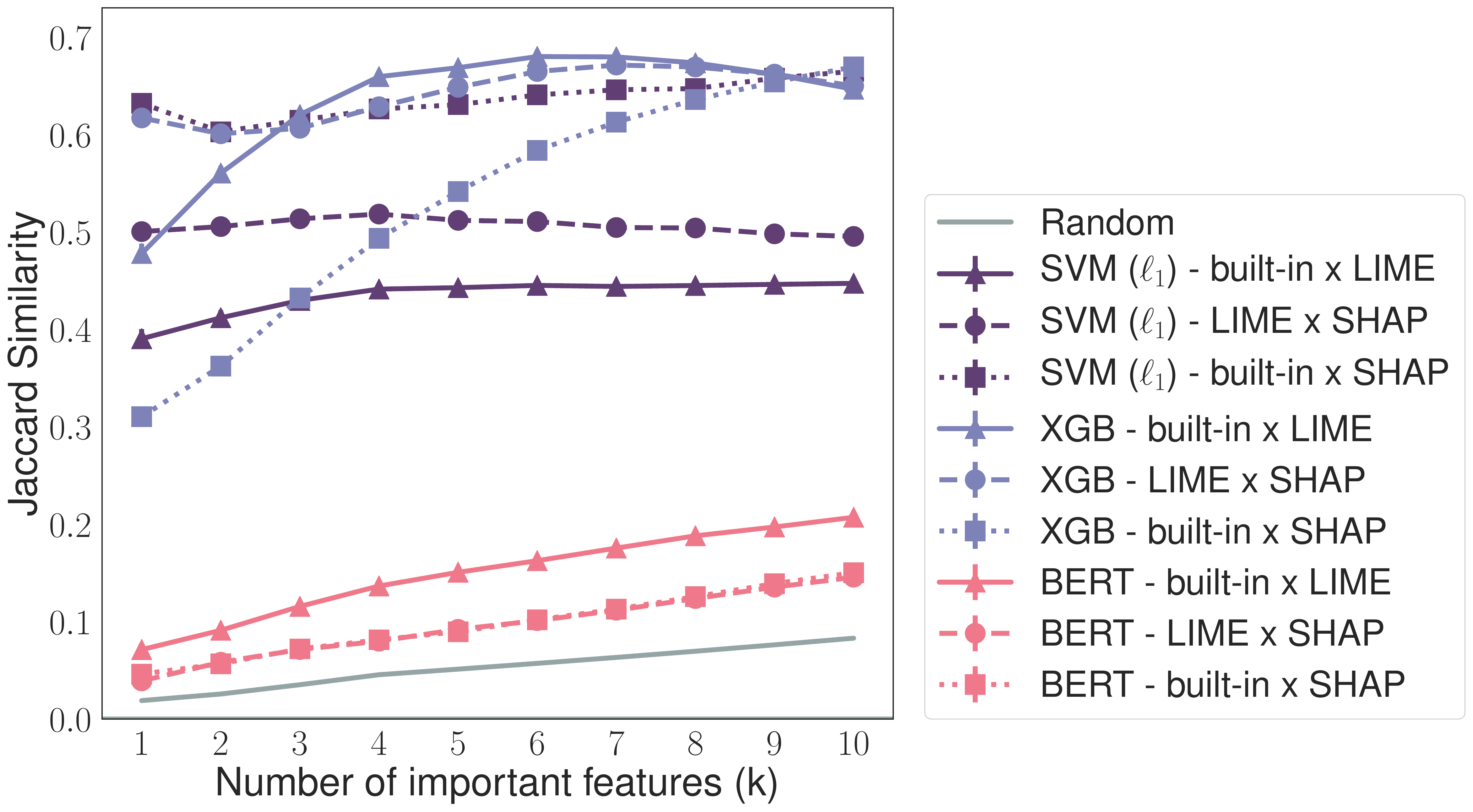}
  \caption{Yelp}
  \label{fig:yelp_models}
\end{subfigure}
\hfill
\begin{subfigure}[t]{0.27\textwidth}
  \centering
 \includegraphics[trim=0 0 7in 0,clip,width=\textwidth]{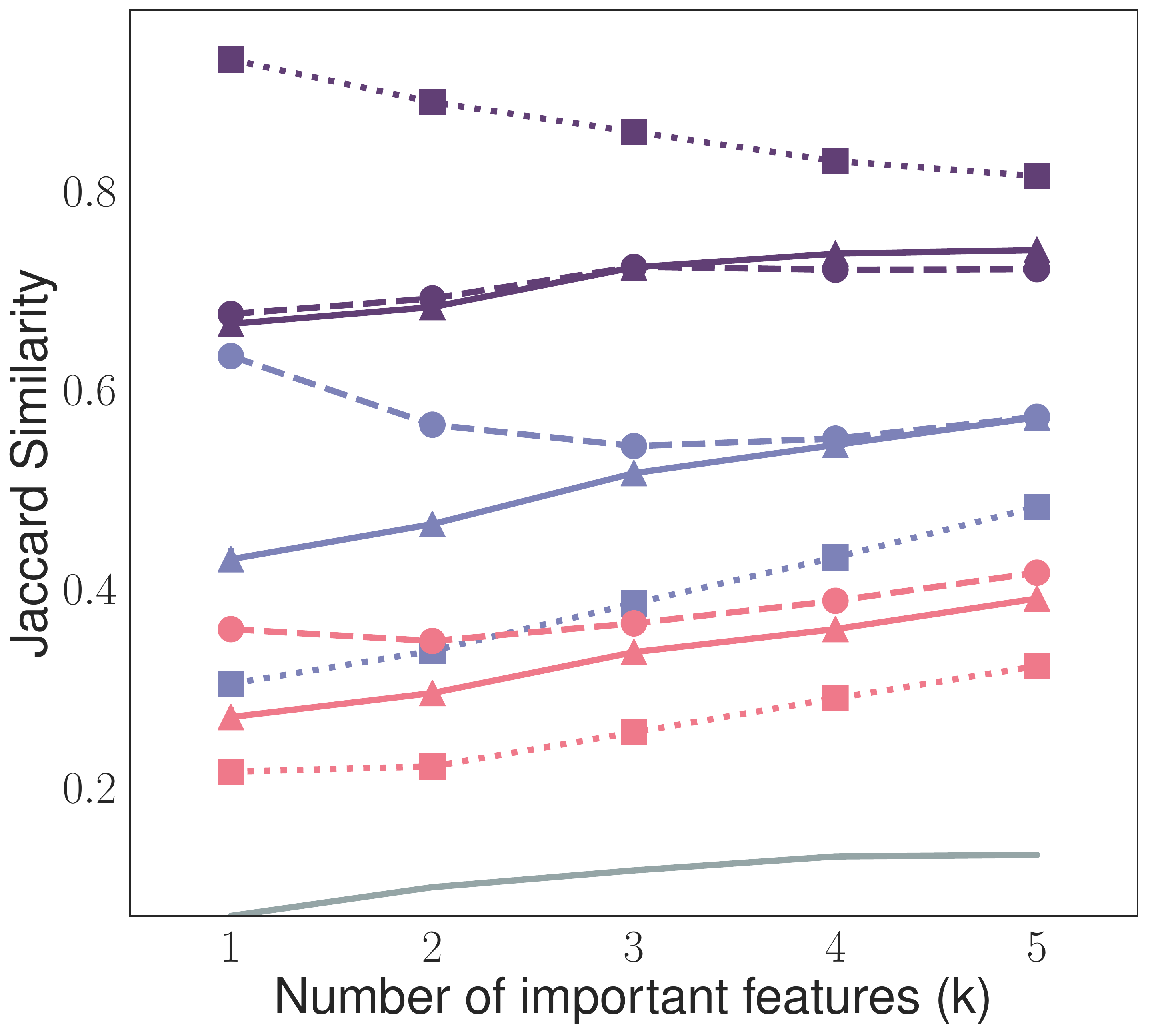}
  \caption{SST}
  \label{fig:sst_models}
\end{subfigure}
\hfill
\begin{subfigure}[t]{0.27\textwidth}
  \centering
  \includegraphics[trim=0 0 7in 0,clip,width=\textwidth]{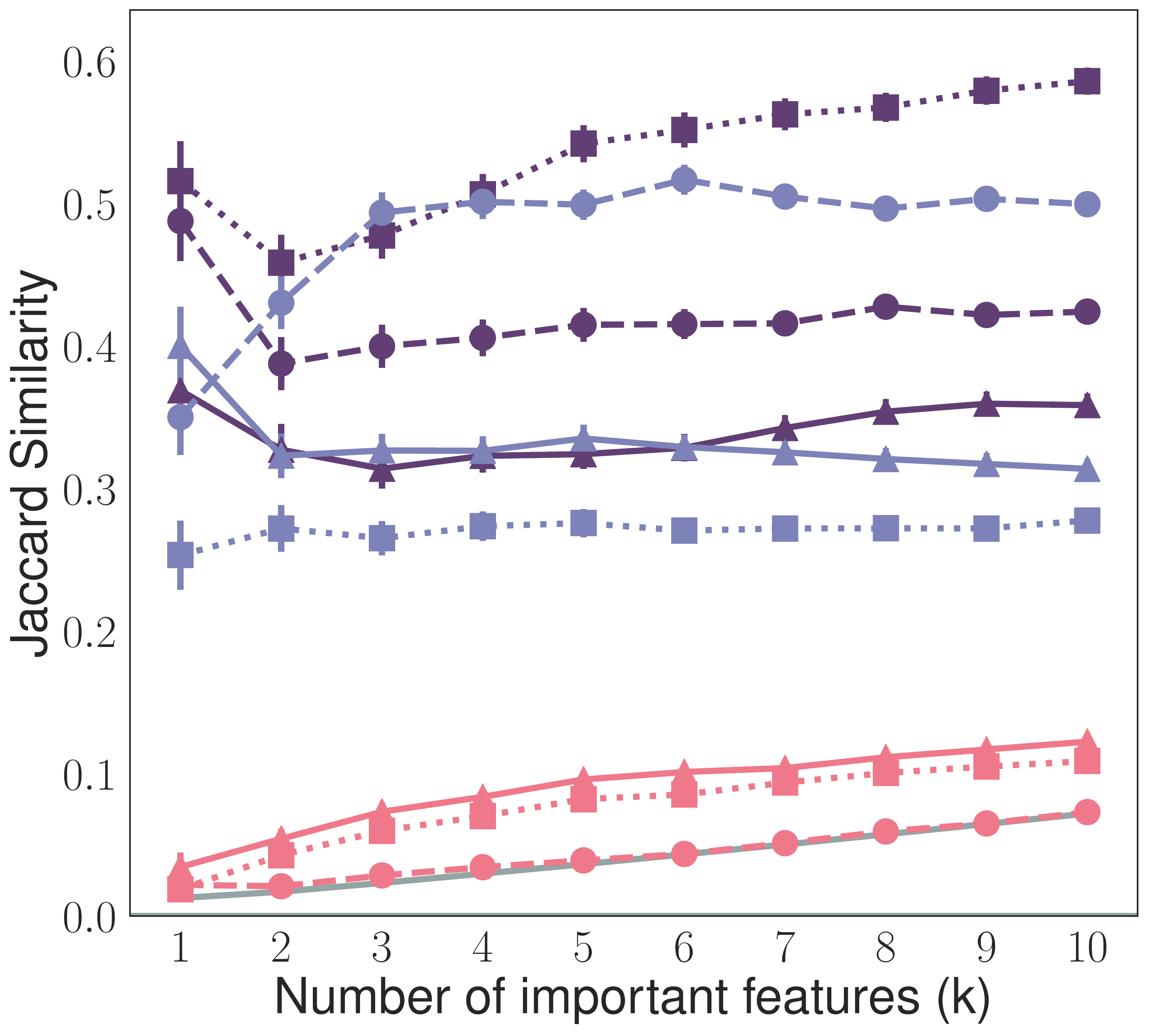}
  \caption{Deception}
  \label{fig:deception_models}
\end{subfigure}
\caption{Similarity comparison between methods using the same model for SVM ($\ell_1$), XGBoost, and BERT.
BERT is much closer to random in deception.
}
\label{fig:similarity_models_supp}
\end{figure*}

\para{Similarity between methods is lower for deep learning models.}
Similar results are observed in SVM ($\ell_1$), XGBoost and BERT.
See \figref{fig:similarity_models_supp}.

\begin{figure*}[t]
\centering
SVM ($\ell_2$) vs. XGBoost \\
\begin{subfigure}[t]{0.42\textwidth}
  \centering
  \includegraphics[width=\textwidth]{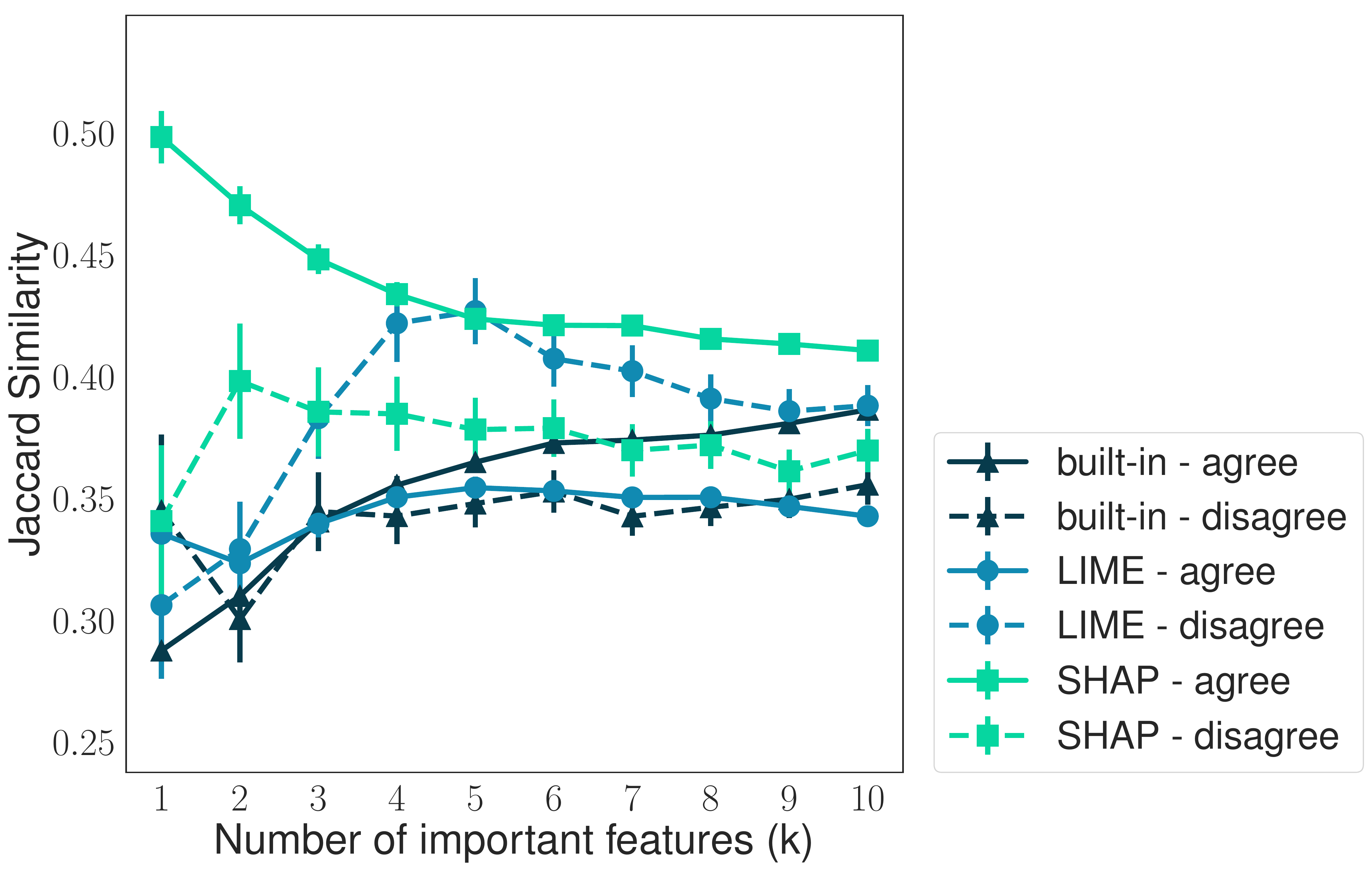}
  \caption{Yelp}
  \label{fig:yelp_svm_xgb_pred}
\end{subfigure}
\hfill
\begin{subfigure}[t]{0.275\textwidth}
  \centering
  \includegraphics[trim=0 0 5.5in 0,clip,width=\textwidth]{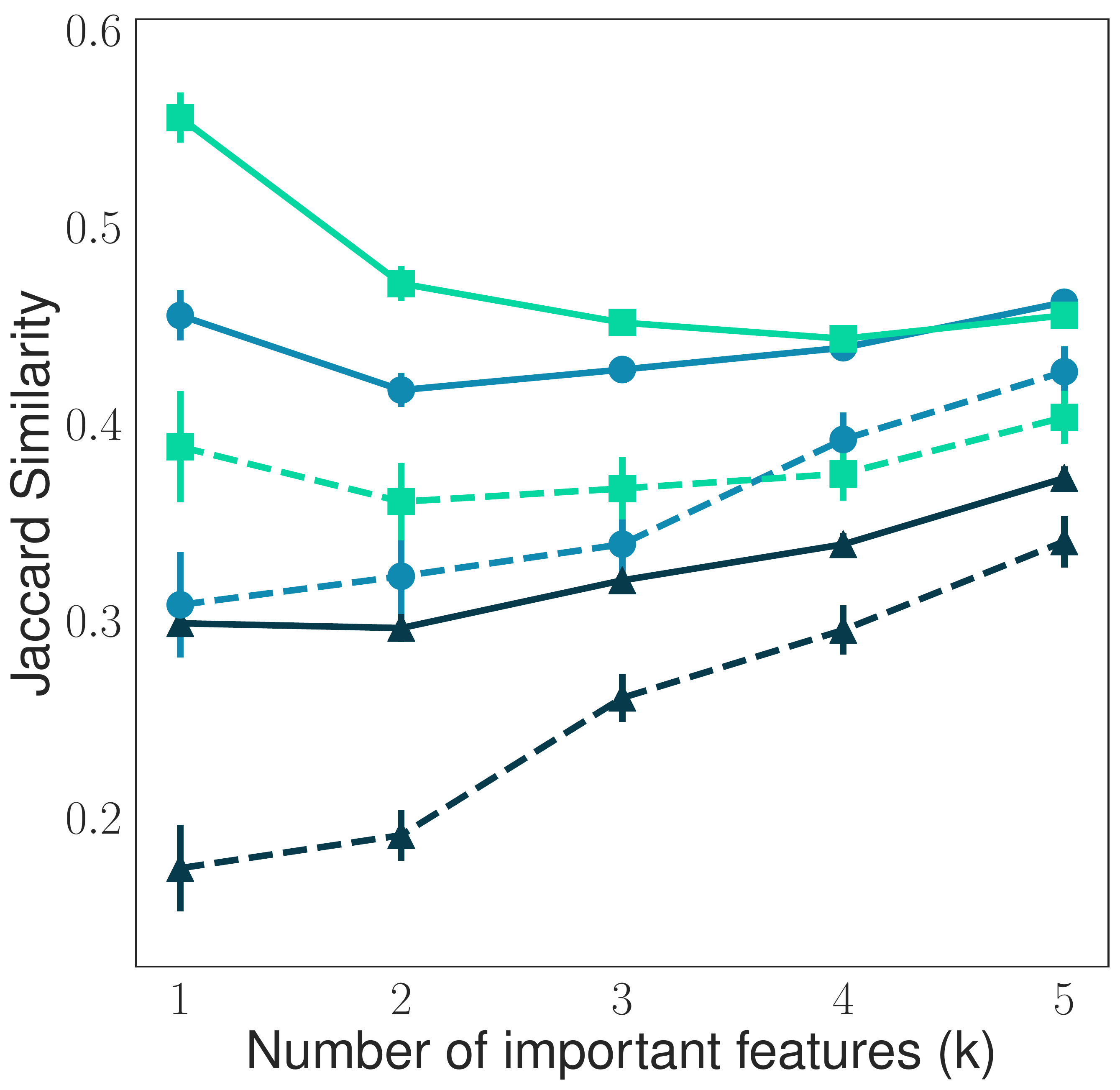}
  \caption{SST}
  \label{fig:sst_svm_xgb_pred}
\end{subfigure}
\hfill
\begin{subfigure}[t]{0.28\textwidth}
  \centering
  \includegraphics[trim=0 0 5.5in 0,clip,width=\textwidth]{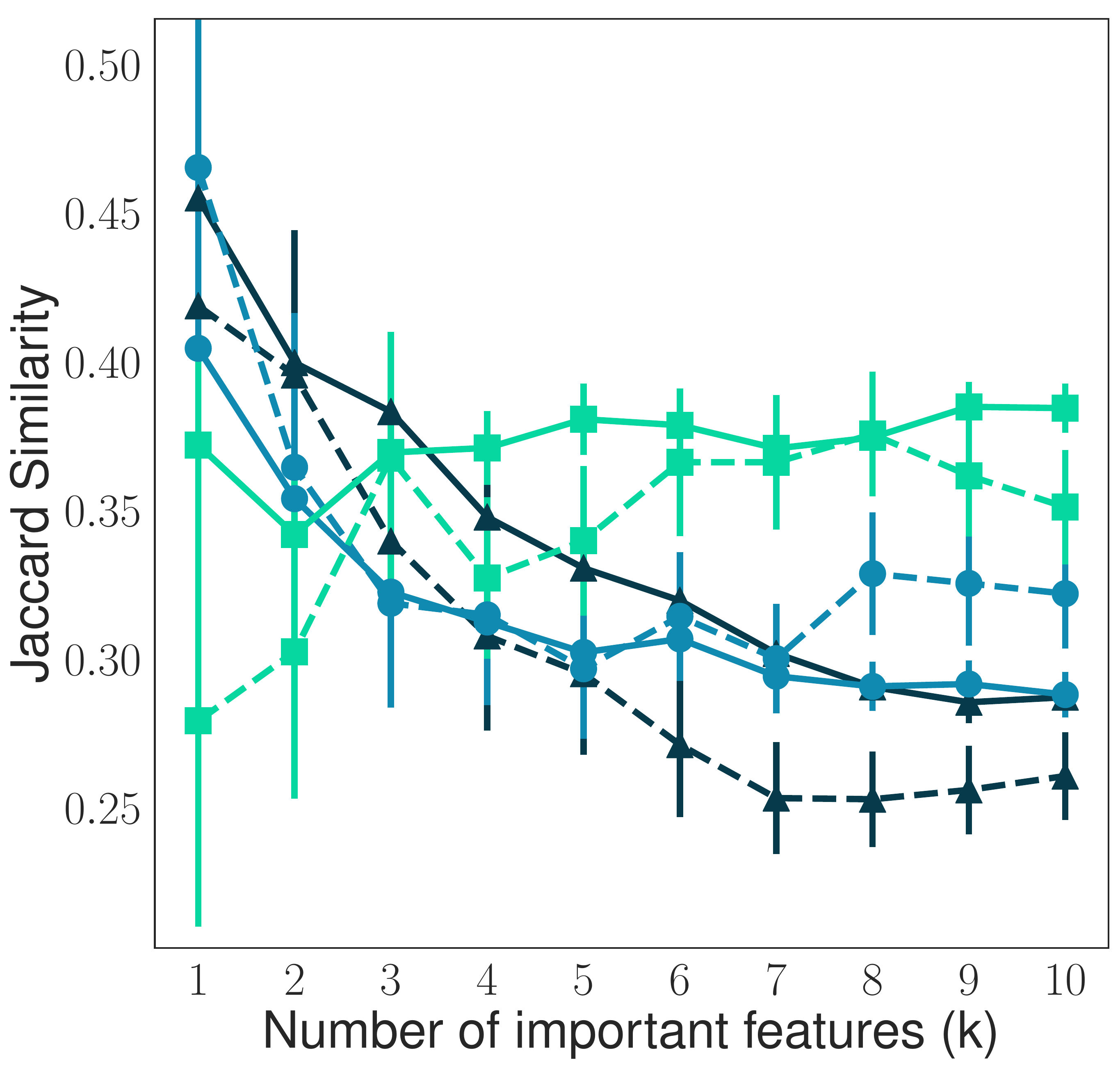}
  \caption{Deception}
  \label{fig:deception_svm_xgb_pred}
\end{subfigure}
XGBoost vs. LSTM with attention\\
\begin{subfigure}[t]{0.42\textwidth}
  \centering
  \includegraphics[width=\textwidth]{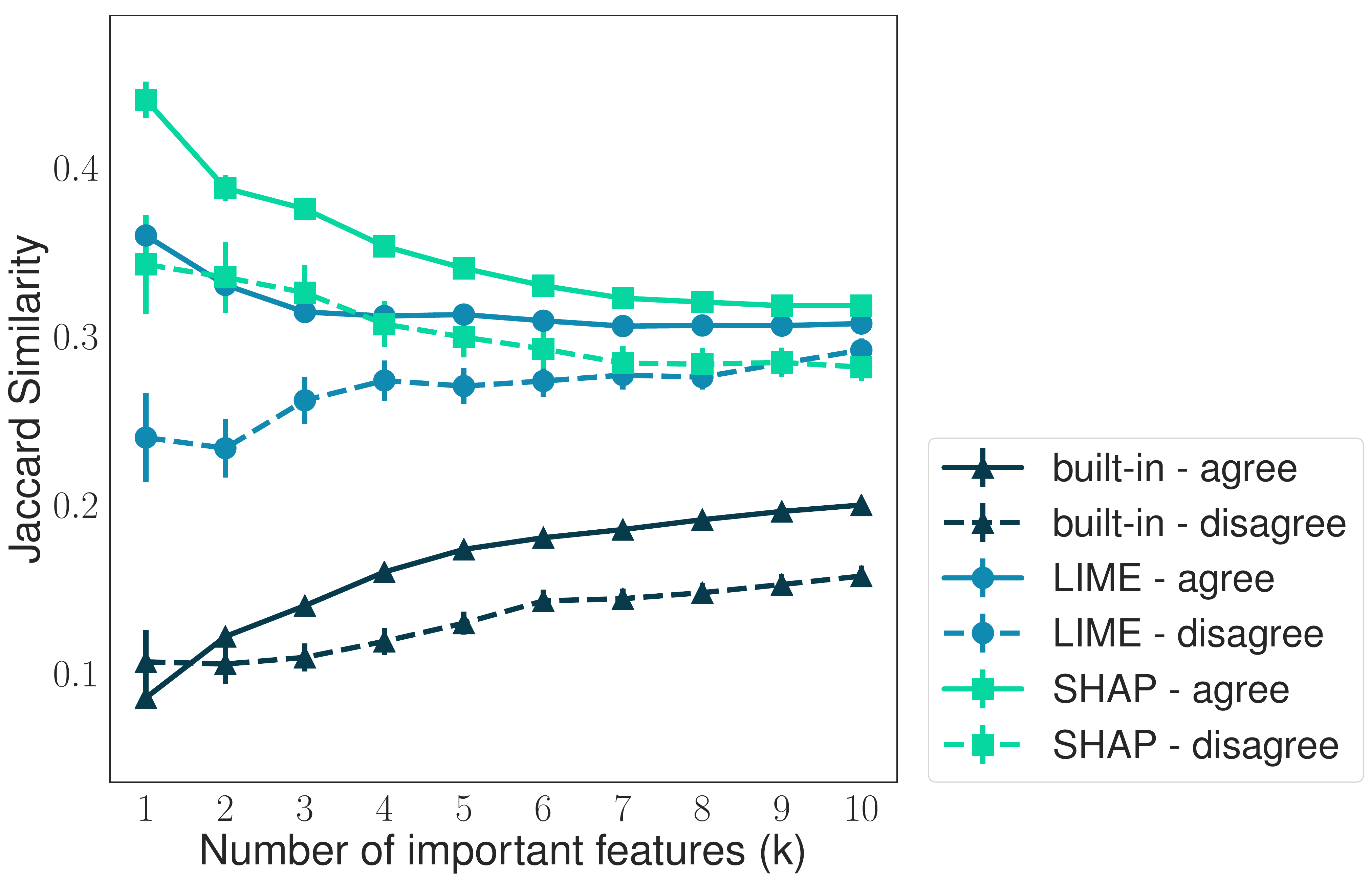}
  \caption{Yelp}
  \label{fig:yelp_xgb_lstm_att_pred}
\end{subfigure}
\hfill
\begin{subfigure}[t]{0.285\textwidth}
  \centering
  \includegraphics[trim=0 0 5.5in 0,clip,width=\textwidth]{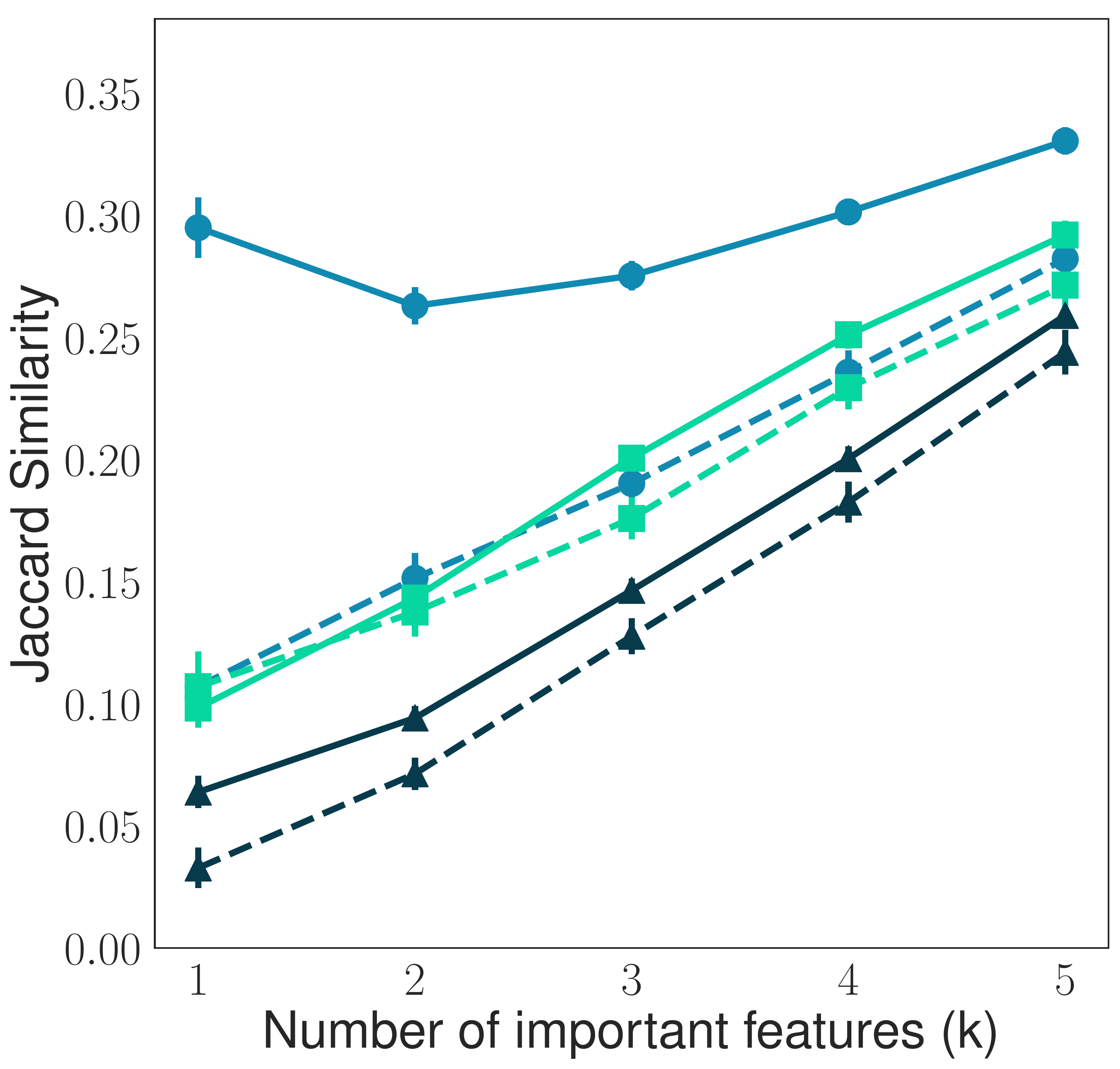}
  \caption{SST}
  \label{fig:sst_xgb_lstm_att_pred}
\end{subfigure}
\hfill
\begin{subfigure}[t]{0.28\textwidth}
  \centering
  \includegraphics[trim=0 0 5.5in 0,clip,width=\textwidth]{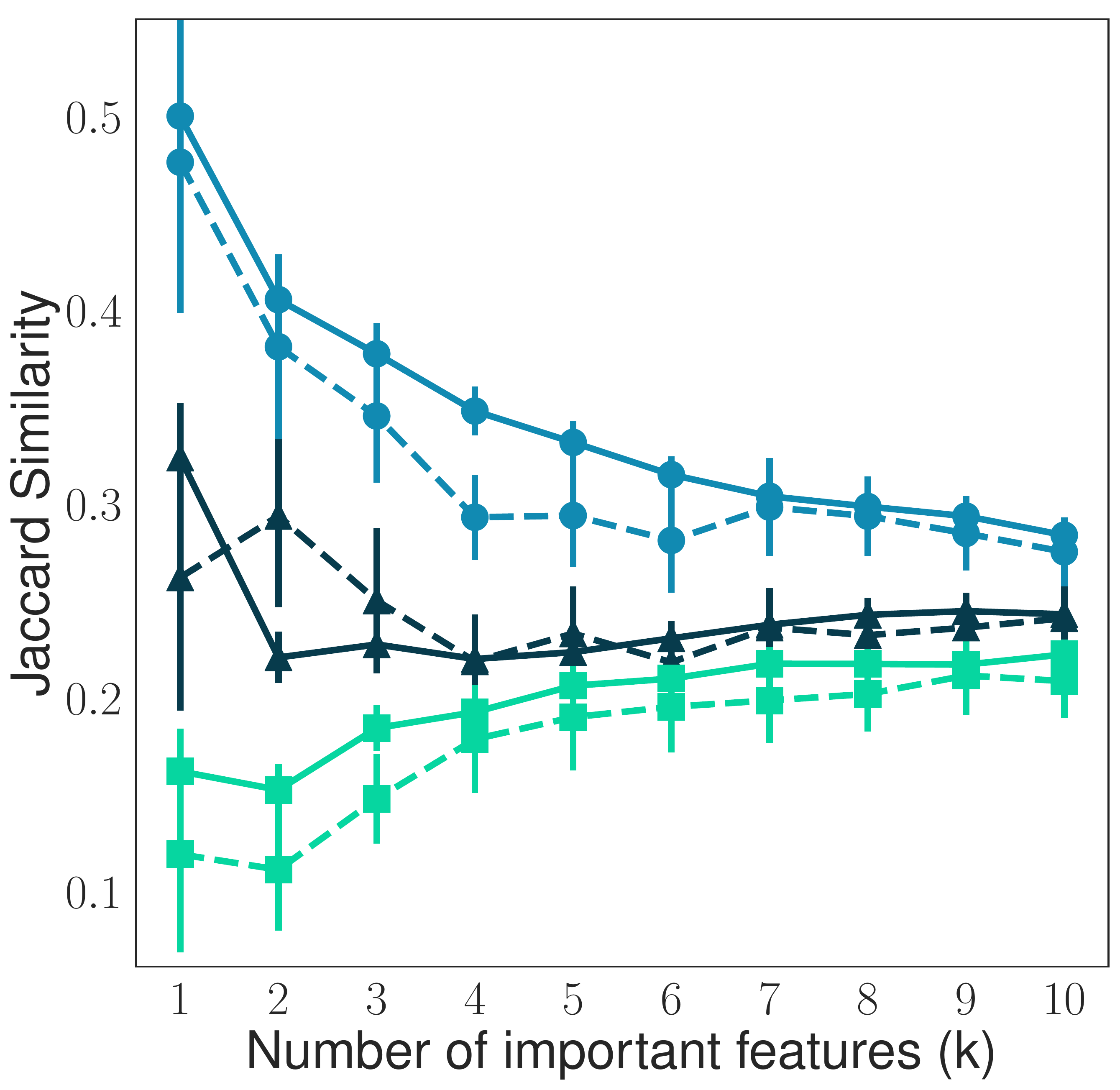}
  \caption{Deception}
  \label{fig:deception_xgb_lstm_att_pred}
\end{subfigure}
\caption{
Similarity between two models is not necessarily greater when they agree on the predictions, and 
sometimes, e.g., 
SVM ($\ell_2$) x XGB with LIME method, it is sometimes lower than when they disagree on the predicted labels.
}
\label{fig:prediction_supp}
\end{figure*}

\begin{figure*}[t]
\centering
SVM ($\ell_1$) vs. XGBoost \\
\begin{subfigure}[t]{0.42\textwidth}
  \centering
  \includegraphics[width=\textwidth]{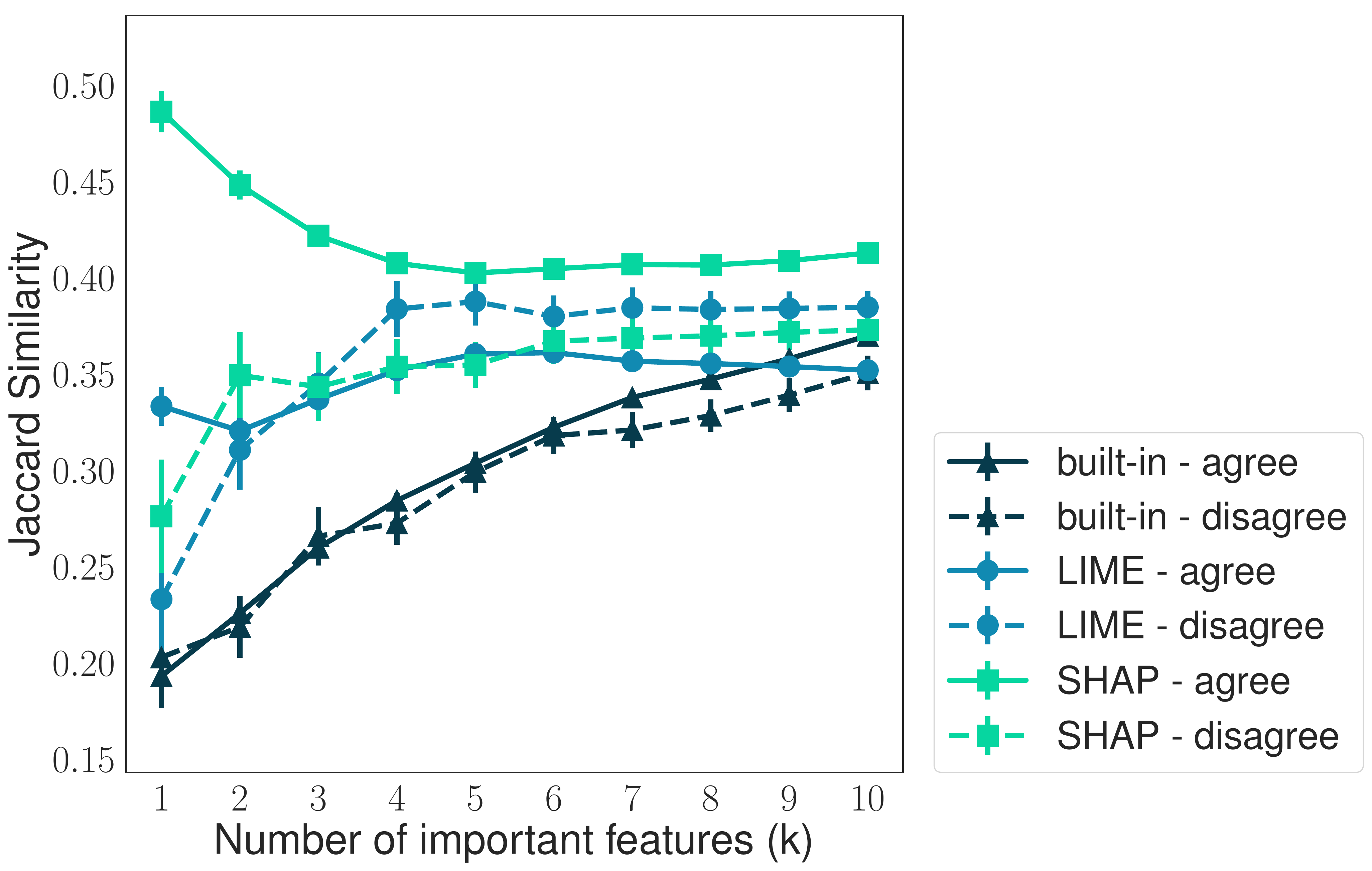}
  \caption{Yelp}
  \label{fig:yelp_svm_xgb_pred}
\end{subfigure}
\hfill
\begin{subfigure}[t]{0.275\textwidth}
  \centering
  \includegraphics[trim=0 0 5.5in 0,clip,width=\textwidth]{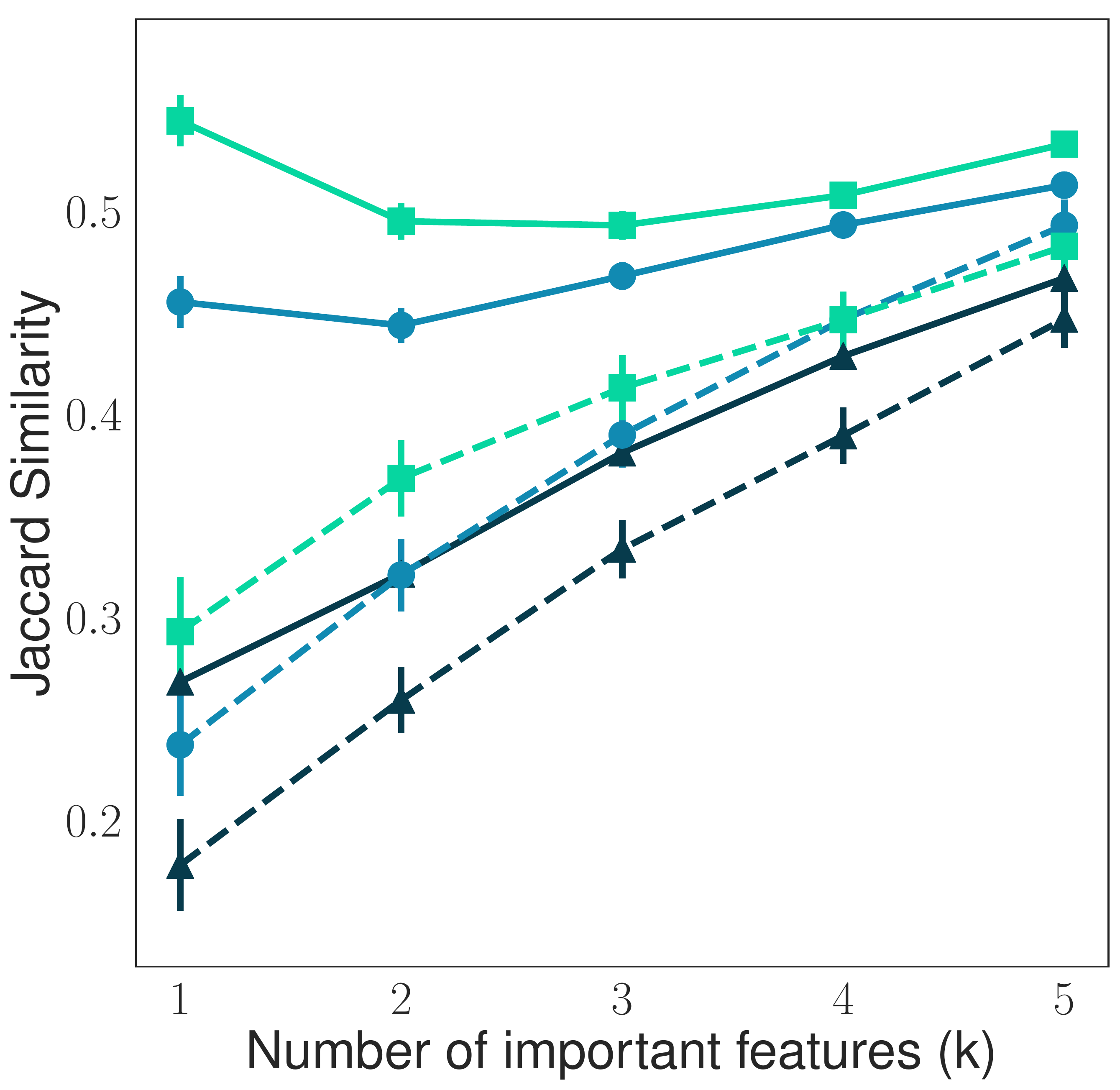}
  \caption{SST}
  \label{fig:sst_svm_xgb_pred}
\end{subfigure}
\hfill
\begin{subfigure}[t]{0.28\textwidth}
  \centering
  \includegraphics[trim=0 0 5.5in 0,clip,width=\textwidth]{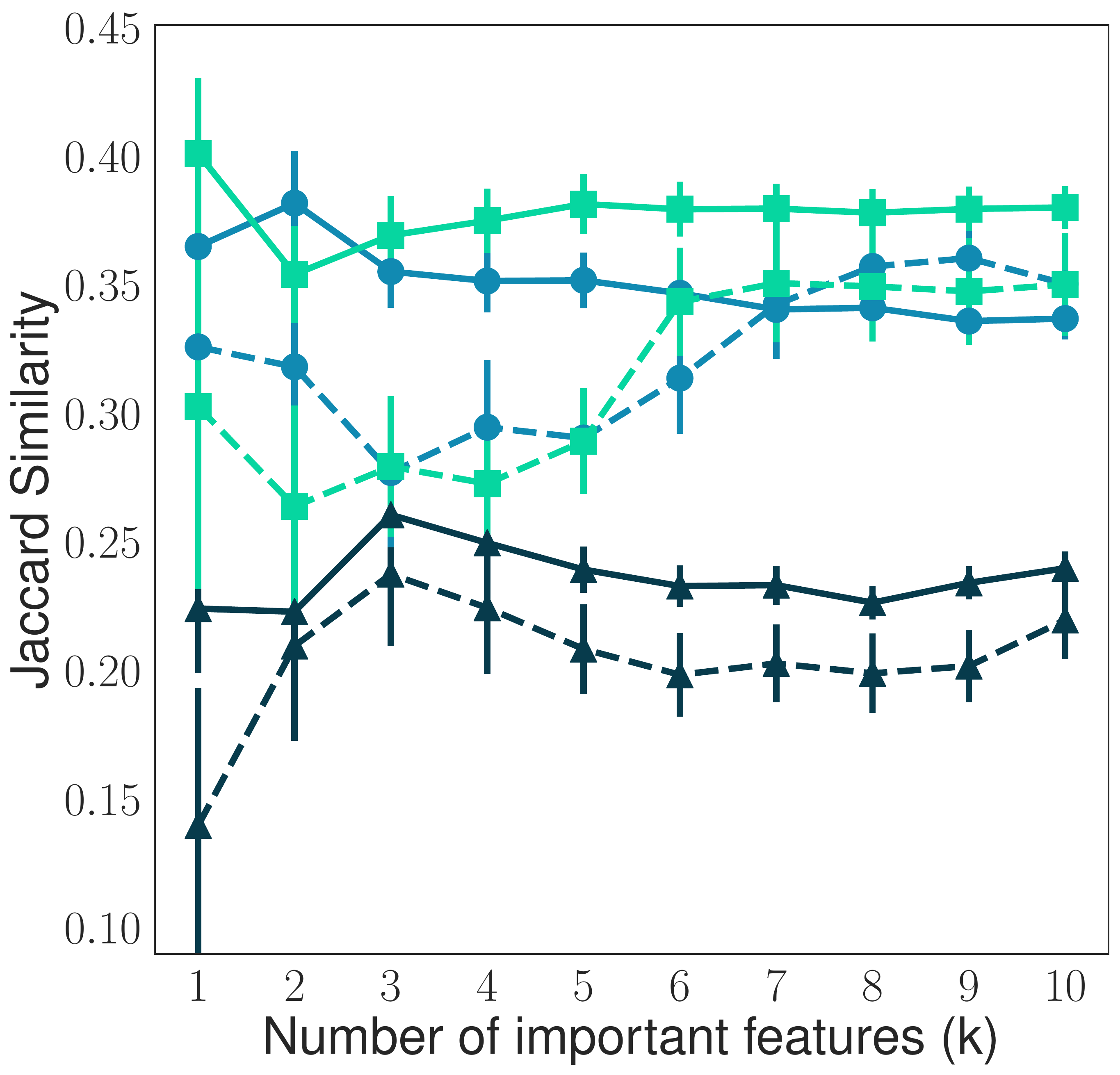}
  \caption{Deception}
  \label{fig:deception_svm_xgb_pred}
\end{subfigure}
XGBoost vs. BERT \\
\begin{subfigure}[t]{0.42\textwidth}
  \centering
  \includegraphics[width=\textwidth]{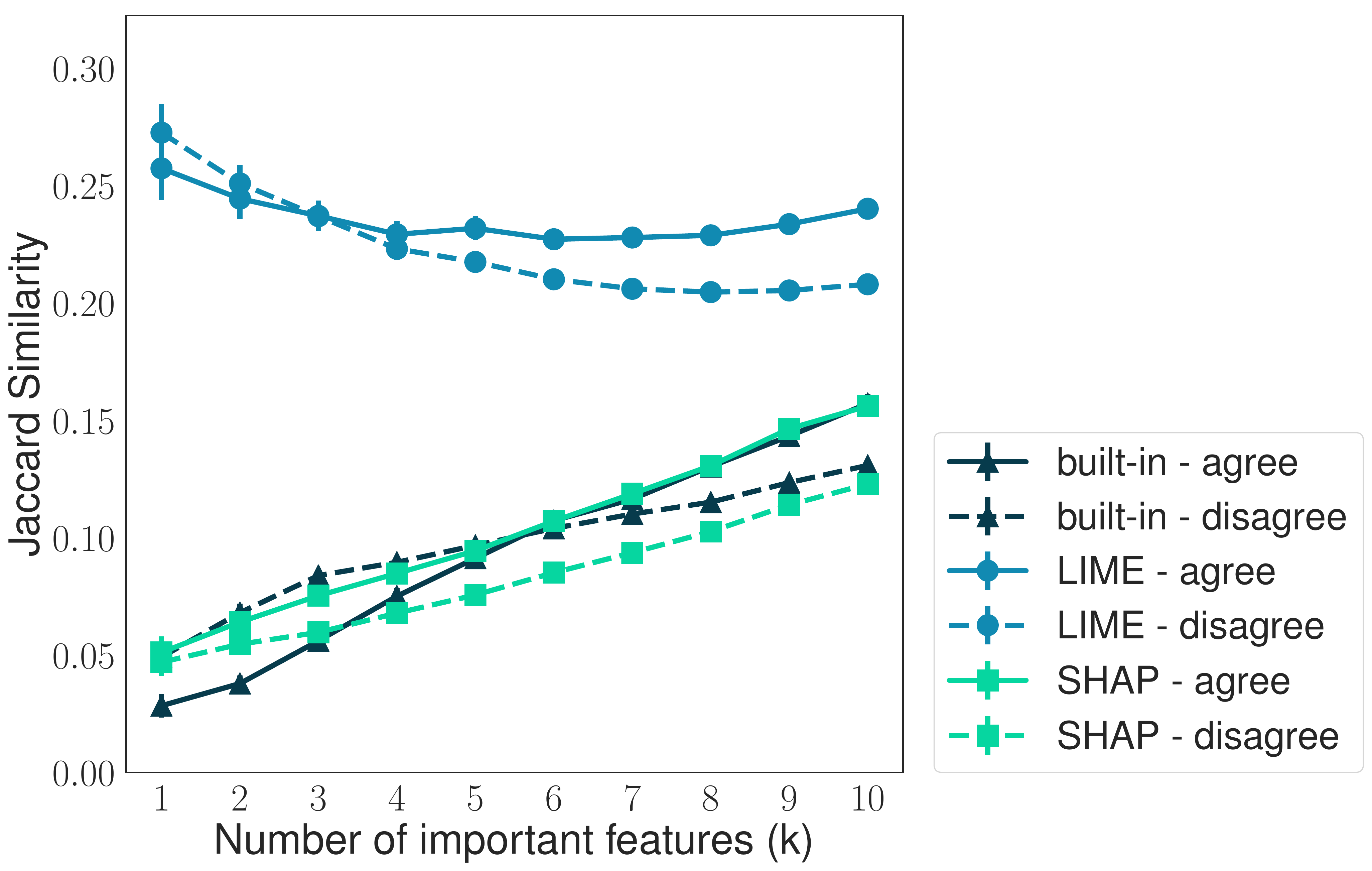}
  \caption{Yelp}
  \label{fig:yelp_xgb_lstm_att_pred}
\end{subfigure}
\hfill
\begin{subfigure}[t]{0.285\textwidth}
  \centering
  \includegraphics[trim=0 0 5.5in 0,clip,width=\textwidth]{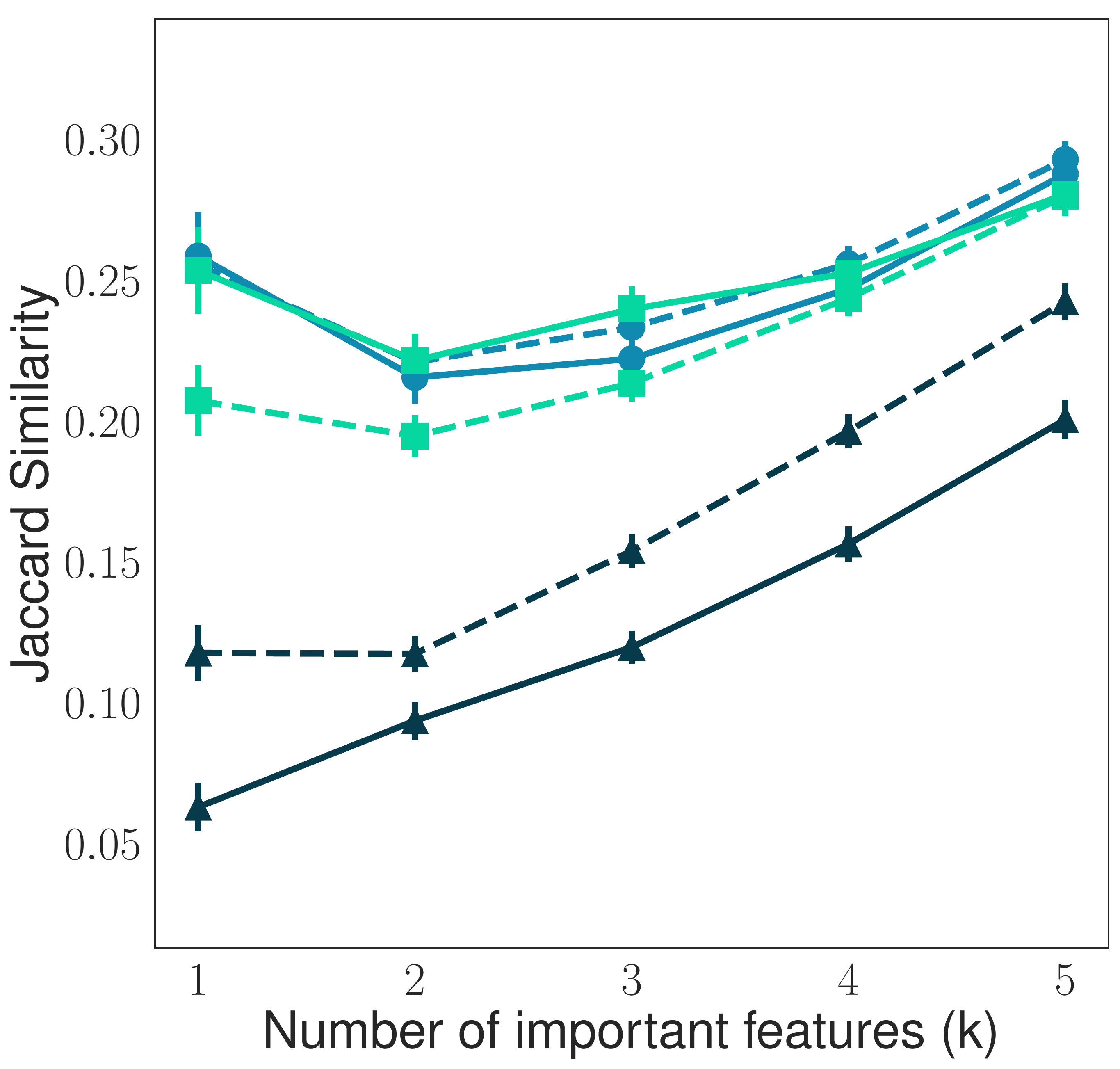}
  \caption{SST}
  \label{fig:sst_xgb_lstm_att_pred}
\end{subfigure}
\hfill
\begin{subfigure}[t]{0.28\textwidth}
  \centering
  \includegraphics[trim=0 0 5.5in 0,clip,width=\textwidth]{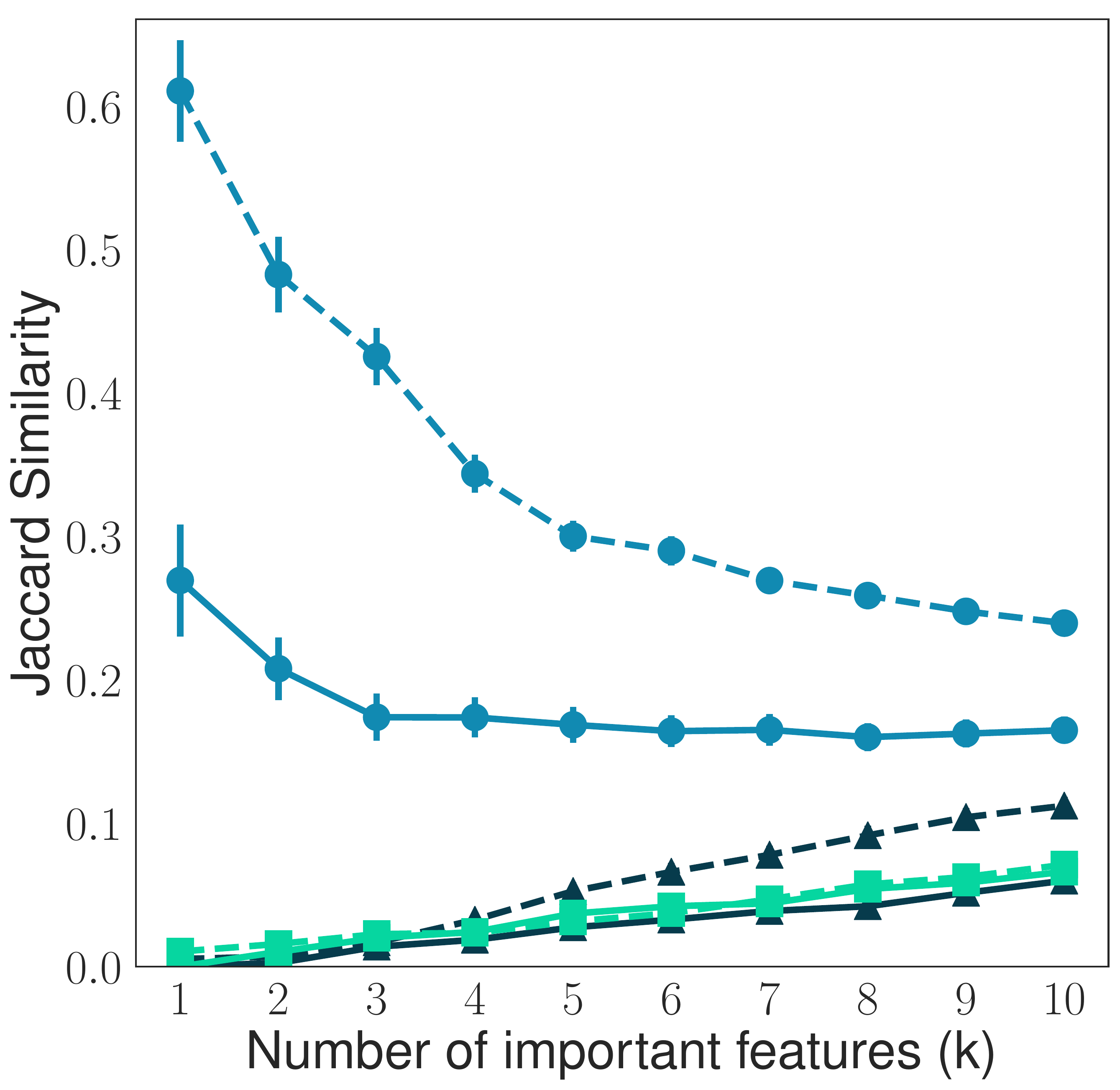}
  \caption{Deception}
  \label{fig:deception_xgb_lstm_att_pred}
\end{subfigure}
\caption{
Similarity between two models is not necessarily greater when they agree on the predictions, and in some scenarios,
e.g.,
SVM ($\ell_1$) x XGB with LIME method, XGB x BERT with LIME method, and XGB x BERT with built-in method, they are sometimes lower than when they disagree on the predicted labels.
}
\label{fig:prediction_supp_2}
\end{figure*}

\para{Similarity vs. predicted labels.}
Similarity is not necessarily higher when predictions agree, it is also not necessarily lower when predictions disagree.
See \figref{fig:prediction_supp} and \figref{fig:prediction_supp_2}.

\begin{figure*}[t!]
\centering
Similarity between different models based on the same method\\
\begin{subfigure}[t]{0.425\textwidth}
  \centering
  \includegraphics[width=\textwidth]{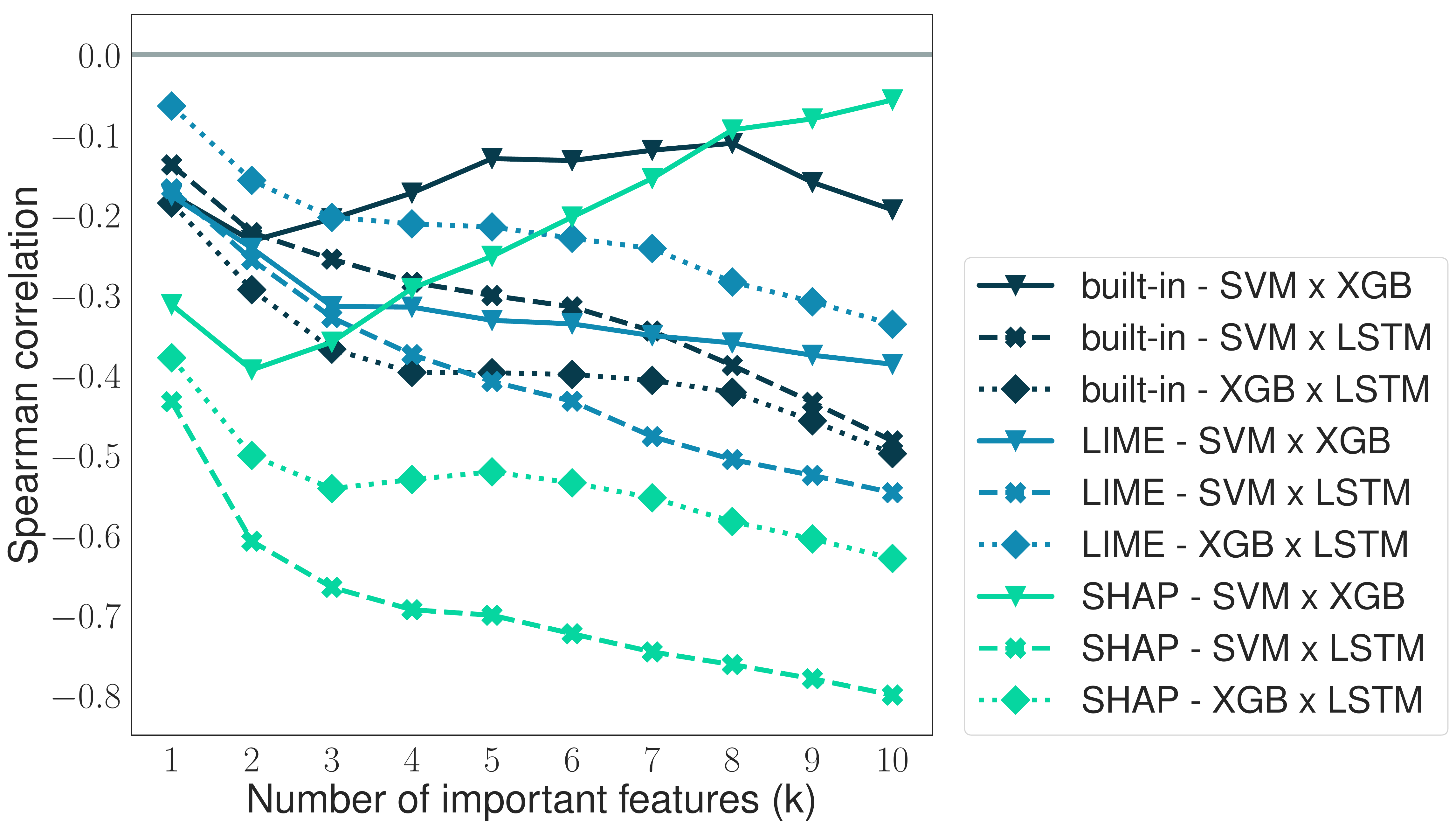}
  \caption{Yelp}
  \label{fig:yelp_methods_length}
\end{subfigure}
\hfill
\begin{subfigure}[t]{0.28\textwidth}
  \centering
  \includegraphics[trim=0 0 6.5in 0,clip,width=\textwidth]{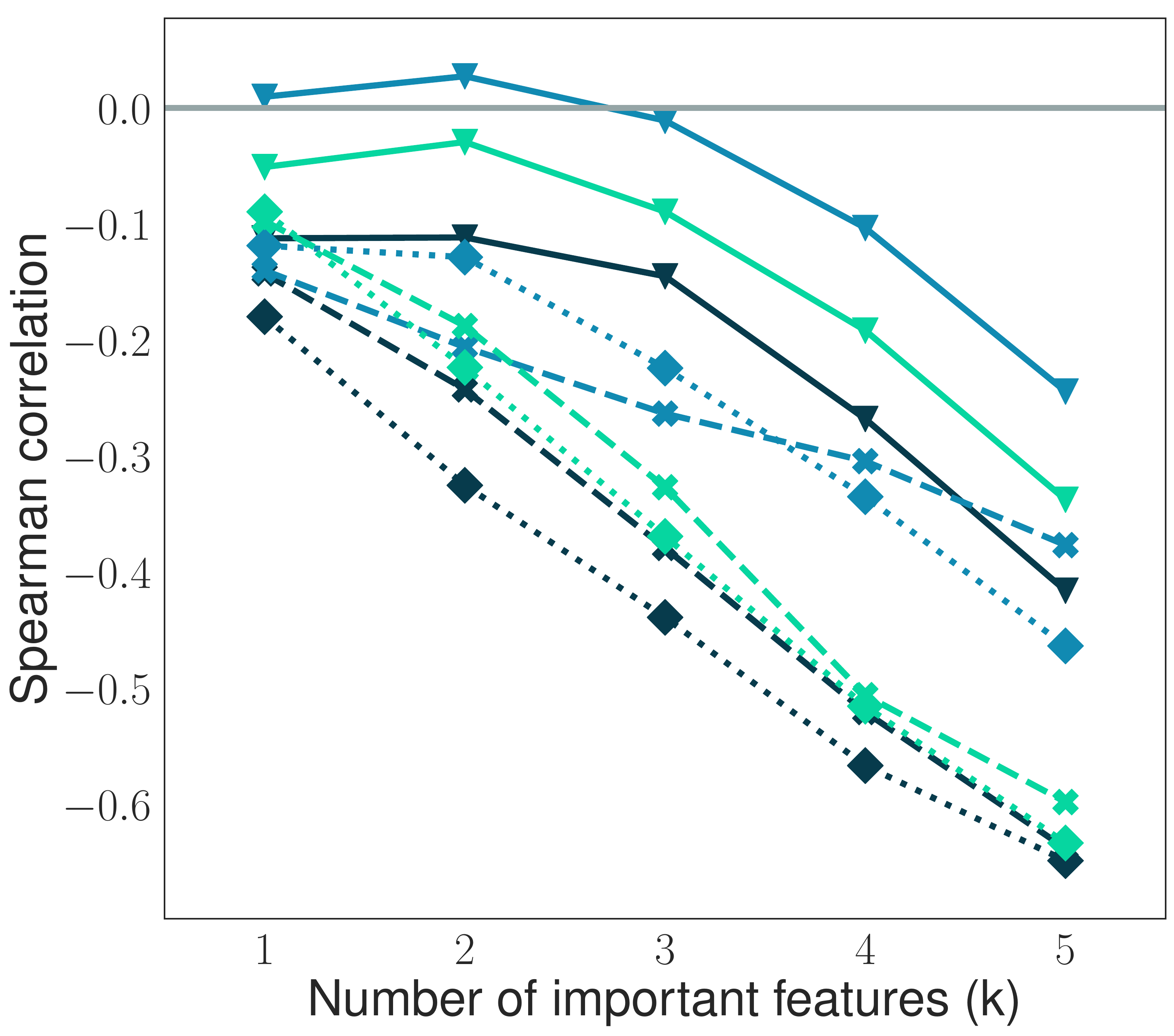}
  \caption{SST}
  \label{fig:sst_methods_length}
\end{subfigure}
\hfill
\begin{subfigure}[t]{0.28\textwidth}
  \centering
  \includegraphics[trim=0 0 6.5in 0,clip,width=\textwidth]{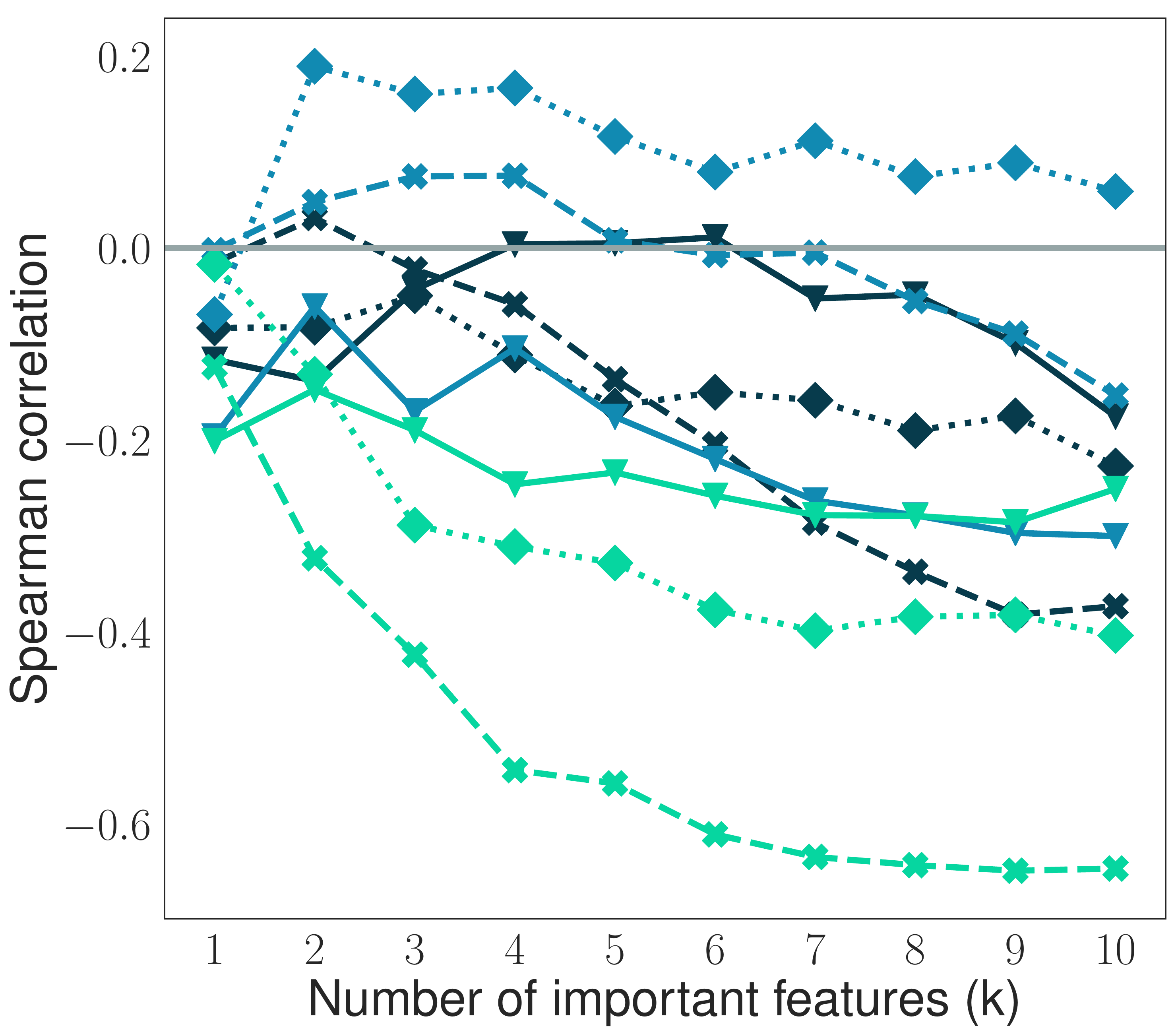}
  \caption{Deception}
  \label{fig:deception_methods_length}
\end{subfigure}\\
Similarity between different models based on the same method for BERT\\
\begin{subfigure}[t]{0.425\textwidth}
  \centering
  \includegraphics[width=\textwidth]{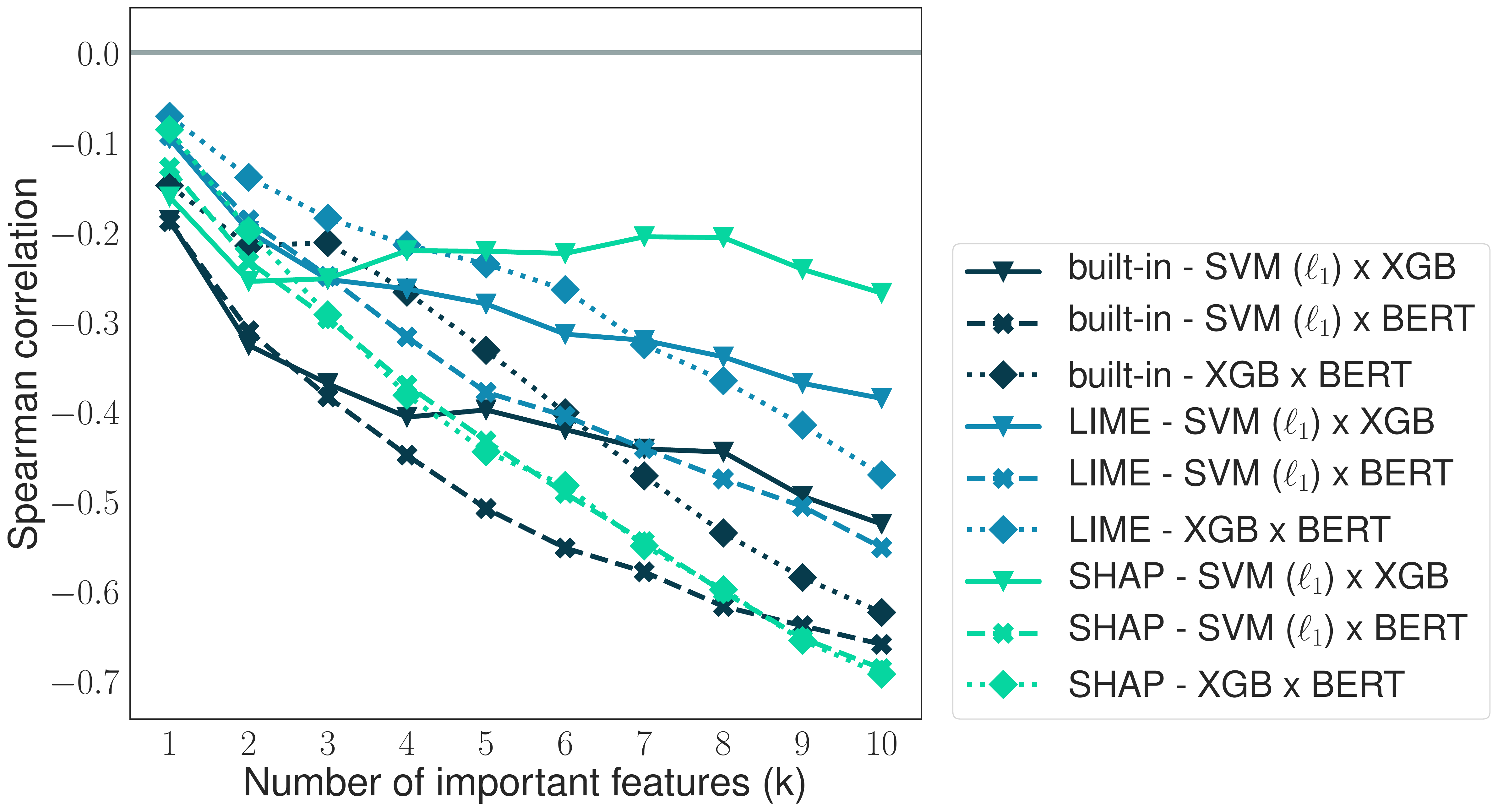}
  \caption{Yelp}
  \label{fig:yelp_methods_length}
\end{subfigure}
\hfill
\begin{subfigure}[t]{0.28\textwidth}
  \centering
  \includegraphics[trim=0 0 7in 0,clip,width=\textwidth]{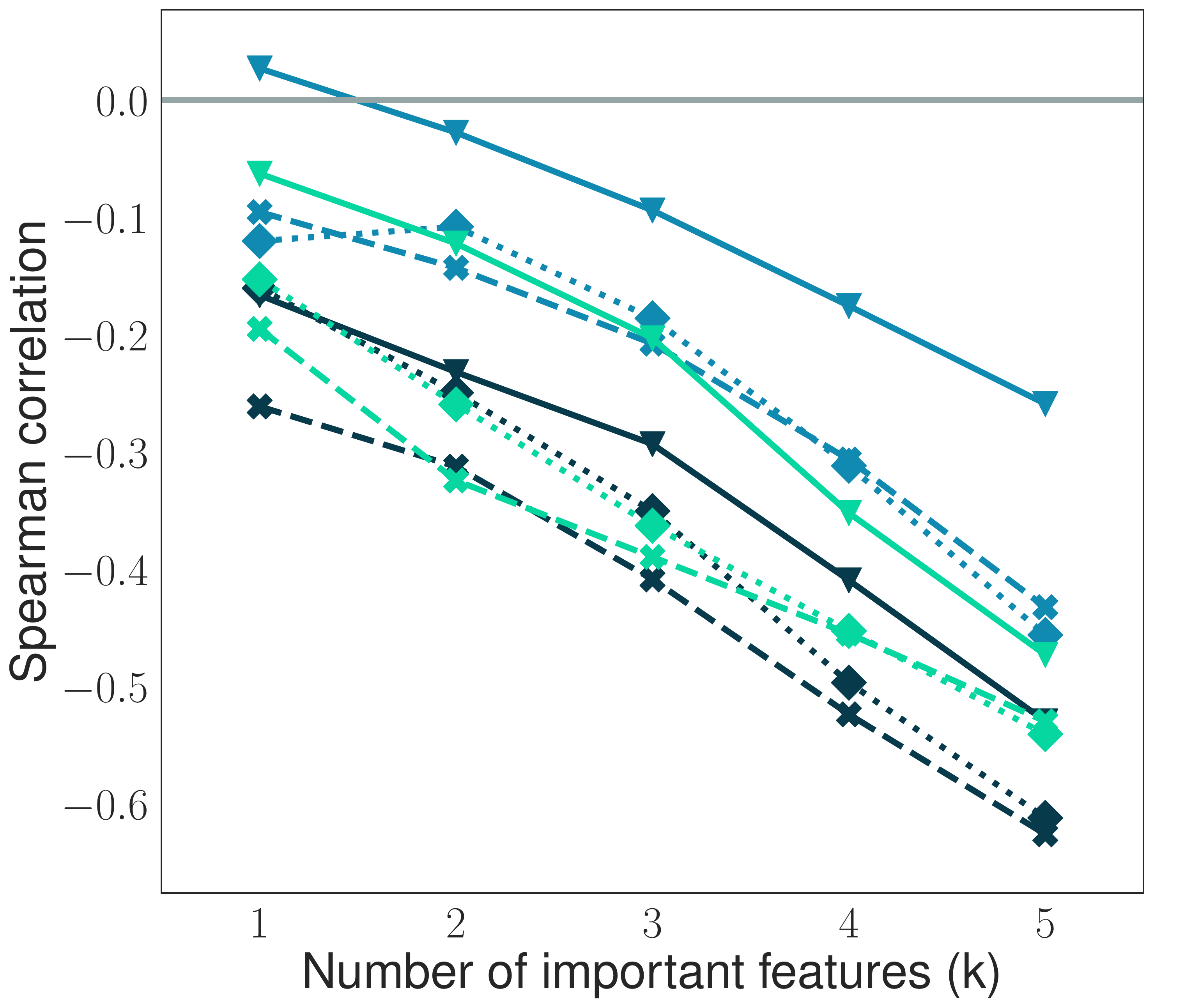}
  \caption{SST}
  \label{fig:sst_methods_length}
\end{subfigure}
\hfill
\begin{subfigure}[t]{0.28\textwidth}
  \centering
  \includegraphics[trim=0 0 7in 0,clip,width=\textwidth]{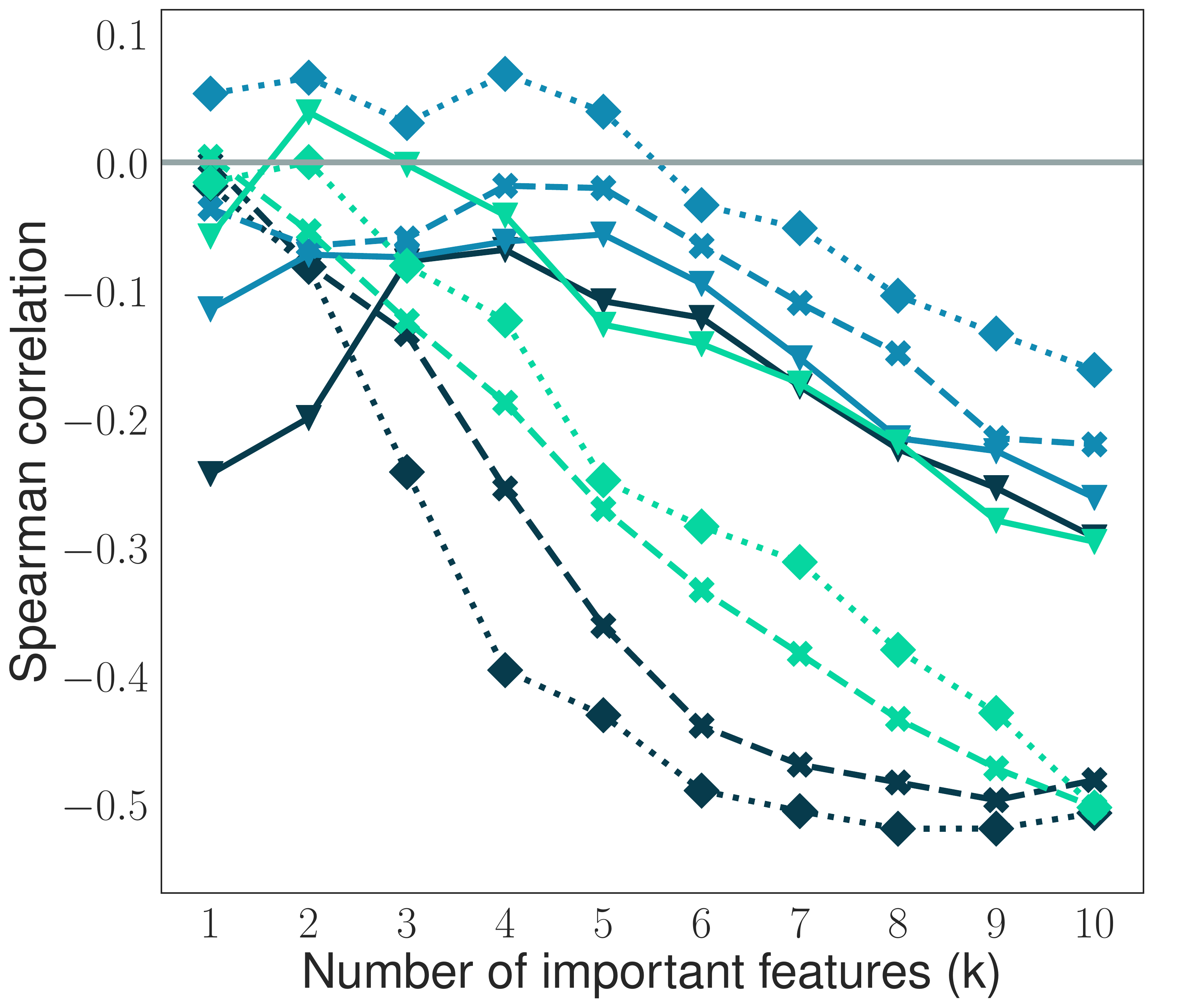}
  \caption{Deception}
  \label{fig:deception_methods_length}
\end{subfigure}\\
Similarity between different methods based on the same model for BERT\\
\begin{subfigure}[t]{0.425\textwidth}
  \centering
  \includegraphics[width=\textwidth]{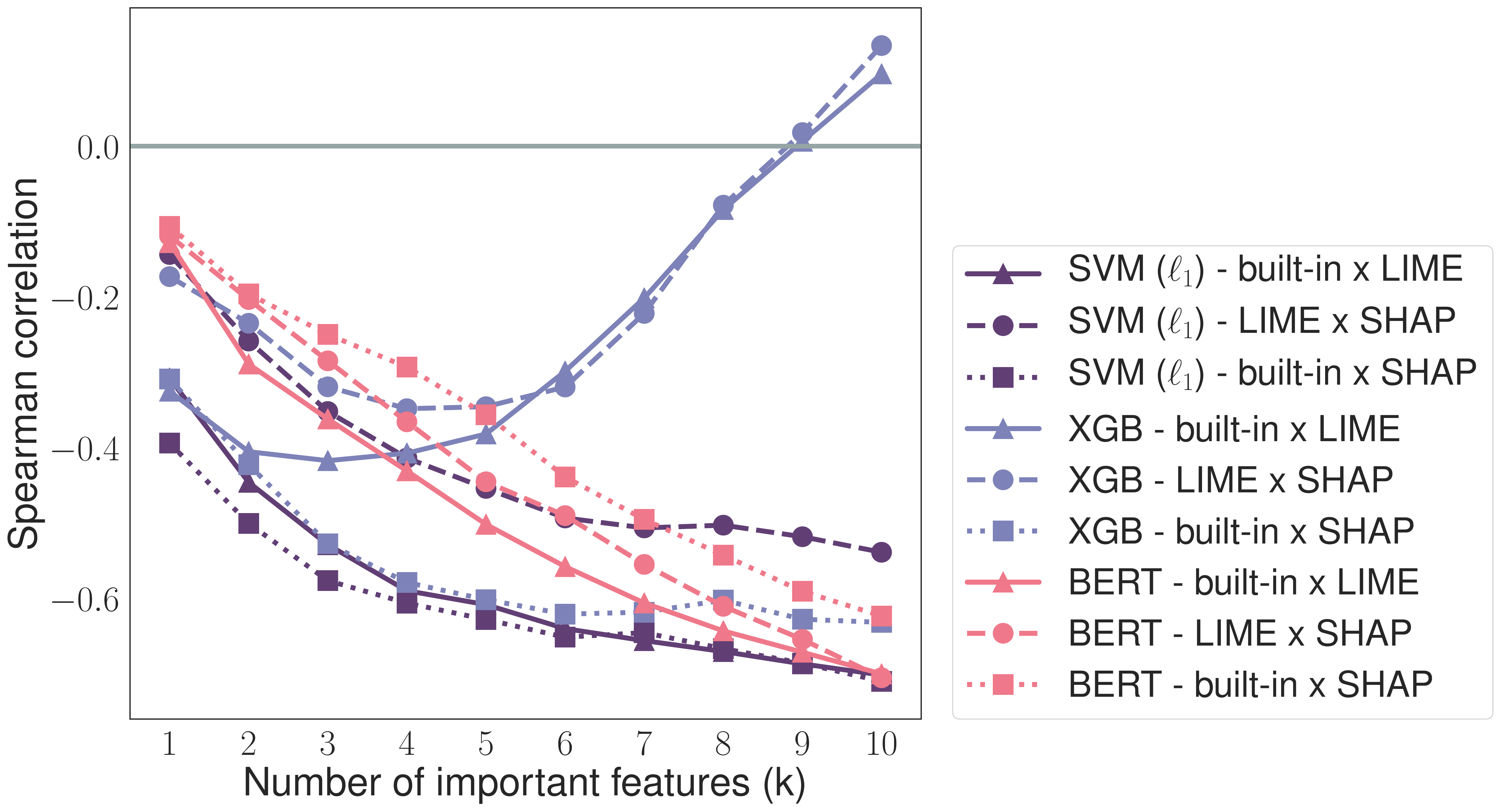}
  \caption{Yelp}
  \label{fig:yelp_models_length}
\end{subfigure}
\hfill
\begin{subfigure}[t]{0.28\textwidth}
  \centering
  \includegraphics[trim=0 0 7in 0,clip,width=\textwidth]{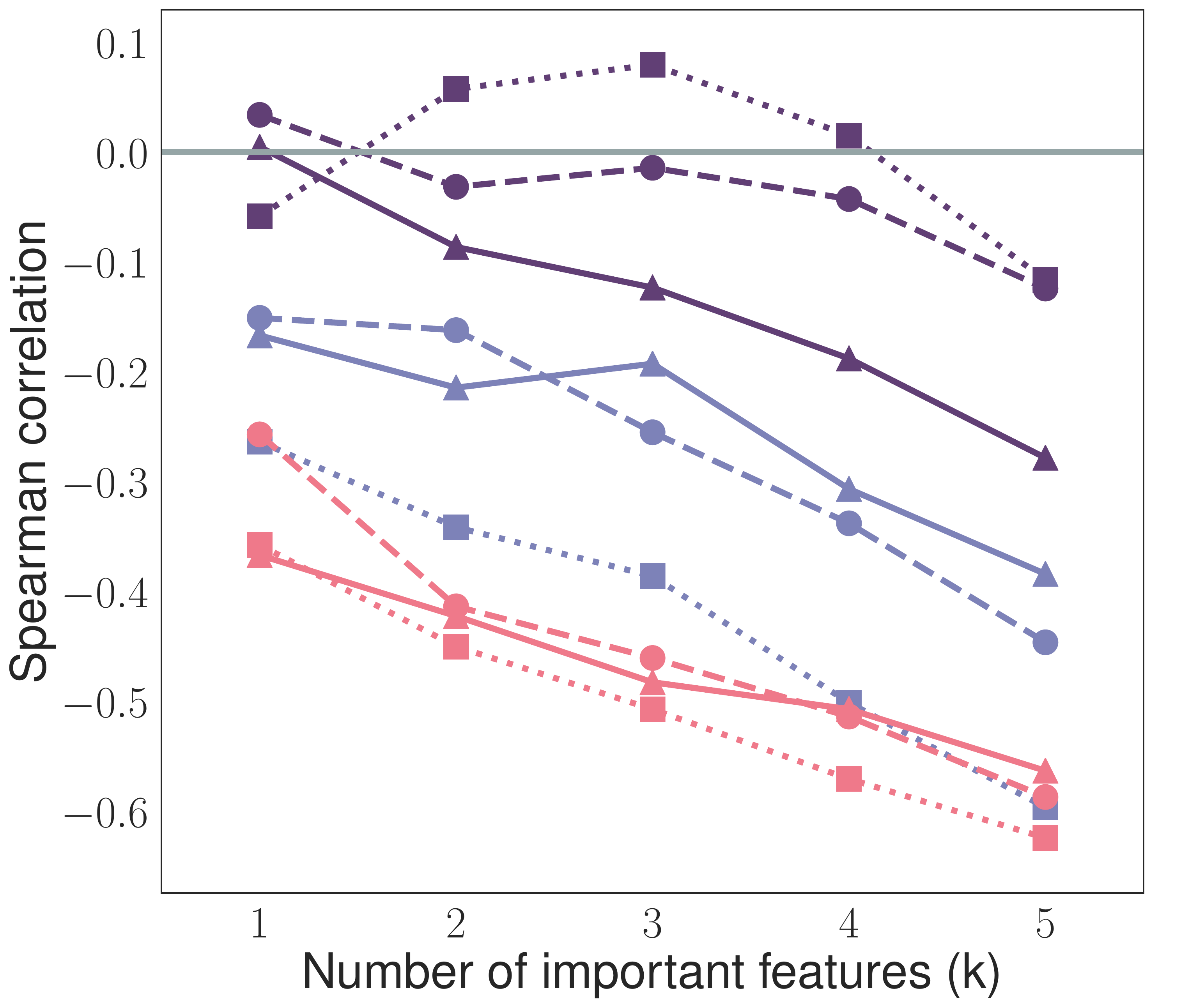}
  \caption{SST}
  \label{fig:sst_models_length}
\end{subfigure}
\hfill
\begin{subfigure}[t]{0.28\textwidth}
  \centering
  \includegraphics[trim=0 0 7in 0,clip,width=\textwidth]{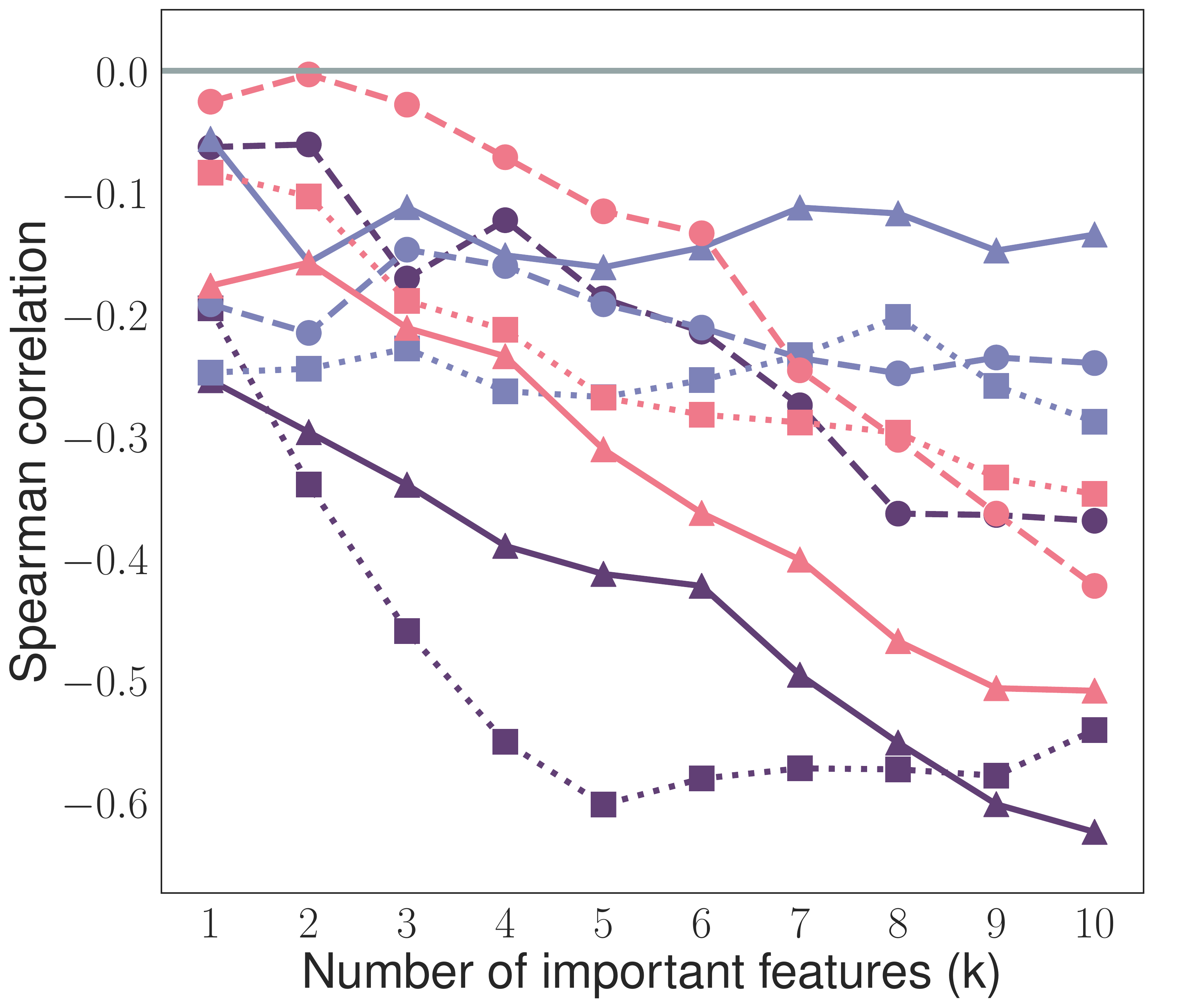}
  \caption{Deception}
  \label{fig:deception_models_length}
\end{subfigure}
\caption{Similarity comparison vs. length. 
The longer the length of an instance, the less similar the important features are. 
The negative correlation becomes stronger as $k$ grows. 
In certain scenarios,
e.g.,
XGB - built-in x LIME and XGB - LIME x SHAP, correlation occasionally goes above 0.
}
\label{fig:length_supp}
\end{figure*}

\para{Similarity vs. length.}
The negative correlation between length and similarity grows stronger as $k$ grows.
See \figref{fig:length_supp}.

\begin{figure*}[t]
\centering
Comparison between models using the same method \\
\begin{subfigure}[t]{0.43\textwidth}
  \centering
  \includegraphics[width=\textwidth]{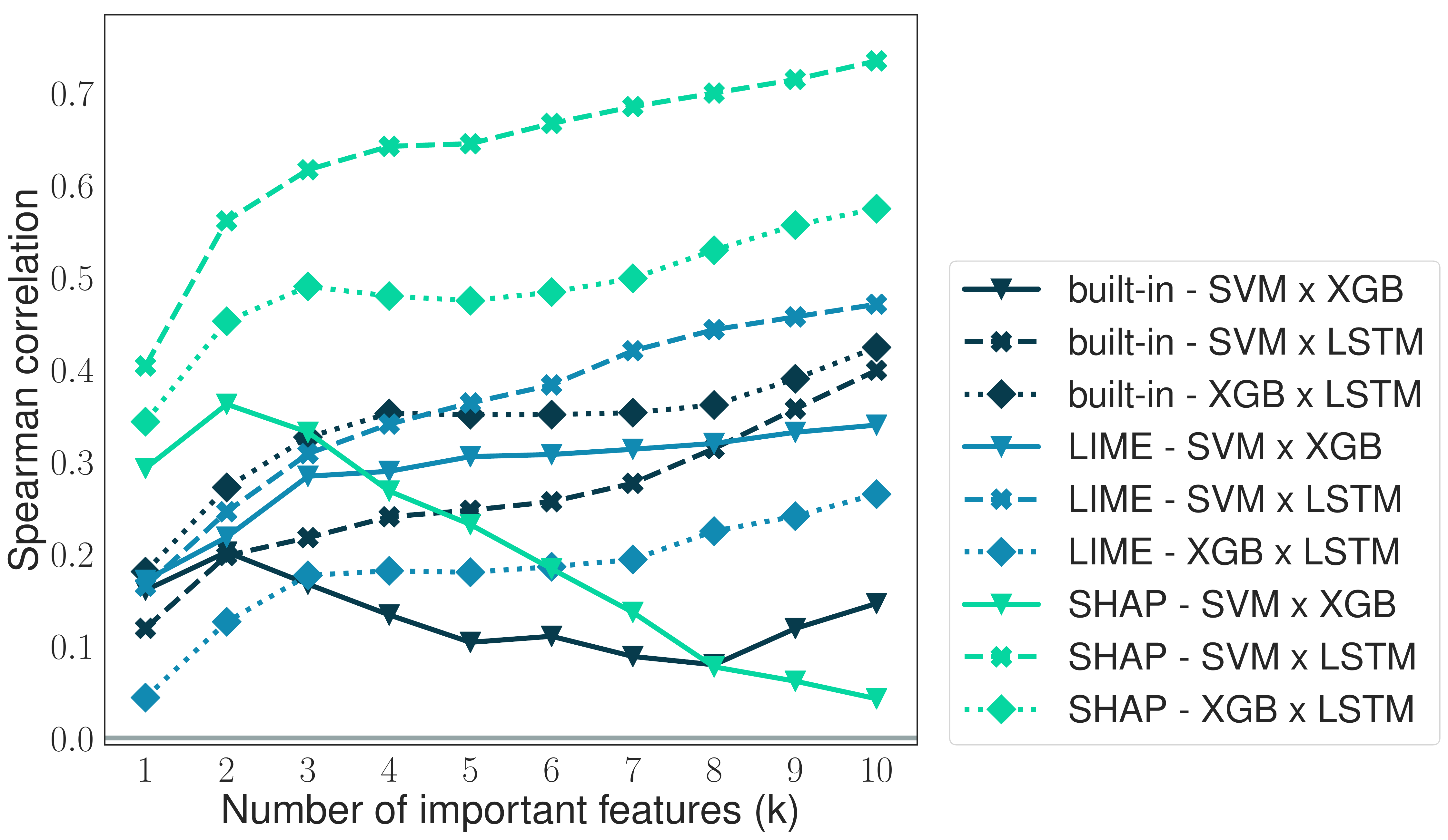}
  \caption{Yelp}
  \label{fig:yelp_methods_ratio}
\end{subfigure}
\hfill
\begin{subfigure}[t]{0.27\textwidth}
  \centering
  \includegraphics[trim=0 0 6.5in 0,clip,width=\textwidth]{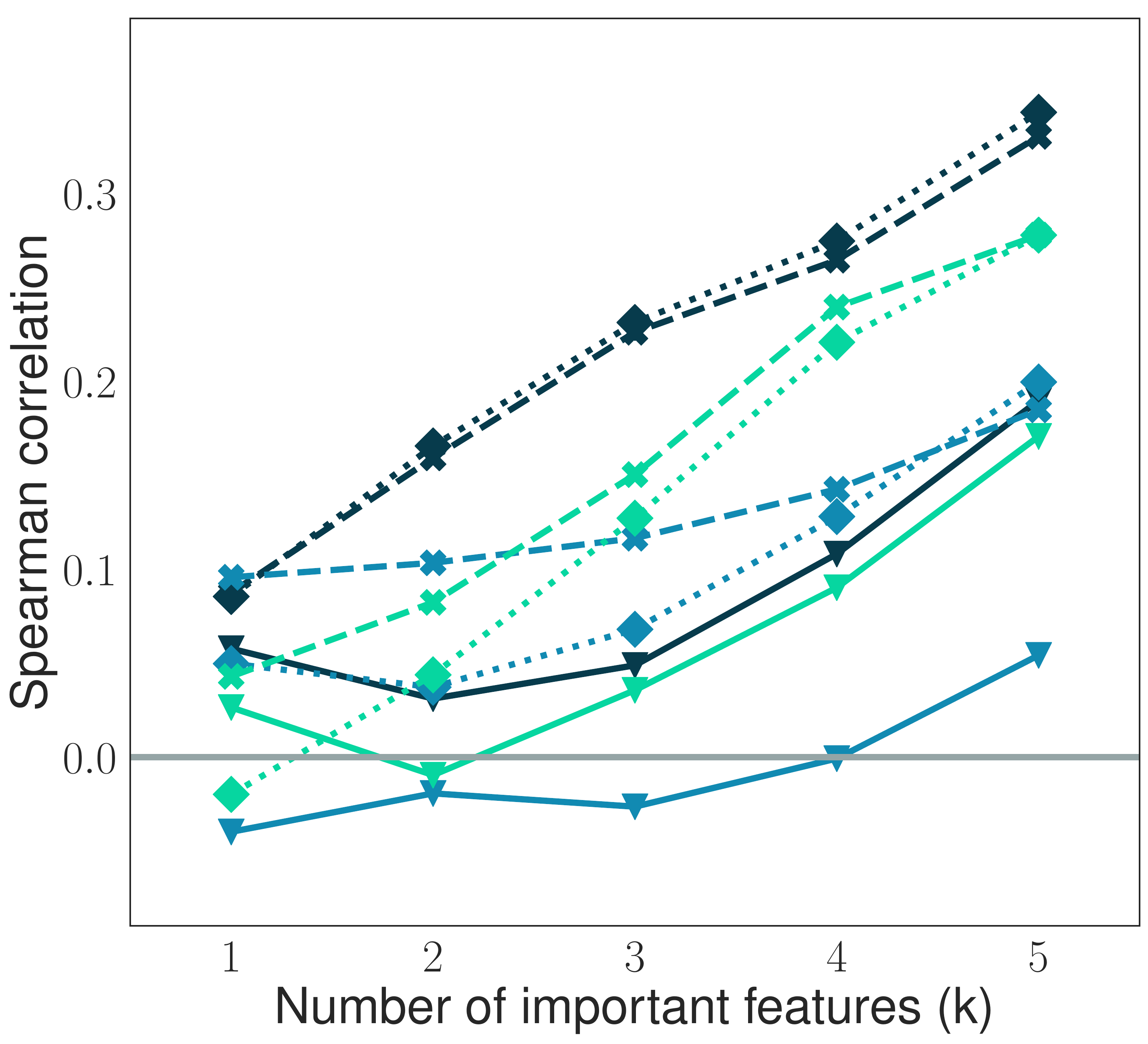}
  \caption{SST}
  \label{fig:sst_methods_ratio}
\end{subfigure}
\hfill
\begin{subfigure}[t]{0.28\textwidth}
  \centering
  \includegraphics[trim=0 0 6.5in 0,clip,width=\textwidth]{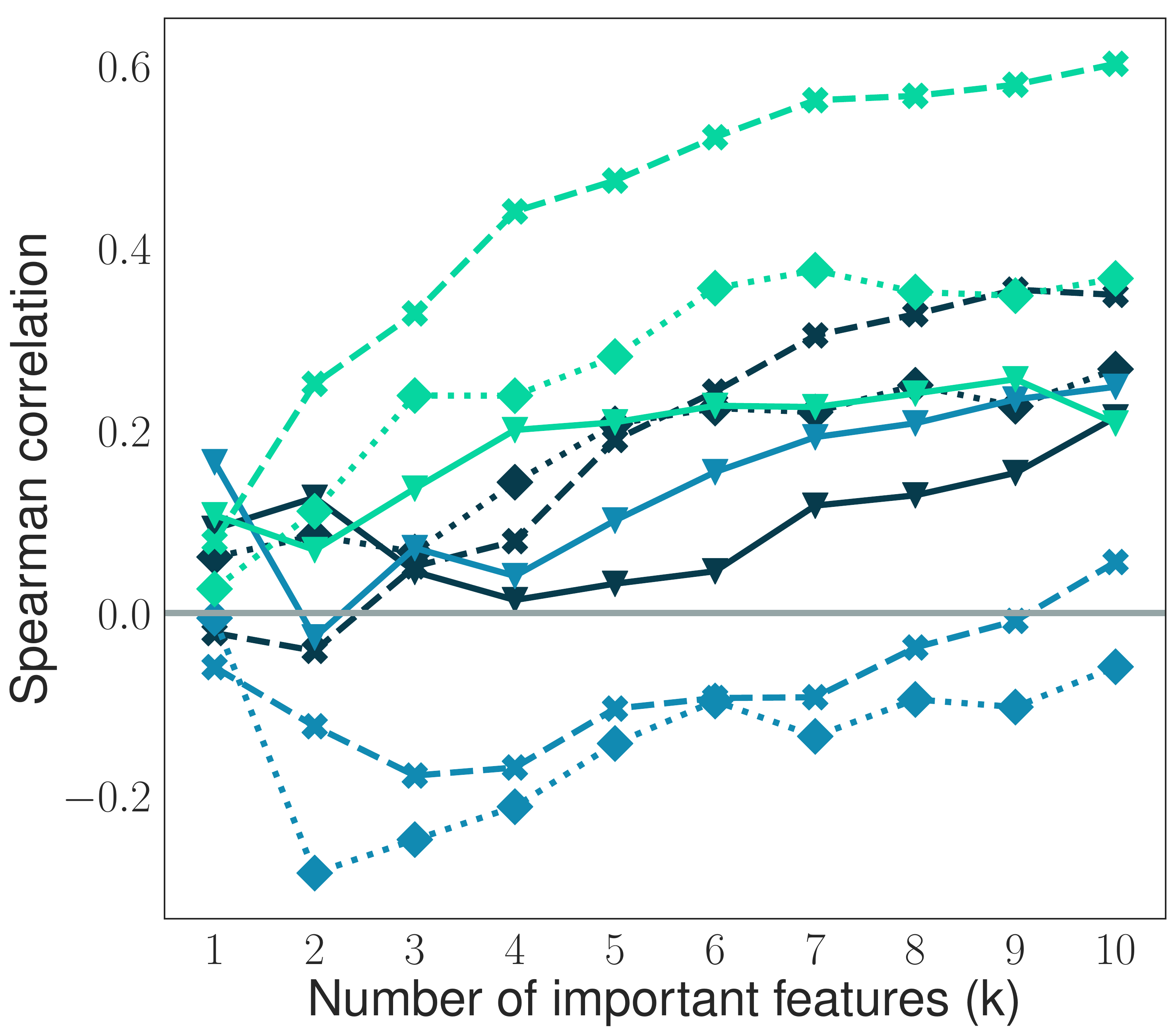}
  \caption{Deception}
  \label{fig:deception_methods_ratio}
\end{subfigure}
\begin{subfigure}[t]{0.43\textwidth}
  \centering
  \includegraphics[width=\textwidth]{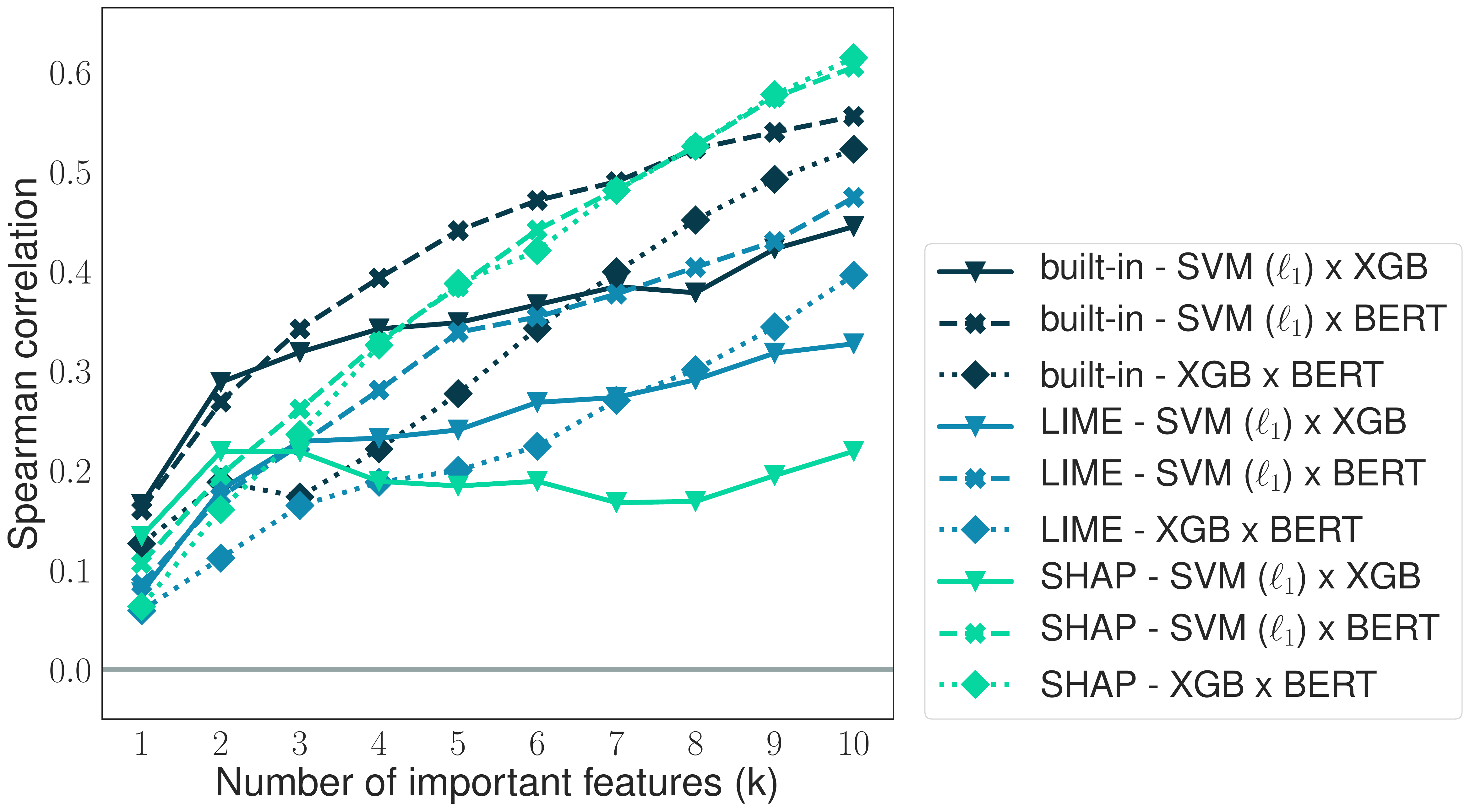}
  \caption{Yelp}
  \label{fig:yelp_methods_ratio}
\end{subfigure}
\hfill
\begin{subfigure}[t]{0.27\textwidth}
  \centering
  \includegraphics[trim=0 0 7.1in 0,clip,width=\textwidth]{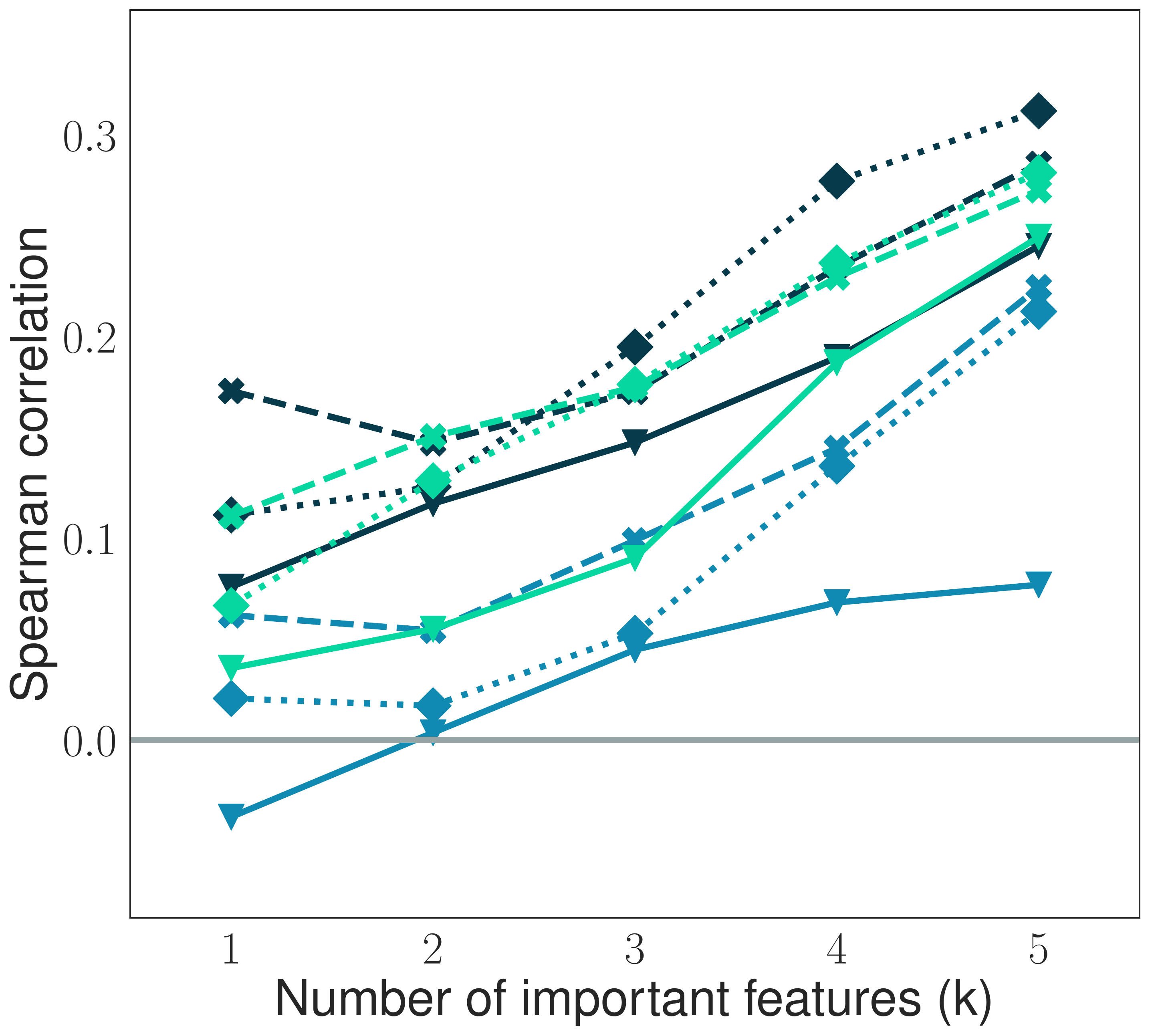}
  \caption{SST}
  \label{fig:sst_methods_ratio}
\end{subfigure}
\hfill
\begin{subfigure}[t]{0.28\textwidth}
  \centering
  \includegraphics[trim=0 0 7.1in 0,clip,width=\textwidth]{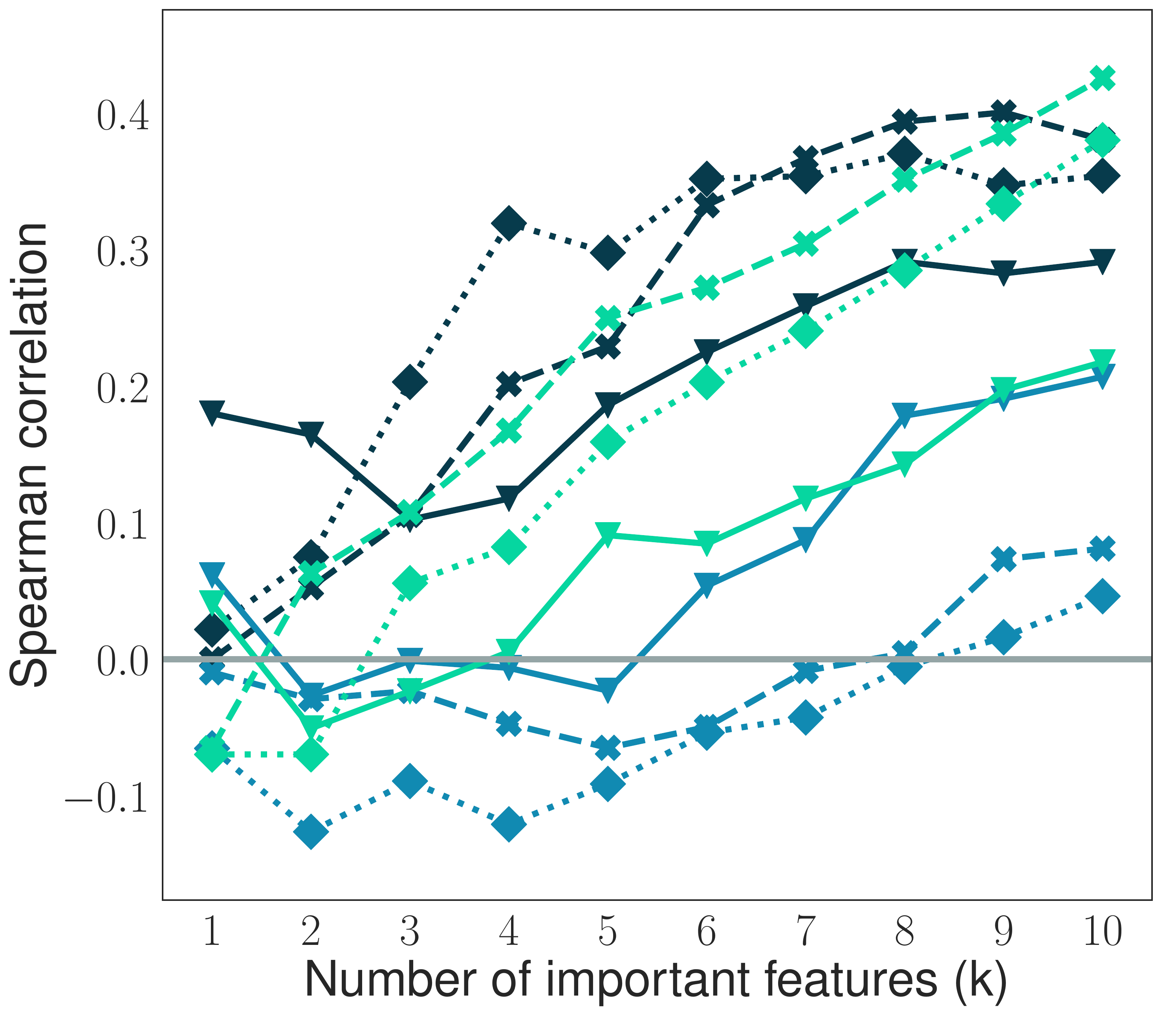}
  \caption{Deception}
  \label{fig:deception_methods_ratio}
\end{subfigure}
\caption{Similarity comparison vs. type-token ratio. 
The higher the type-token ratio, the more similar the important features are. 
The positive correlation becomes stronger as $k$ grows. 
In some cases,
e.g.,
LIME method on deception dataset, correlation becomes weaker as $k$ grows.
}
\label{fig:ratio}
\end{figure*}

\begin{figure*}[t]
\centering
Similarity comparison between methods using the same model\\
\begin{subfigure}[t]{0.43\textwidth}
  \centering
  \includegraphics[width=\textwidth]{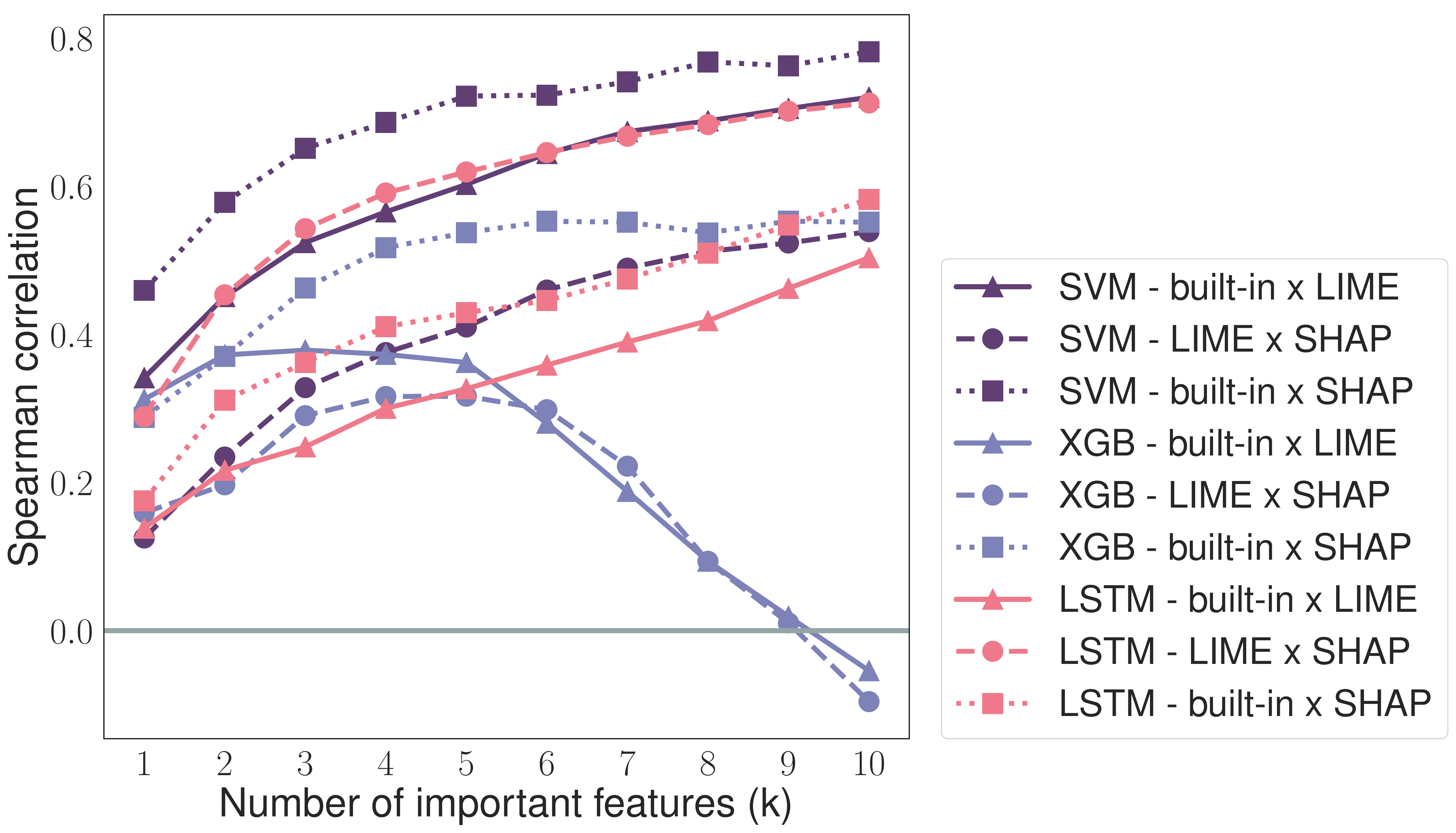}
  \caption{Yelp}
  \label{fig:yelp_models_ratio}
\end{subfigure}
\hfill
\begin{subfigure}[t]{0.275\textwidth}
  \centering
  \includegraphics[trim=0 0 6.5in 0,clip,width=\textwidth]{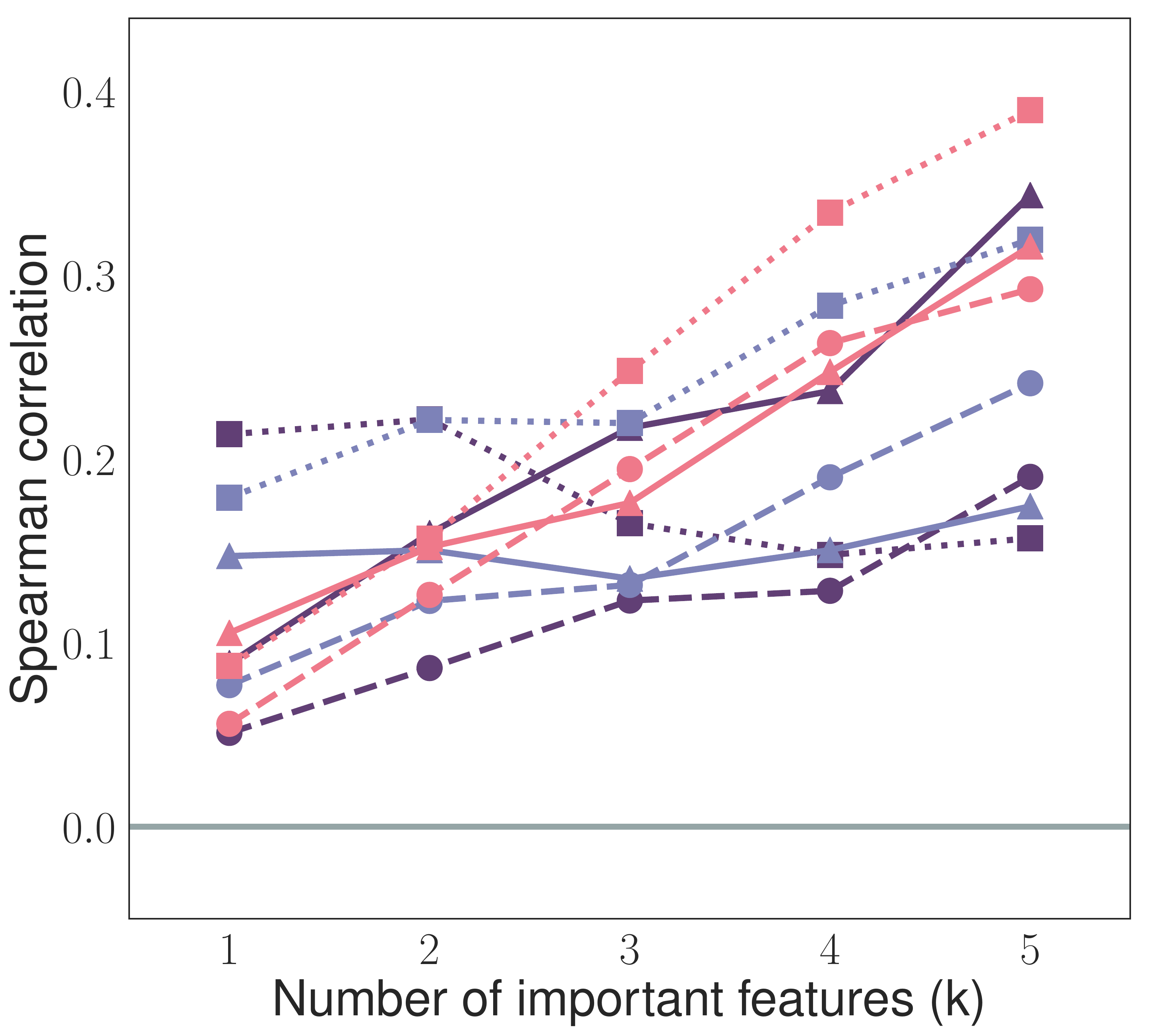}
  \caption{SST}
  \label{fig:sst_models_ratio}
\end{subfigure}
\hfill
\begin{subfigure}[t]{0.27\textwidth}
  \centering
  \includegraphics[trim=0 0 6.5in 0,clip,width=\textwidth]{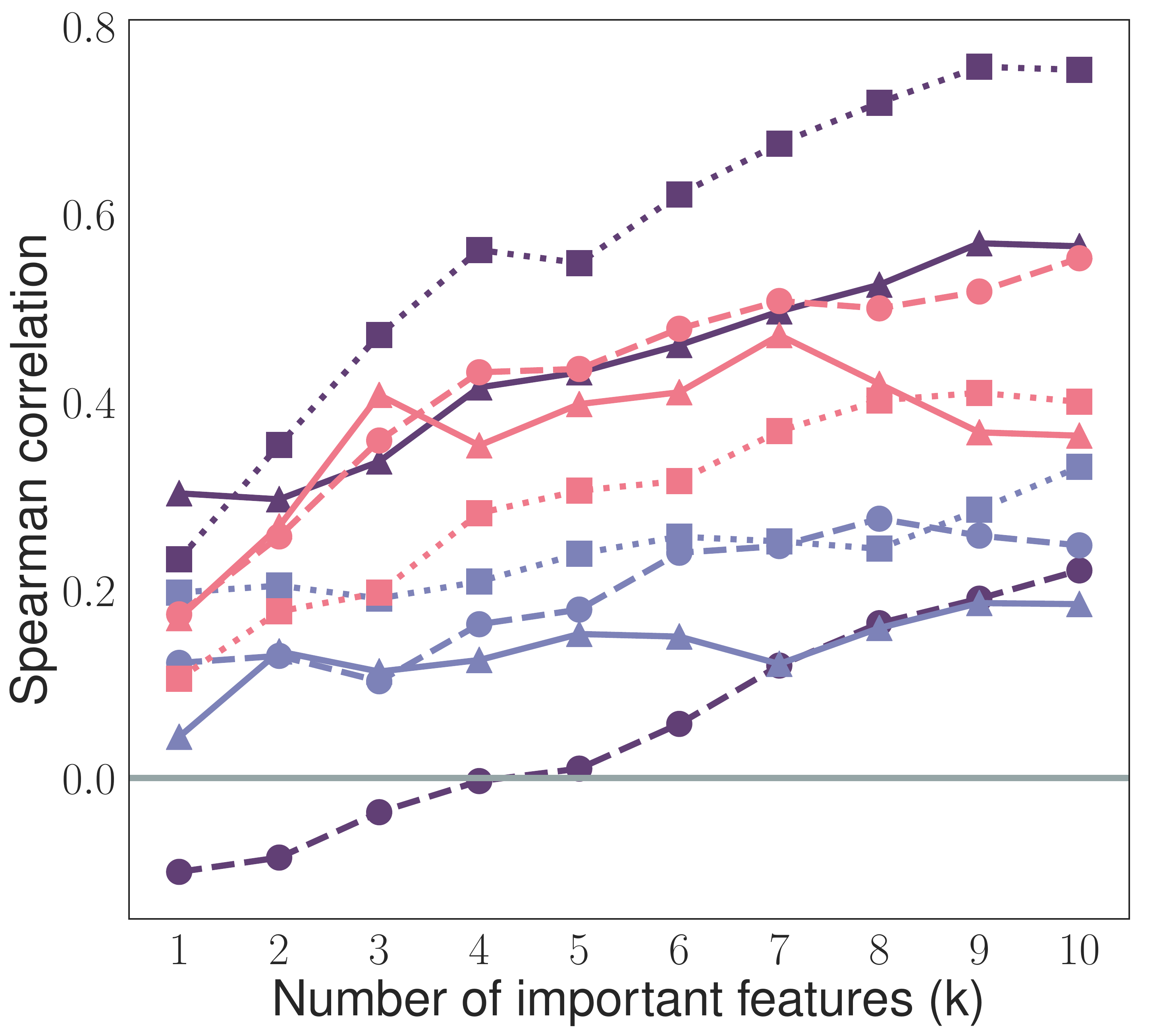}
  \caption{Deception}
  \label{fig:deception_models_ratio}
\end{subfigure}
\begin{subfigure}[t]{0.43\textwidth}
  \centering
  \includegraphics[width=\textwidth]{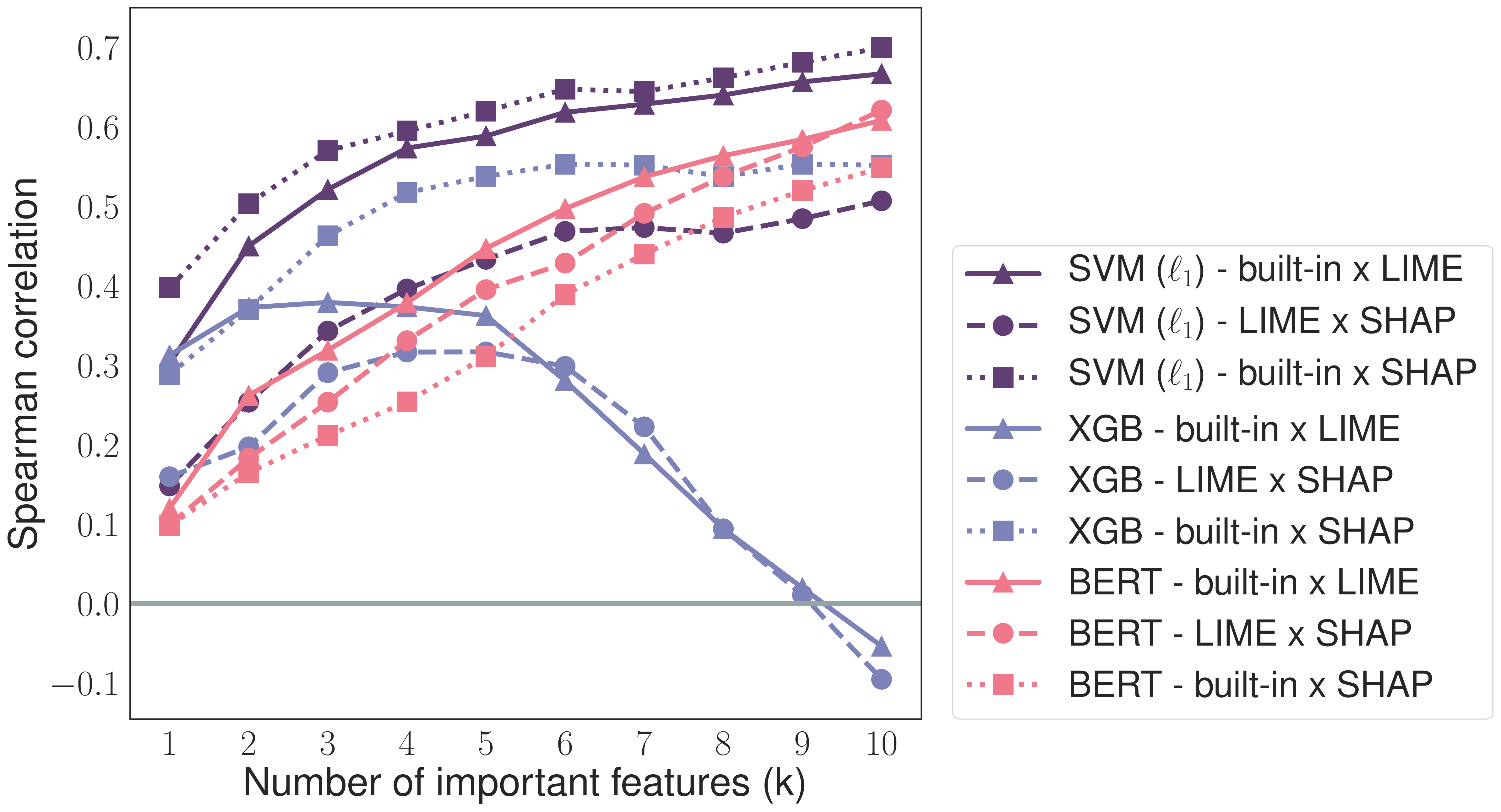}
  \caption{Yelp}
  \label{fig:yelp_models_ratio}
\end{subfigure}
\hfill
\begin{subfigure}[t]{0.283\textwidth}
  \centering
  \includegraphics[trim=0 0 7.1in 0,clip,width=\textwidth]{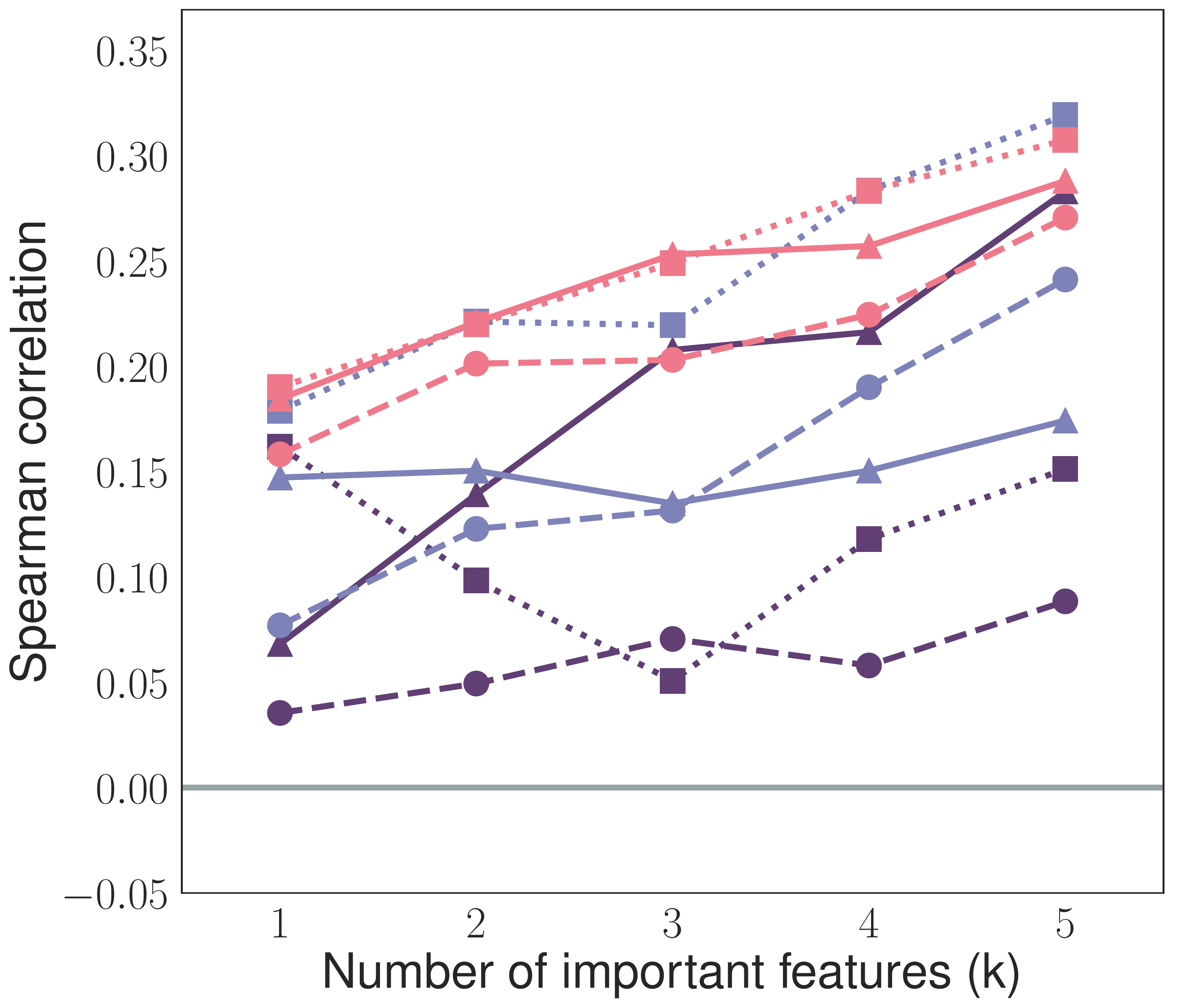}
  \caption{SST}
  \label{fig:sst_models_ratio}
\end{subfigure}
\hfill
\begin{subfigure}[t]{0.27\textwidth}
  \centering
  \includegraphics[trim=0 0 7.1in 0,clip,width=\textwidth]{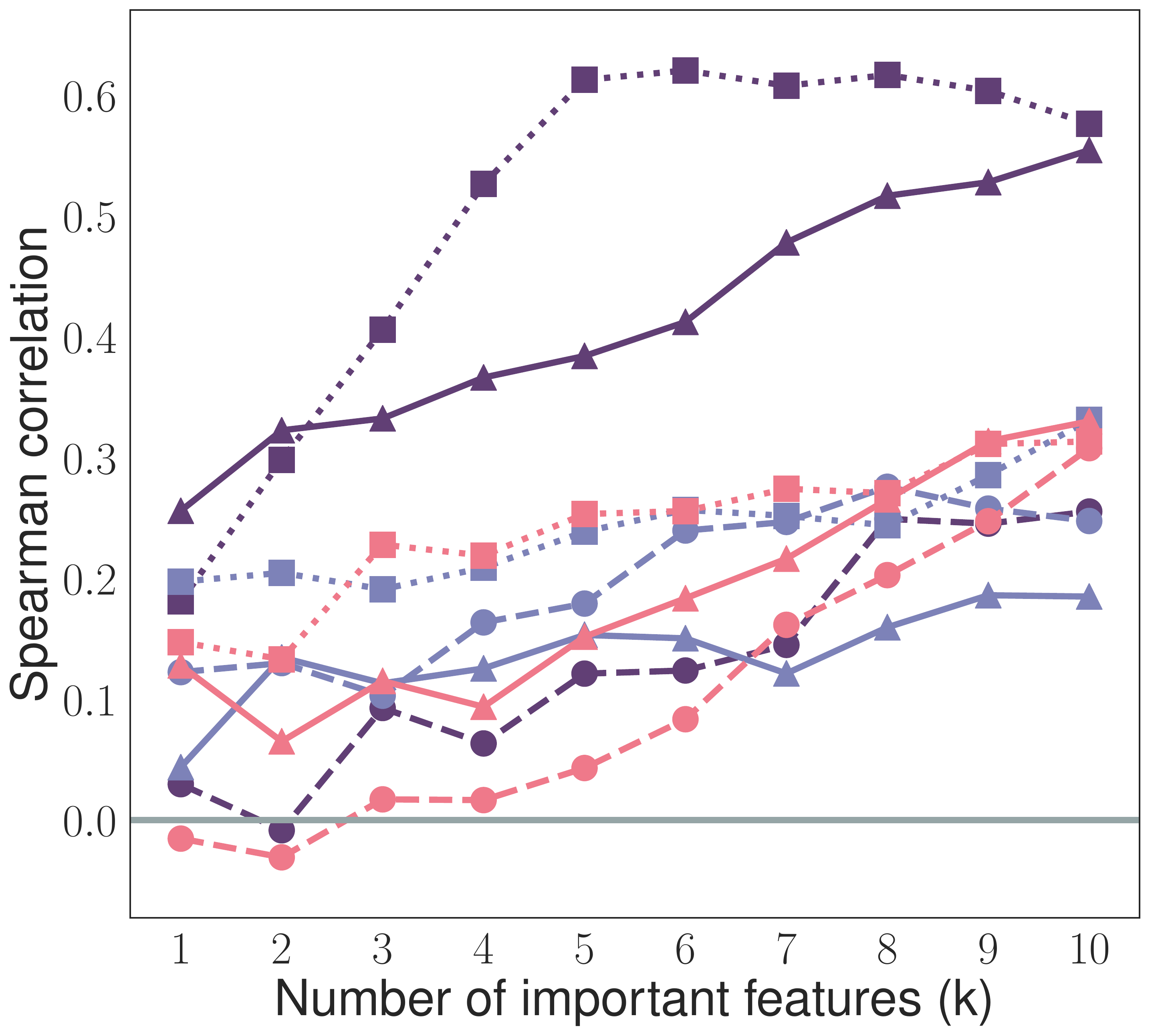}
  \caption{Deception}
  \label{fig:deception_models_ratio}
\end{subfigure}
\caption{Similarity comparison vs. type-token ratio. 
The higher the type-token ratio, the more similar the important features are. 
The positive correlation becomes stronger as $k$ grows. 
In some cases,
e.g.,
 XGB - built-in and LIME and XGB - LIME and SHAP on Yelp dataset, correlation becomes weaker as $k$ grows.
}
\label{fig:ratio_2}
\end{figure*}

\para{Similarity vs. type-token ratio.}
The positive correlation between type-token ratio and similarity grows stronger as $k$ grows.
See \figref{fig:ratio} and \figref{fig:ratio_2}.

\begin{figure*}[t]
\centering
\begin{subfigure}[t]{0.4\textwidth}
  \centering
  \includegraphics[width=\textwidth]{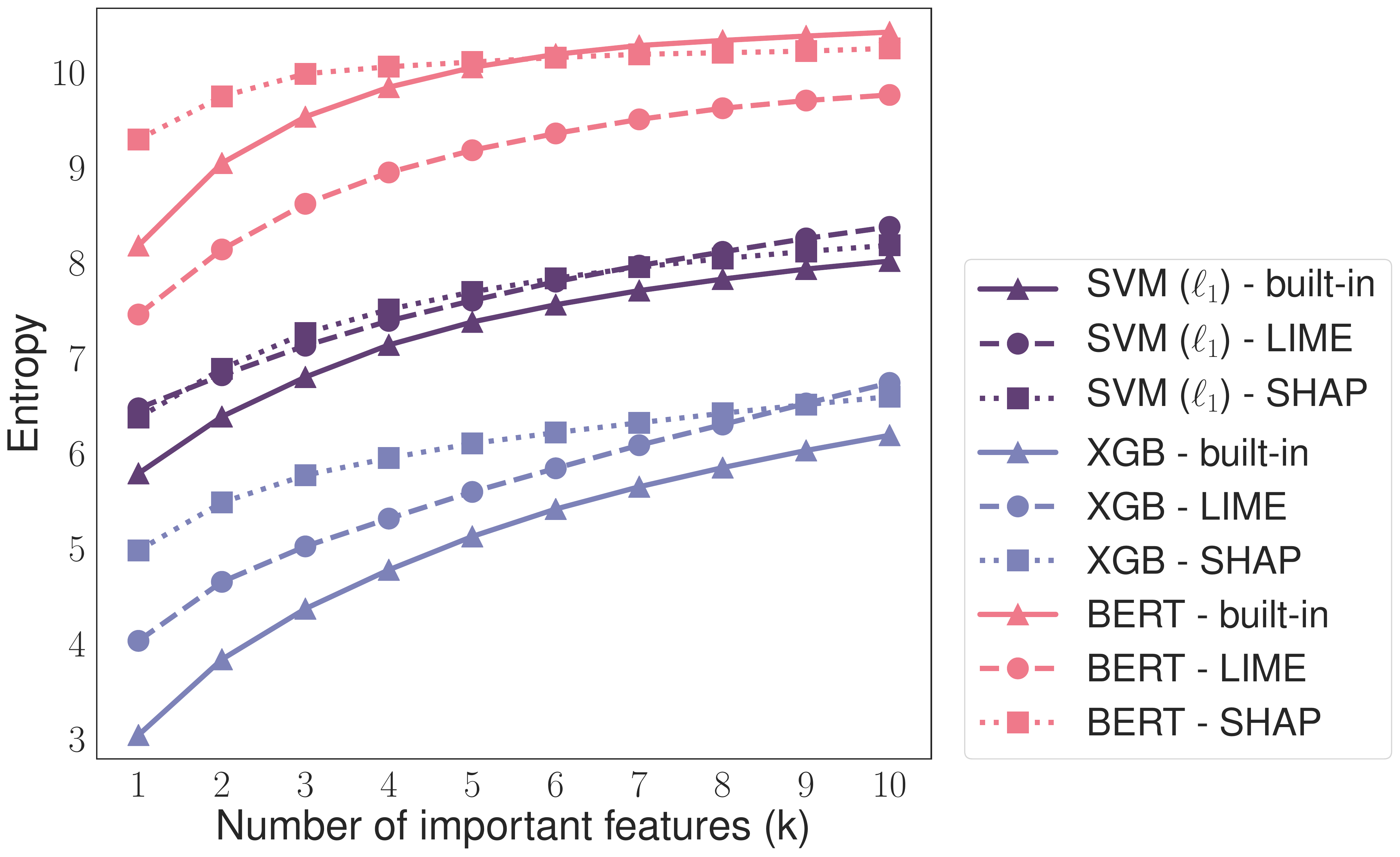}
  \caption{Yelp}
  \label{fig:yelp_models_entropy}
\end{subfigure}
\hfill
\begin{subfigure}[t]{0.275\textwidth}
  \centering
  \includegraphics[trim=0 0 5.5in 0,clip,width=\textwidth]{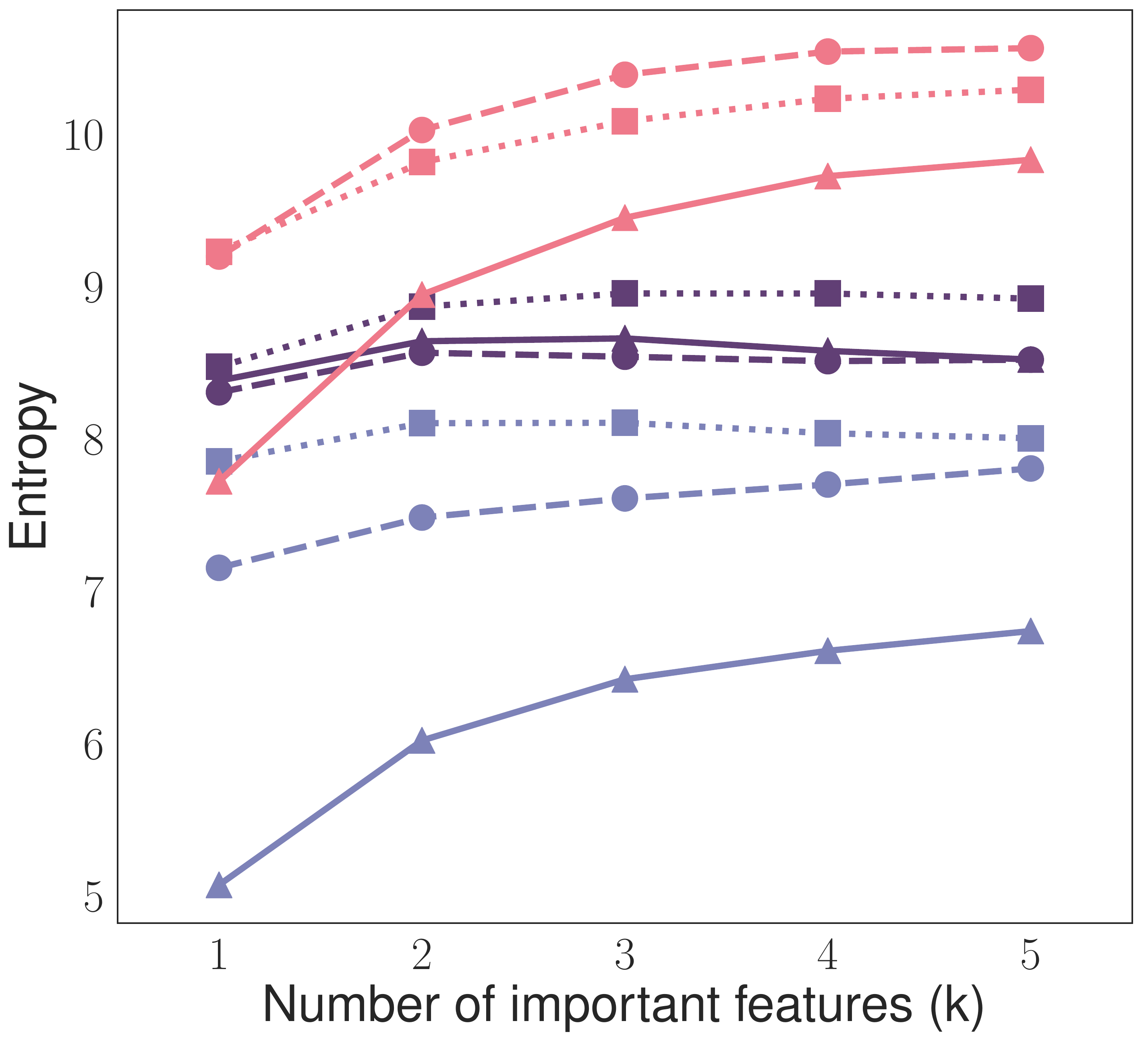}
  \caption{SST}
  \label{fig:sst_models_entropy}
\end{subfigure}
\hfill
\begin{subfigure}[t]{0.27\textwidth}
  \centering
  \includegraphics[trim=0 0 5.5in 0,clip,width=\textwidth]{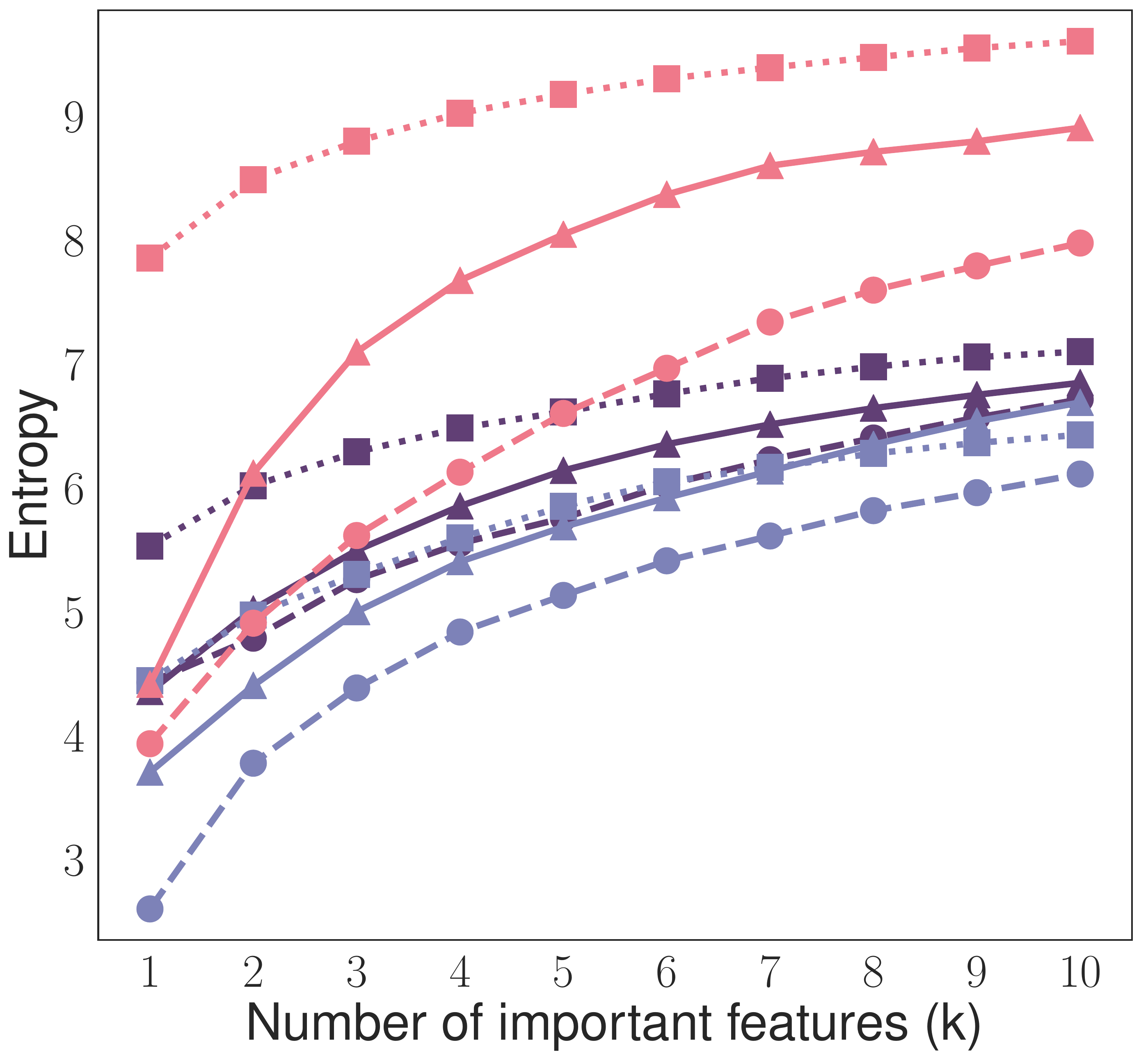}
  \caption{Deception}
  \label{fig:deception_models_entropy}
\end{subfigure}
\caption{The entropy of important features. 
In general, BERT generates more diverse important features than SVM ($\ell_1$) and XGBoost.
}
\label{fig:entropy_supp}
\end{figure*}

\para{Entropy.}
Deep learning models generate more diverse important features than traditional models.
See \figref{fig:entropy_supp}.

\begin{figure*}[t]
\centering
\begin{subfigure}[t]{0.443\textwidth}
  \centering
  \includegraphics[width=\textwidth]{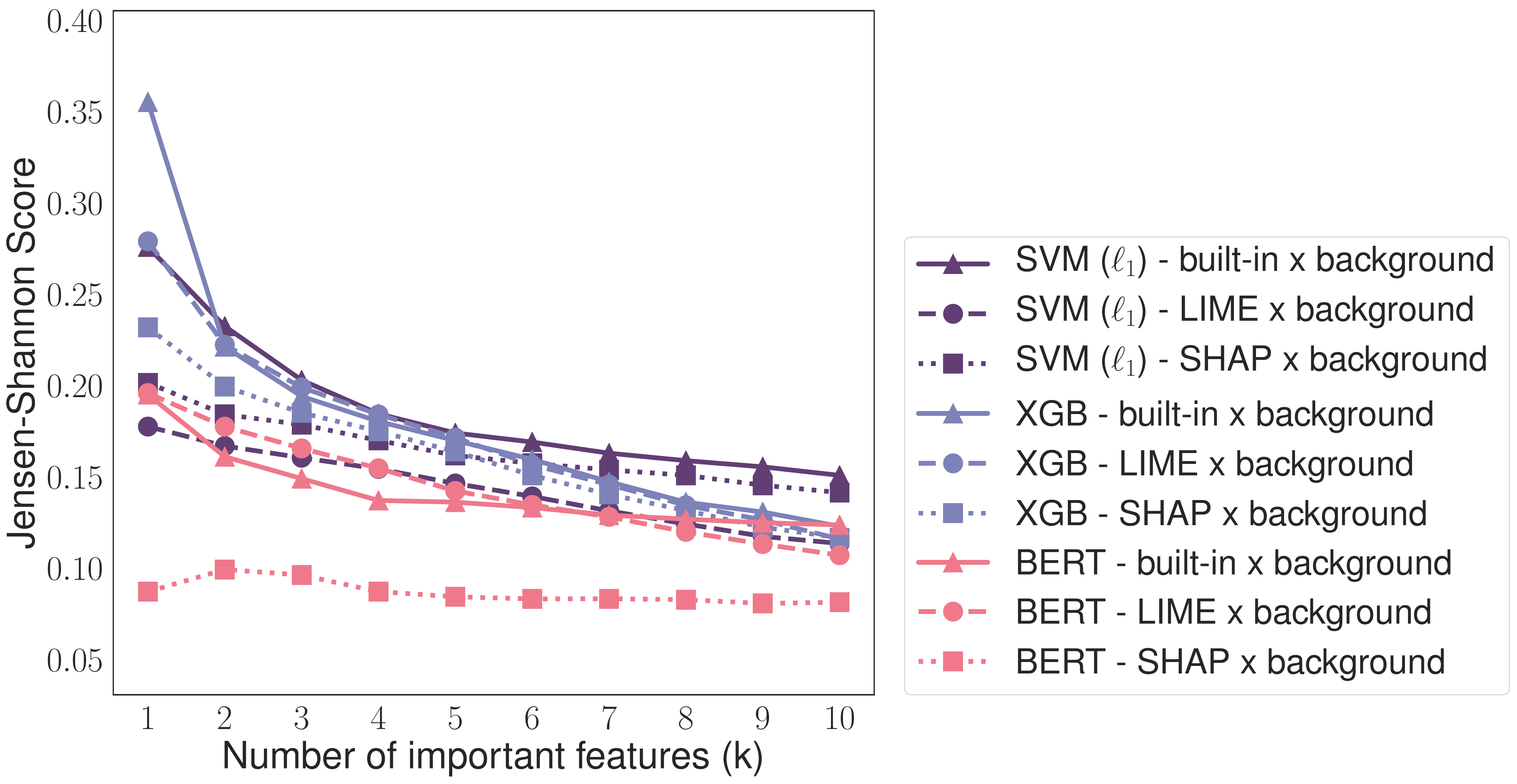}
  \caption{Yelp}
  \label{fig:yelp_models_jensenshannon_pos}
\end{subfigure}
\hfill
\begin{subfigure}[t]{0.275\textwidth}
  \centering
  \includegraphics[trim=0 0 8.2in 0,clip,width=\textwidth]{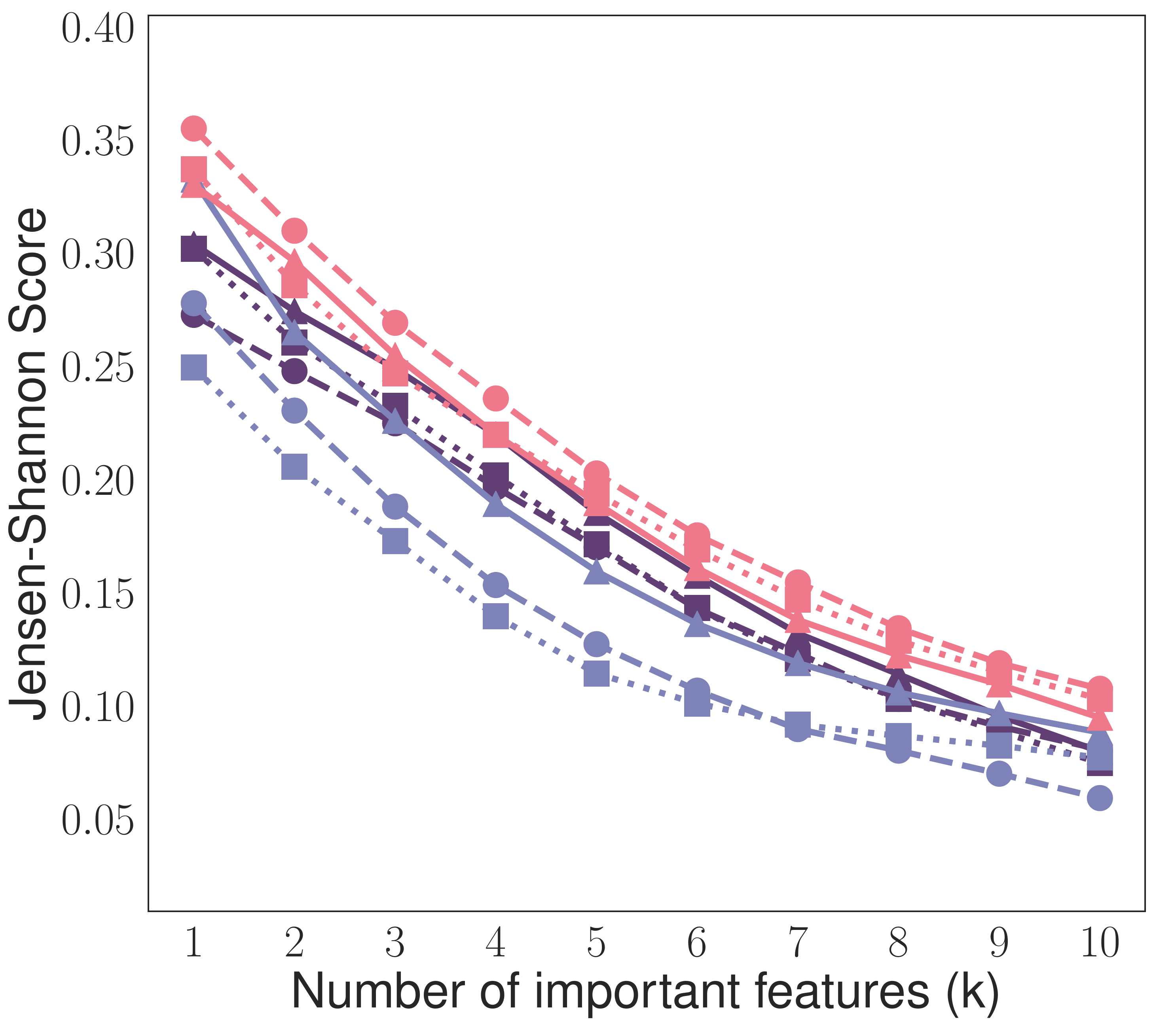}
  \caption{SST}
  \label{fig:sst_models_jensenshannon_pos}
\end{subfigure}
\hfill
\begin{subfigure}[t]{0.27\textwidth}
  \centering
  \includegraphics[trim=0 0 8.2in 0,clip,width=\textwidth]{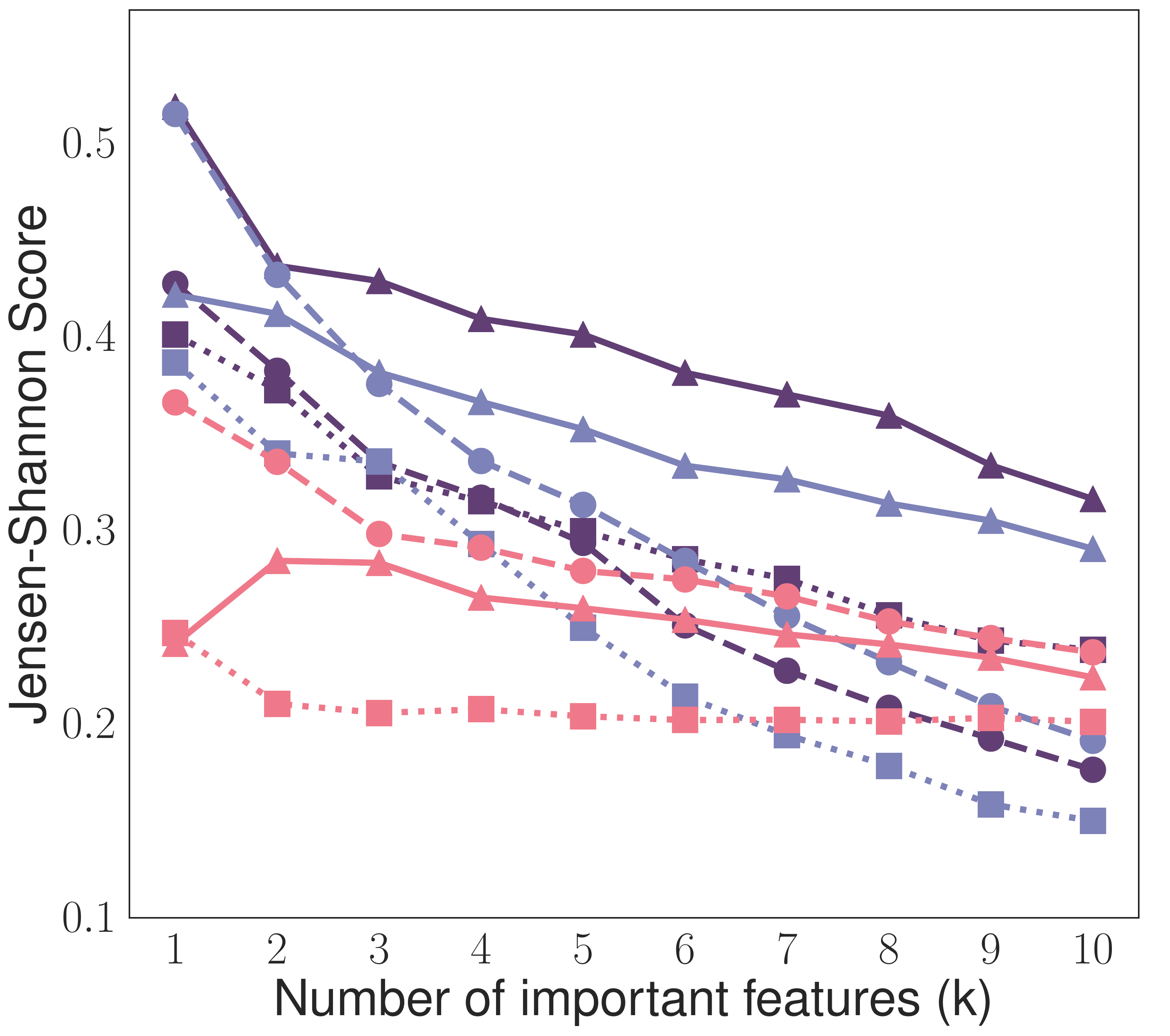}
  \caption{Deception}
  \label{fig:deception_models_jensenshannon_pos}
\end{subfigure}
\caption{Distance of the part-of-speech tag distributions between important features and all words (background).
Distance is generally smaller with post-hoc methods for all models, 
although some exceptions exist for LSTM with attention and BERT.
}
\label{fig:jensenshannon_pos}
\end{figure*}

\para{Jensen-shannon distance between POS.}
Distance of part-of-speech tag distributions between important features and all words is generally smaller with post-hoc methods for traditional models.
See \figref{fig:jensenshannon_pos}.

\end{document}